\documentclass{article}

\PassOptionsToPackage{numbers, compress}{natbib}
\usepackage[main,final]{neurips_2025}

\usepackage{amsmath,amssymb}
\usepackage{color}
\usepackage[dvipsnames]{xcolor}

\usepackage{times}
\usepackage{bm}
\usepackage{natbib}
\usepackage{algorithm}
\usepackage{algorithmic}
\usepackage{enumitem}
\usepackage{amssymb}
\usepackage{mathrsfs}
\usepackage{graphicx}
\usepackage{bbm}
\usepackage{mathtools}
\usepackage{rotating,subfigure}
\usepackage[flushleft]{threeparttable}
\usepackage{comment}
\usepackage{enumitem} %%%enumitem
\usepackage{arydshln}
\usepackage{pstricks}
\usepackage{tikz}

\usepackage[utf8]{inputenc} % allow utf-8 input
\usepackage[T1]{fontenc}    % use 8-bit T1 fonts
\usepackage[breaklinks=true,backref=section,colorlinks=true]{hyperref}       % hyperlinks
\hypersetup{
 citecolor=blue, linkcolor=blue}
\usepackage{url}            % simple URL typesetting
\usepackage{booktabs}       % professional-quality tables
\usepackage{amsfonts}       % blackboard math symbols
\usepackage{nicefrac}       % compact symbols for 1/2, etc.
\usepackage{microtype}      % microtypography
\usepackage{xcolor}         % colors
\definecolor{PCAcolor}{cmyk}{0, 0.85, 0.85, 0.50}
\newcommand{\px}{{\normalfont\texttt{\color{PCAcolor}PCA+}}}

\newcommand{\pxx}{{\normalfont\texttt{\color{PCAcolor}PCA++}}}

\title{PCA++: How Uniformity Induces Robustness to Background Noise in Contrastive Learning}
\author{Mingqi Wu\\
McGill University\\
Mila\\
\texttt{mingqi.wu@mail.mcgill.ca}
\And Qiang Sun \\
University of Toronto \\ MBZUAI\\
\texttt{qsunstats@gmail.com} \\
Corresponding author
\And 
Archer Y. Yang\\
McGill University \\ Mila\\
\texttt{archer.yang@mcgill.ca}\\
Corresponding author
}

\newcommand{\RR}{\mathbb{R}}
\newcommand{\EE}{\mathbbm{E}}
\newcommand{\diag}{{\rm diag}}

\ifx\BlackBox\undefined
\newcommand{\BlackBox}{\rule{1.5ex}{1.5ex}}  % end of proof
\fi

\ifx\QED\undefined
\def\QED{~\rule[-1pt]{5pt}{5pt}\par\medskip}
\fi

\ifx\proof\undefined
\newenvironment{proof}{\par\noindent{\bf Proof\ }}{\hfill\BlackBox\vspace{2mm}}
\fi

\ifx\theorem\undefined
\newtheorem{theorem}{Theorem}
\fi
\ifx\example\undefined
\newtheorem{example}[theorem]{Example}
\fi
\ifx\property\undefined
\newtheorem{property}{Property}
\fi
\ifx\lemma\undefined
\newtheorem{lemma}[theorem]{Lemma}
\fi
\ifx\proposition\undefined
\newtheorem{proposition}[theorem]{Proposition}
\fi
\ifx\remark\undefined
\newtheorem{remark}[theorem]{Remark}
\fi
\ifx\corollary\undefined
\newtheorem{corollary}[theorem]{Corollary}
\fi
\ifx\definition\undefined
\newtheorem{definition}[theorem]{Definition}
\fi
\ifx\conjecture\undefined
\newtheorem{conjecture}[theorem]{Conjecture}
\fi
\ifx\fact\undefined
\newtheorem{fact}[theorem]{Fact}
\fi
\ifx\claim\undefined
\newtheorem{claim}[theorem]{Claim}
\fi
\ifx\assumption\undefined
\newtheorem{assumption}[theorem]{Assumption}
\fi
\numberwithin{equation}{section}
\numberwithin{theorem}{section}

\def\##1\#{\begin{align}#1\end{align}}
\def\$#1\${\begin{align*}#1\end{align*}}

\newcommand{\tr}{\textrm{tr}}

\newcommand{\Rom}[1]{\text{\uppercase\expandafter{\romannumeral #1\relax}}}

\newcommand{\ndata}{n}
\newcommand{\ndim}{d}
\newcommand{\aspratio}{c}
\newcommand{\nsig}{k}
\newcommand{\nbkg}{m}
\newcommand{\neig}{s}

\newcommand{\qas}{\quad \text{a.s.}}
\newcommand{\npinfty}{\ndata,\ndim \rightarrow +\infty}
\newcommand{\overlineninfty}{\xrightarrow{\ndata \rightarrow +\infty}}
\newcommand{\spann}{\text{span}}

\newcommand{\noise}{\varepsilon}
\newcommand{\noisemat}{Z}
\newcommand{\sigload}{w}
\newcommand{\sigloadmat}{W}
\newcommand{\bkgload}{h}
\newcommand{\bkgloadmat}{H}
\newcommand{\errormat}{E}
\newcommand{\eigspace}{\mathcal{U}}

\newcommand{\covmatsample}{S}

\newcommand {\eigval}{\lambda}
\newcommand {\eigvec}{v}

\newcommand{\sig}{A}
\newcommand{\bkg}{B}

\newcommand{\iid}{\rm i.i.d.}

\usepackage{txfonts}

\usepackage{geometry}

\usepackage[subfigure]{tocloft}% http://ctan.org/pkg/tocloft
\makeatletter
\renewcommand{\numberline}[1]{%
  \@cftbsnum #1\@cftasnum~\@cftasnumb%
}
\makeatother

\usepackage{minitoc}
\usepackage{transparent}
\definecolor{adbskyblue}{HTML}{c1d8f0}
\definecolor{adbskyyellow}{HTML}{F2D22E}
\definecolor{adbskyred}{HTML}{ecccad}

\newcommand{\xcolorbox}[2]{%
  \tikz[baseline=(T.base)]\node[fill=#1, fill opacity=0.3, text opacity=1, inner sep=2pt, anchor=base] (T) {\ensuremath{#2}};%
}

\usepackage{subcaption}

\begin{document}

\maketitle

\begin{abstract}

High-dimensional data often contain low-dimensional signals obscured by structured background noise, which limits the effectiveness of standard PCA. Motivated by contrastive learning, we address the problem of recovering shared signal subspaces from positive pairs, paired observations sharing the same signal but differing in background. Our baseline, $\px$, uses alignment-only contrastive learning and succeeds when background variation is mild, but fails under strong noise or high-dimensional regimes. To address this, we introduce $\pxx$, a hard uniformity-constrained contrastive PCA that enforces identity covariance on projected features. $\pxx$ has a closed-form solution via a generalized eigenproblem, remains stable in high dimensions, and provably regularizes against background interference. We provide exact high-dimensional asymptotics in both fixed-aspect-ratio and growing-spike regimes, showing uniformity’s role in robust signal recovery. Empirically, $\pxx$ outperforms standard PCA and alignment-only $\px$ on simulations, corrupted-MNIST, and single-cell transcriptomics, reliably recovering condition-invariant structure. More broadly, we clarify uniformity’s role in contrastive learning, showing that explicit feature dispersion defends against structured noise and enhances robustness.

%%%High-dimensional data often contain low-dimensional signals obscured by structured background noise, which limits the effectiveness of standard PCA. Motivated by contrastive learning, we address the problem of recovering shared signal subspaces from positive pairs, paired observations sharing the same signal but differing in background. Our baseline, PCA+, uses alignment-only contrastive learning and succeeds when background variation is mild, but fails under strong noise or high-dimensional regimes. To address this, we introduce PCA++, a hard uniformity-constrained contrastive PCA that enforces identity covariance on projected features. PCA++ has a closed-form solution via a generalized eigenproblem, remains stable in high dimensions, and provably regularizes against background interference. We provide exact high-dimensional asymptotics in both fixed-aspect-ratio and growing-spike regimes, showing uniformity's role in robust signal recovery. Empirically, PCA++ outperforms standard PCA and alignment-only PCA+ on simulations, corrupted-MNIST, and single-cell transcriptomics, reliably recovering condition-invariant structure. More broadly, we clarify uniformity’s role in contrastive learning, showing that explicit feature dispersion defends against structured noise and enhances robustness.

\end{abstract}

\section{Introduction}

Real-world data often hides a simple, low-dimensional signal beneath layers of structured noise and random variation.  In genomics, batch effects blur true biological differences \citep{leek2010tackling,haghverdi2018batch}. In medical imaging, scanner differences obscure true clinical signals \citep{fortin2017harmonization}. In finance, the performance of individual assets is masked by market-wide trends \citep{fama1993common,campbell2001have}.  Yet standard Principal Component Analysis (PCA) indiscriminately captures all dominant directions, failing to distinguish the signal of interest from unwanted backgrounds.

\begin{figure}[ht]
\centering
\subfigure{    
\includegraphics[width=.45\linewidth]{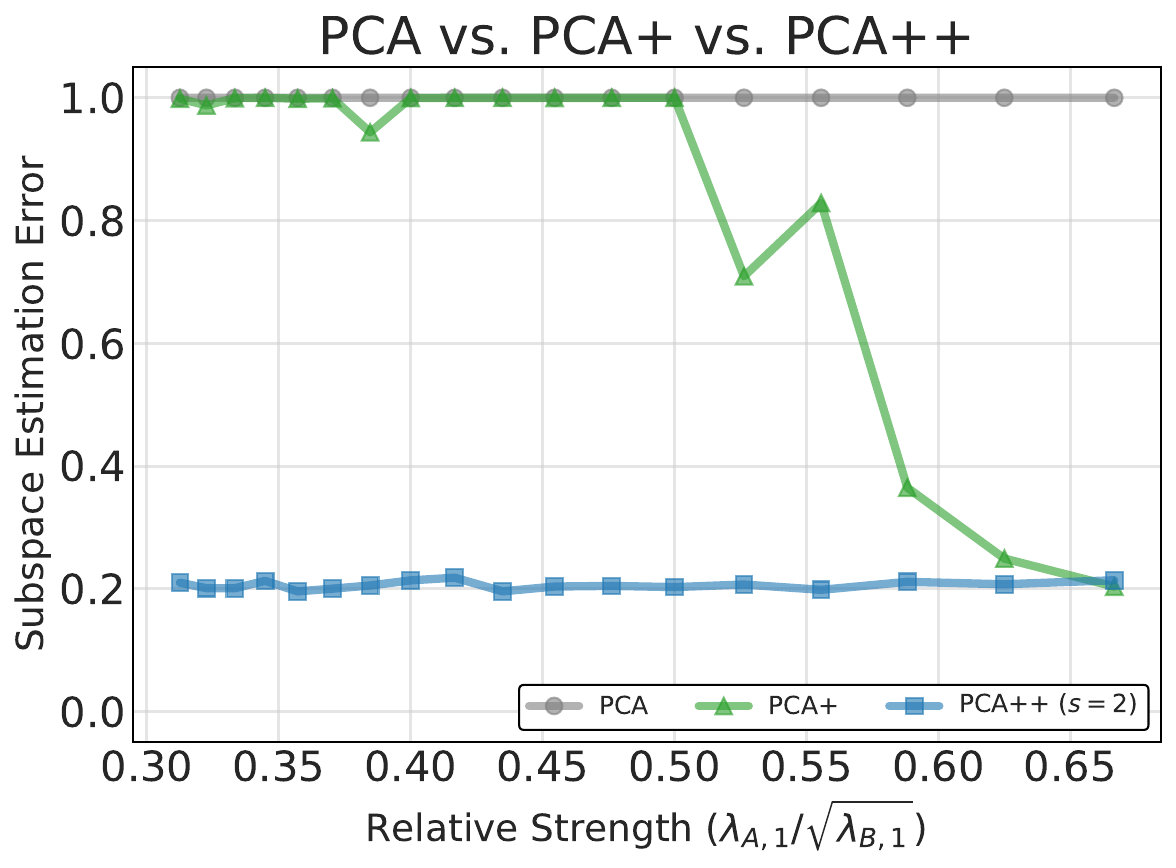}
}
\hspace{0.5cm}% ← adjust this length to taste
\subfigure{
\vspace{1pt}
\includegraphics[width=.45\linewidth]{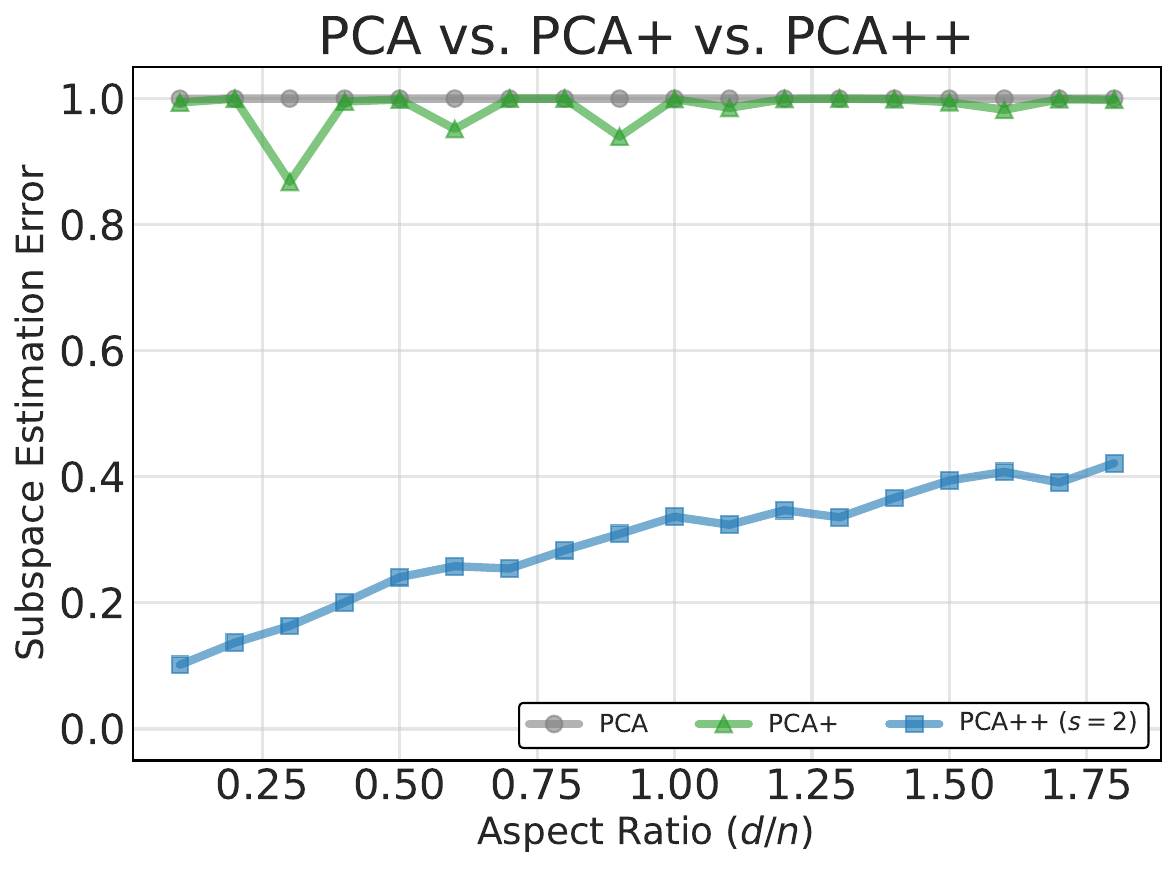}
}
    \caption{\small Subspace estimation error for standard PCA, $\px$, $\pxx$. Results are for Example \ref{ex:counterexample}.
    \textbf{Left:} (varying relative strength of the signal $\lambda_{A,1}/\sqrt{\lambda_{B,1}}$)  As background strength grows, $\px$ deteriorates sharply while $\pxx$ keeps its error uniformly low.
    \textbf{Right:} (varying aspect ratio $d/n$) Across all regimes, $\pxx$ outperforms both PCA and $\px$.
    }
    \label{fig:pcaplot_1}
\end{figure}

\textbf{Alignment-only contrastive learning and $\px$.} Contrastive learning \citep{gutmann2010noise,oord2018representation,chen2020simple,chen2021exploring} offers a natural remedy. By comparing paired datasets $X,X^+\in \RR^{n\times d}$, each with $n$ samples in $d$ dimensions that share the same signal but experience different backgrounds, we can cancel out unwanted variation. 
Motivated by this idea, we introduce $\px$, a vanilla contrastive PCA method that forms 
a ``contrastive" covariance $S_n^+ = \frac{1}{2n}(X^\top X^+ + X^{+\top} X)$ from positive pairs, and applies ordinary PCA to $S_n^+$. At the population level, $\px$ perfectly recovers the shared signal subspace. In finite‐sample, low‐dimensional regimes, $\px$ again succeeds. However, as background spikes (eigenvalue $\sqrt{\lambda_{B,1}}$) increase relative to signal spikes (Figure \ref{fig:pcaplot_1}, left) or the ambient dimension $d$ grows relative to $n$ (Figure \ref{fig:pcaplot_1}, right), its estimated directions can drift into background‐dominated components and miss the true signal.

\textbf{Hard uniformity constraint and $\pxx$.} To guard against this failure mode, we propose $\pxx$, which augments the contrastive objective with a ``hard uniformity” constraint \citep{wang2020understanding}: the projected features must have identity covariance. Equivalently, $\pxx$ solves
\begin{equation}
\max_{V\in\RR^{\ndim\times\nsig}}\ \tr(V^{\top}\covmatsample_{\ndata}^{+}V),\qquad\text{s.t.}\qquad V^{\top}S_{n}V=I_{\nsig},
\end{equation}
where
$S_n = \frac{1}{\ndata}X^\top X$  is the usual sample covariance. 
This constraint, forcing the learned features to be uniformly distributed, acts like an ``even‐spread" regularizer, protecting the signal
representation from distortion by strong background noise. 
We show that $\pxx$ admits a closed‐form solution via a \emph{generalized eigenproblem} and remains stable (with a low‐rank truncation of $S_n$) even when $d\gg n$. Empirically, $\pxx$ maintains tight alignment with the ground‐truth signal across a wide range of $d/n$, outperforming both standard PCA and $\px$ (Figure \ref{fig:pcaplot_1}).

\textbf{Theoretical guarantees.} We analyze $\px$ and $\pxx$ under a tractable \textbf{contrastive factor model} where each paired observation decomposes as orthogonal signal and background. Our main results include: (1). \emph{Population-level consistency} of $\px$, plus finite-sample bounds on its subspace error--and a precise characterization of when strong background spikes overwhelm it. (2). \emph{Exact high‐dimensional asymptotics} for $\pxx$ in two regimes: (i) Fixed aspect ratio $d/n\rightarrow c>0$, under a Baik-Ben Arous-P\'{e}ch\'{e} (BBP) detectability condition~\citep{baik2006eigenvalues}, we derive a closed‐form limit for the subspace error; (ii) Growing‐spike regime: when each spike scales with $d/n$,  the limiting subspace estimation error simplifies to a function of the weakest signal's effective signal-to-noise-ratio. These analyses quantify how the uniformity constraint regularizes against background interference.

\textbf{Broader impact on contrastive learning.} Our work also sheds light on an open question in contrastive learning theory: the practical role of uniformity. Recent theoretical results~\citep{wang2020understanding,pmlr-v195-parulekar23a} show that contrastive losses naturally encourage uniformity while preserving meaningful class structures. Yet, the precise theoretical reason uniformity improves downstream performance, particularly in noisy, high-dimensional environments, remained unclear. We rigorously show how an explicit uniformity constraint can defend against structured noise in high-dimensional regimes, clarifying why uniformity enhances robustness and generalization.

In summary, our contributions to the community include:
\begin{enumerate}
    \item A novel contrastive PCA framework that leverages paired observations to perfectly isolate shared signal subspaces at the population level and provably bounds finite-sample subspace recovery error under a tractable linear factor model.
    \item A hard uniformity–constrained $\pxx$ algorithm that enforces identity covariance on projected features, admits a closed-form generalized eigenproblem solution, and remains stable even when background eigenvalues and ambient dimension grow arbitrarily large.
    \item Exact high‐dimensional asymptotic characterizations of $\pxx$ in both fixed aspect‐ratio and growing‐spike regimes, yielding closed‐form limits for subspace estimation error that quantify how uniformity regularizes against structured noise.
    \item A theoretical explanation for uniformity’s power in contrastive learning, rigorously showing how enforcing feature dispersion defends signal recovery from strong background interference and clarifies uniformity’s role as a robust regularizer.
\end{enumerate}

\subsection{Related work}

\textbf{Foundations of contrastive learning.}
Early theory framed InfoNCE through mutual information (MI) maximization~\citep{oord2018representation,bachman2019learning}, and initial analyses linked the contrastive objective to downstream class separability guarantees~\citep{saunshi2019theoretical}. But MI alone proved insufficient: tighter MI bounds sometimes hurt performance~\citep{mcallester2020formal,tschannen2019mutual}. A major advance came with the \emph{alignment–uniformity} framework of~\citet{wang2020understanding}, showing that InfoNCE naturally pulls positive pairs together while spreading features uniformly on the hypersphere. This insight inspired methods like Barlow Twins~\citep{zbontar2021barlow} and VICReg~\citep{bardes2021vicreg}, which explicitly shape feature covariances, and it was further extended by~\citet{chen2021intriguing}, who parameterized alignment and uniformity terms directly. More recently, \citet{pmlr-v195-parulekar23a} proved that InfoNCE promotes uniformity in the population limit, and other work has highlighted how inductive biases steer the final representations~\citep{saunshi2022understanding,haochen2022theoretical}. Yet a precise, theory‐grounded explanation for why uniformity bolsters robustness, especially under structured noise, remains open, and it motivates our current study.

\textbf{Spectral views.}
Another strand casts contrastive learning as a form of spectral learning. The spectral contrastive loss~\citep{haochen2021provable,haochen2022beyond} directly manipulates feature covariances, and in fact it and InfoNCE share the same population solution~\citep{Daniel2023}. This spectral view connects CL to graph‐based dimensionality reduction (e.g.\ t‐SNE and spectral clustering) \citep{tan2024contrastive}.

\textbf{Linear models and high-dimensional analysis.} 
In parallel, tractable linear and asymptotic models have shed light on CL’s strengths. For instance, \citet{ji2023power} showed that CL recovers latent signal directions more reliably than classic unsupervised methods, while \citet{bansal2024understanding} used Gaussian mixtures to prove InfoNCE can identify optimal low‐dimensional subspaces. Our work also builds on a tractable model, embedding a linear contrastive factor model in high-dimensional random matrix theory to isolate how a hard uniformity constraint protects signal recovery under strong, structured backgrounds.

{\color{black}
\textbf{Foreground-background cPCA.} A significant line of work, initiated by \texttt{cPCA} \citep{abid2018exploring}, seeks directions of high variance in a \emph{foreground} dataset that are absent in a separate \emph{background} dataset. This methodology, including subsequent probabilistic and sparse extensions \citep{li2020probabilistic, boileau2020exploring, zhou2023sparse, zhang2025contrastive}, is fundamentally designed for \emph{case-control} settings in which a pure background dataset is explicitly available. Formally, these methods take as input two distinct datasets: a target dataset $X$, which contains both signal and nuisance variation, and a background-only dataset $Y$, which contains only nuisance variation. In contrast, our setting is structurally different. The proposed methods ($\px$ and $\pxx$) address the \emph{positive-pair} scenario, which is central to modern self-supervised learning (e.g., SimCLR \citep{chen2020simple}, MoCo \citep{he2020momentum}). Instead of relying on a clean background sample, we are given paired observations $(X, X^{+})$ that share the same latent signal but are corrupted by independent background variations. The objective is to identify directions that are \emph{shared across positive pairs}, i.e., invariant to nuisance variability, without requiring an explicit background dataset. See additional discussion in Appendix \ref{app:cpca_pcaplus}.

\textbf{Canonical correlation analysis (CCA).}
Although there is a conceptual connection between our work and CCA  \citep{cca}, the two approaches differ fundamentally. Both aim to identify shared structure between paired datasets, but they impose different objectives and constraints, which leads to substantial performance differences in noisy, high-dimensional settings. A detailed discussion of the relationship to CCA is provided in Appendix \ref{app:cca}.

}

\section{Contrastive Factor Model}\label{secPre}

We observe paired high‐dimensional samples $\{(x_{i}, x_{i}^+)\}_{i=1}^{n}$
generated by a \textbf{contrastive factor model}, which decomposes each point into a shared signal, an independent background, and noise:
\begin{equation}\label{eq:model}
x_{i} =  \xcolorbox{adbskyyellow}{\sig\sigload_{i}} + \xcolorbox{adbskyred}{\bkg \bkgload_{i}} + \xcolorbox{adbskyred}{\noise_{i}},\qquad
x_{i}^{+} =  \xcolorbox{adbskyyellow}{\sig\sigload_{i}} + \xcolorbox{adbskyblue}{\bkg \bkgload_{i}'} + \xcolorbox{adbskyblue}{\noise_{i}'}, \qquad i=1,
\ldots,n,
\end{equation}
where $\sig\in \RR^{\ndim \times \nsig}$ and $\bkg\in \RR^{\ndim \times \nbkg}$ have orthonormal columns spanning the signal and background subspaces, respectively; $\sigload_{i}\in \RR^{k}$ is the \emph{shared} signal factor, while $\bkgload_{i}$, $\bkgload_{i}'\in \RR^{m}$ are the independent background factors; $\noise_{i}$, $\noise_{i}'\in \RR^{d}$ are independent \emph{random noise}. 

Equivalently, we may write
\[
\sig \coloneqq \left[ \sqrt{\eigval_{\sig,1}} \eigvec_{\sig,1}, \ldots, \sqrt{\eigval_{\sig,\nsig}} \eigvec_{\sig,\nsig} \right], \quad
\bkg \coloneqq \left[ \sqrt{\eigval_{\bkg,1}} \eigvec_{\bkg,1}, \ldots, \sqrt{\eigval_{\bkg,\nbkg}} \eigvec_{\bkg,\nbkg} \right],
\]
each column $\eigvec_{\sig,j}$ (resp. $\eigvec_{\bkg,j}$) is a principal direction with variance
$\eigval_{\sig,j}$ (resp. $\eigval_{\bkg,j}$). 
By contrasting each pair
$(x_i,x_i^+)$, which share the same signal $\xcolorbox{adbskyyellow}{\sig\sigload_{i}}$ but differ in background $\xcolorbox{adbskyred}{\bkg \bkgload_{i}}$,  $\xcolorbox{adbskyblue}{\bkg \bkgload_{i}'}$ and noise, our goal is to isolate and recover the low‐dimensional signal subspace spanned by the columns of $A$.

To make the model analytically tractable, we impose three assumptions:

\begin{assumption}[Orthogonal signal and background]
\label{asm:orthogonal}
The column spaces of the signal and background loading matrices are mutually orthogonal, i.e., $\mathrm{span}(\sig)\,\perp\,\mathrm{span}(\bkg)$.
\end{assumption}

\begin{assumption}[Gaussian latent factors]
\label{asm:latent_gaussian}
$\sigload_i \;\overset{\mathrm{iid}}{\sim}\;\mathcal N(0, I_{\nsig})$, $\bkgload_i,\;\bkgload_i' \;\overset{\mathrm{iid}}{\sim}\;\mathcal N(0, I_{\nbkg})$, all independent across \(i\).
\end{assumption}

\begin{assumption}[Isotropic  noise]
\label{asm:noise}
$\noise_i,\noise_{i}'\;\overset{\mathrm{iid}}{\sim}\;\mathcal N(0,\,I_{\ndim}),$
independent of $z_i$, $w_i$, $w_{i}'$.
\end{assumption}

{\color{black}
\textbf{On the orthogonality assumption.} Orthogonality assumption is made primarily for analytical tractability by forcing the signal and background into disjoint subspaces. Our framework's core mechanism is, however, robust to its violation. The key insight is that even if signal and background subspaces overlap, the \emph{contrastive energy} (as analyzed in Lemma \ref{lm:highdim_bound}) of the shared directions remains strictly positive, while that of pure background directions is zero. This allows $\pxx$ to distinguish the full signal space from the background. The main effect of overlap is a reduction in the generalized eigenvalue for the shared directions, but they remain detectable.  

\textbf{On the Gaussian latent factor and noise assumption.} This assumption was also made for analytical convenience, as it allows for the clean derivation of exact constants and closed-form error rates. However, we expect the core results to hold more generally for sub-Gaussian distributions. Many of the key results from random matrix theory that we rely on (e.g., in Lemmas \ref{lm:HDspike}
and \ref{lm:HDspike2}) have well-known extensions beyond the Gaussian case, often requiring only a few finite moments.  

\textbf{On the isotropic noise assumption.} We fix the noise variance to one without loss of generality, since any other scale can be absorbed into the singular values of $A$ and $B$. 

We discuss the potential to relax these assumptions in Appendix~\ref{sec:Assumption}.}

\section{\texorpdfstring{$\px$}{px}: Contrastive PCA via Alignment Only}\label{sec:cpca}

A key insight from \cite{wang2020understanding,chen2021intriguing,pmlr-v195-parulekar23a} is that contrastive objectives implicitly promote two complementary geometric forces in representation space: \textbf{Alignment} (bringing positive pairs close), and \textbf{Uniformity} (evenly spreading all features on the hypersphere). Specifically, for a feature map $f:\mathcal{X}\rightarrow\RR^k$,  a general \emph{spectral contrastive learning} objective \citep{haochen2022beyond}  with uniformity‐weight $\tau>0$ is 
\begin{equation}\label{eq:full_constrastive_loss}
\mathcal{L}_{\text{sp}}(f) = \underset{(x,x^{+})}{\EE}\left[\;\big\|f(x)-f(x^{+})\; \big\|_{2}^{2}\right] \ + \  \tau\cdot \Big\|\;\underset{x}{\EE}\left[f(x)f(x)^{\top}\right]-I_{k}\;\Big\|_{F}.
\end{equation}
{\color{black} In this work, we consider a closely related objective, adopting the alignment term from \cite{haochen2021provable}:
\begin{equation}\label{eq:full_constrastive_loss1}
\mathcal{L}(f) = \underbrace{-\underset{(x,x^{+})}{\EE}\left[\; f(x)^{\top}f(x^{+})    \right]}_{\text{\color{PCAcolor}\emph{alignment}}} \ + \  \tau\cdot\underbrace{\Big\|\;\underset{x}{\EE}\left[f(x)f(x)^{\top}\right]-I_{k}\;\Big\|_{F}}_{\text{\color{PCAcolor}\emph{uniformity}}}.
\end{equation}
Both objectives in \eqref{eq:full_constrastive_loss} and \eqref{eq:full_constrastive_loss1} capture the alignment-uniformity trade-off observed in many popular contrastive learning frameworks \citep{chen2020simple, oord2018representation, NIPS2016_6b180037, 8578491}.
}

In this section, we ``turn off" uniformity ($\tau=0$) in \eqref{eq:full_constrastive_loss1} and study a linear encoder that optimizes \emph{only} alignment under our contrastive factor model; the following section then analyzes the effect of enforcing uniformity. We show that an “alignment‐only” method ($\px$), i.e.,  applying PCA to the contrastive covariance, can successfully recover the signal subspace when background variation is mild. However, in high‐dimensional regimes with strong background noise, this pure‐alignment approach can fail completely. This failure highlights the necessity of a uniformity‐type constraint: by enforcing feature dispersion, we regain robust signal recovery even when background components overwhelmingly dominate.

To isolate the role of alignment in contrastive learning, we study a linear encoder
\[
f(x) = V^\top x,\quad V \in \RR^{\ndim \times \nsig},
\]
and maximize alignment between positive pairs $(x_i, x_i^+)$ without  uniformity objective ($\tau=0$) in \eqref{eq:full_constrastive_loss1}:
\[
\min_{V^\top V = I_k}\;-\frac{1}{n}\sum_{i=1}^n  (V^{\top} x_i)^{\top} (V^{\top} x^+_i)
\quad\Leftrightarrow\quad
\max_{V^\top V = I_k}\;\tr \left(\frac{1}{n} V^\top X^\top X^+\,V\right),
\]
where $-\frac{1}{n}\sum_{i=1}^n  (V^{\top} x_i)^{\top} (V^{\top} x^+_i)$ is the empirical version of $-\EE[(V^{\top}x)^{\top}(V^{\top} x^{+})]$, and $X = (x_1, \ldots, x_{\ndata})^\top$, 
$X^{+} = (x_1^{+}, \ldots, x_{\ndata}^{+})^\top$ with $X,X^{+} \in \RR^{\ndata \times \ndim}$. To prevent representational collapse in such an alignment-focused setup, an orthogonality constraint $V^\top V = I_{\nsig}$ is necessary. Note that here $\frac{1}{n}X^\top X^{+}$ need not be symmetric, so its eigenvalues can be complex and its eigenvectors need not be orthogonal. A simple fix is to symmetrize  
\[
\covmatsample_{\ndata}^{+} = \frac{1}{2 \ndata}(X^\top X^{+} + X^{+\top} X)
\]
which is real-symmetric and shares the same population expectation (shown in Theorem \ref{thm:consistency_vanillaPCA}). Plugging $\covmatsample_{\ndata}^{+}$ into the alignment objective yields
\begin{equation}
\max_{V^\top V = I_{\nsig}}  \ \tr (V^\top \covmatsample_{n}^+ V),
\end{equation}
which is exactly PCA on the “contrastive” covariance $\covmatsample_{\ndata}^{+}$.
We call this method $\px$--our baseline that captures pure alignment under an orthogonality constraint. 

The estimated signal subspace is taken to be the span of the top $k$ eigenvectors of $S_{n}^{+}$. We first show that $S_{n}^{+}$ is an unbiased estimator of the true signal covariance:
\begin{theorem}[Unbiasedness of the contrastive covariance estimator]\label{thm:consistency_vanillaPCA}

Under Assumptions \ref{asm:orthogonal}--\ref{asm:noise},  the contrastive covariance estimator $\covmatsample_{\ndata}^{+}$ satisfies $\EE\bigl[\covmatsample_{\ndata}^{+}] =\sig \sig^\top.$
\end{theorem}
In other words, in expectation, the contrastive covariance matrix $\covmatsample_{\ndata}^{+}$ perfectly cancels out both background and noise, recovering the population signal covariance $\sig \sig^\top$.

Having established the behaviour of $\px$ estimator at the population level, we now turn to its finite-sample performance. Our main result is a non-asymptotic upper bound on the subspace estimation error of $\px$, incurred by applying PCA to the contrastive covariance matrix \( \covmatsample_{\ndata}^{+} \). 
Denote the true subspace span by columns of $A$ by \(\eigspace_{\sig} \coloneqq \spann\{\eigvec_{\sig,1}, \ldots, \eigvec_{\sig,\nsig}\}\) and let  \(\widehat{\eigspace}_{\sig}\) be our estimator. We measure the estimation error between \(\eigspace_{\sig}\) and \(\widehat{\eigspace}_{\sig}\) via their subspace distance  based on the \emph{principal angles} between \(\eigspace_{\sig}\) and \(\widehat{\eigspace}_{\sig}\). The definition of the subspace distance is provided in Appendix \ref{sec:principleangles}. 
 
\begin{theorem}[Finite-sample performance of $\px$]\label{thm:CPCAlowdim}
    Suppose Assumptions  \ref{asm:orthogonal}--\ref{asm:noise} hold, and that the sample size $\ndata$ obeys:
    \begin{equation*}
        \ndata \geq  \frac{C}{\eigval_{\sig, \nsig}^{2}} \left( \nsig \eigval_{\sig, 1}^{2} + \nbkg \eigval_{\bkg, 1}^{2} + \max(\nsig, \nbkg) \eigval_{\sig, 1}\eigval_{\bkg, 1} + \ndim \left(\eigval_{\sig, 1} + \eigval_{\bkg, 1} + 1\right) \right) \log^3 (\ndata + \ndim),
    \end{equation*}
    for a sufficiently large universal $C$. Then, with probability at least $1 - \mathcal{O}((\ndata + \ndim)^{-10})$, the distance between the estimated subspace $\widehat{\eigspace}_{\sig}$ and the true subspace $\eigspace_{\sig}$ satisfies:
\begin{equation*}
    \begin{split}
    \operatorname{dist}(\widehat{\eigspace}_{\sig}, \eigspace_{A}) &\lesssim \frac{1}{\lambda_{\sig, \nsig}} \Bigg( \eigval_{\sig, 1} \sqrt{\frac{\nsig}{\ndata}} + \eigval_{\bkg, 1} \sqrt{\frac{\nbkg}{\ndata}} + \sqrt{\eigval_{\sig, 1}\eigval_{\bkg, 1}} \sqrt{\frac{\max(\nsig, \nbkg)}{\ndata}}\\ 
    &\quad + \left(\sqrt{\eigval_{\sig, 1}} + \sqrt{\eigval_{\bkg, 1}} + 1\right)\sqrt{\frac{\ndim}{\ndata}} \ \Bigg) \log^{1/2}(\ndata + \ndim).
    \end{split}
\end{equation*}
\end{theorem}
This bound shows that, when the smallest signal spike $\lambda_{\sig, \nsig}$ is large relative to the largest background spikes $\eigval_{\bkg, 1}$ and the noise level, $\px$ recovers the signal subspace accurately from finite data. However, if the background strength grows or the eigengap $\lambda_{\sig, \nsig}$ shrinks, the error can blow up. Indeed, in the extreme one‐signal-one‐background case below, standard $\px$ fails completely to align with the true signal:

\begin{example}[One‑signal, one‑background]\label{ex:counterexample}
    Let $k=m=1$ in model~\eqref{eq:model} and
    \begin{equation*}
        \sig: = \left[ \sqrt{\eigval_{\sig, 1}} e_{1} \right] \quad \text{and} \quad \bkg: = \left[ \sqrt{\eigval_{\bkg, 1}} e_{2} \right],
    \end{equation*}
    where $e_{1} \coloneqq (1, 0, \dots, 0)^\top$ and $e_{2} \coloneqq (0, 1, 0, \dots, 0)^\top$ are the first two standard basis vectors in $\RR^{\ndim}$.
\end{example}

\begin{theorem}[Failure under strong background]\label{thm:counterexample}
    Under this model with $\ndim/\ndata \rightarrow \aspratio \in (0, + \infty)$ as $\npinfty$, the leading $\px$ direction $\hat{\eigvec}_{1}$  of $\covmatsample_{\ndata}^{+}$ satisfies
    \begin{equation*}
        \lim_{\npinfty} (\hat{\eigvec}_{1}^\top e_{1})^2 \leq 2 \frac{\eigval_{\sig, 1}}{\sqrt{\eigval_{\bkg, 1} \aspratio}}.
    \end{equation*}
\end{theorem}
This result highlights a limitation of $\px$: when the leading background eigenvalue becomes sufficiently large,  $\sqrt{\eigval_{\bkg,1}} \gtrsim \eigval_{\sig, 1}/\sqrt{\aspratio}$, the top empirical eigenvector fails to align with the true signal direction, with the alignment remaining bounded away from one. To address this issue, we propose a \emph{constrained} contrastive PCA method in the following section that incorporates uniformity to suppress background interference and achieve robust signal recovery even under strong background conditions.

\section{Enforcing Uniformity: \texorpdfstring{$\pxx$}{pxx} and Its Analysis}\label{sec:ConcPCA}
% Let $\{(\hat{\eigval}_{i}, \hat{\eigvec}_{i}) \}_{j=1}^{\ndim}$ represent the eigenvalue-eigenvector pairs of the matrix $X^\top X/\ndata$, ordered such that $\eigval_{i}$ are in increasing order. The estimated subspace $\widehat{\eigspace}_{\sig}$ is defined as the span of the top $\nsig$ ($\nsig\leq d$) distinct vectors that maximize the following criterion:
% \begin{equation*}
% \begin{split}
% &\max_{\eigvec}\frac{1}{\eigval} \eigvec^\top \covmatsample_{n}^+ \eigvec,\\
% \text{s.t.} ~~ &(\eigval, \eigvec) \in \{(\hat{\eigval}_{i}, \hat{\eigvec}_{i}) \}_{j=1}^{\nsig+\nbkg}
% \end{split}
% \end{equation*}

The previous section exposed a flaw of pure‐alignment PCA ($\px$): it can no longer recover the true signal when background noise is too strong. We now show that enforcing perfect uniformity, i.e.,  constraining the projected features to have identity covariance, acts as a powerful regularizer, neutralizing even very strong background.

Concretely, we introduce the \emph{hard‐uniformity contrastive PCA} ($\pxx$):
\begin{equation}\label{eq:constrained_opt}
\max_{V\in\RR^{\ndim\times\nsig}}\ \underbrace{\tr(V^{\top}\covmatsample_{\ndata}^{+}V)}_{\text{\color{PCAcolor}\emph{alignment}}},\qquad\text{s.t.}\qquad\underbrace{V^{\top}S_{n}V=I_{\nsig}}_{\text{\color{PCAcolor}\emph{uniformity}}},
\end{equation}
where
$S_n = \frac{1}{\ndata}X^\top X$. The hard constraint $V^\top S_n V = I_k$ enforces $\frac{1}{n}\sum_{i=1}^{n}(Vx_i)(Vx_i)^\top=I_k$, i.e.  perfect uniformity of the projected features.  \eqref{eq:full_constrastive_loss}.

The $\pxx$ objective \eqref{eq:constrained_opt} can be solved in one shot via a single \emph{\color{PCAcolor}generalized eigenvalue decomposition} (see proof in Appendix \ref{sec:GEV_proof}), making it both conceptually simple and computationally efficient.   Specifically, one computes  
\begin{equation}\label{eq:geneigen}
    \covmatsample_{n}^+ \eigvec_{j} = \eigval_{j}  S_n \eigvec_{j}, \quad \forall~ j \in \{1, \dots, \ndim\}.
\end{equation}
for $\{\eigval_j,\eigvec_j\}$,  where $\covmatsample_{\ndata}^{+} = \frac{1}{2 \ndata}(X^\top X^{+} + X^{+\top} X)$ and $S_n = \frac{1}{\ndata}X^\top X$. The top-$\nsig$ generalized eigenvalues $\eigvec_1,\ldots,\eigvec_k$ (those with largest $\eigval_j$) form the columns of the $\pxx$ solution $V \in \RR^{\ndim \times \nsig}$. When $S_n$ is invertible, \eqref{eq:geneigen} reduces to the ordinary eigenproblem for $S_n^{-1}\covmatsample_n^+$.  For implementation details see Algorithm \ref{alg:generalized_eig} in Appendix \ref{sec:Generalized_eigenvalue}, and for a full derivation of generalized eigenvalue solvers refer to \citep{parlett1998symmetric,ghojogh2019eigenvalue}.

\begin{figure}[ht]
\centering
\subfigure{
\vspace{0pt}
    \includegraphics[width=.46\linewidth]{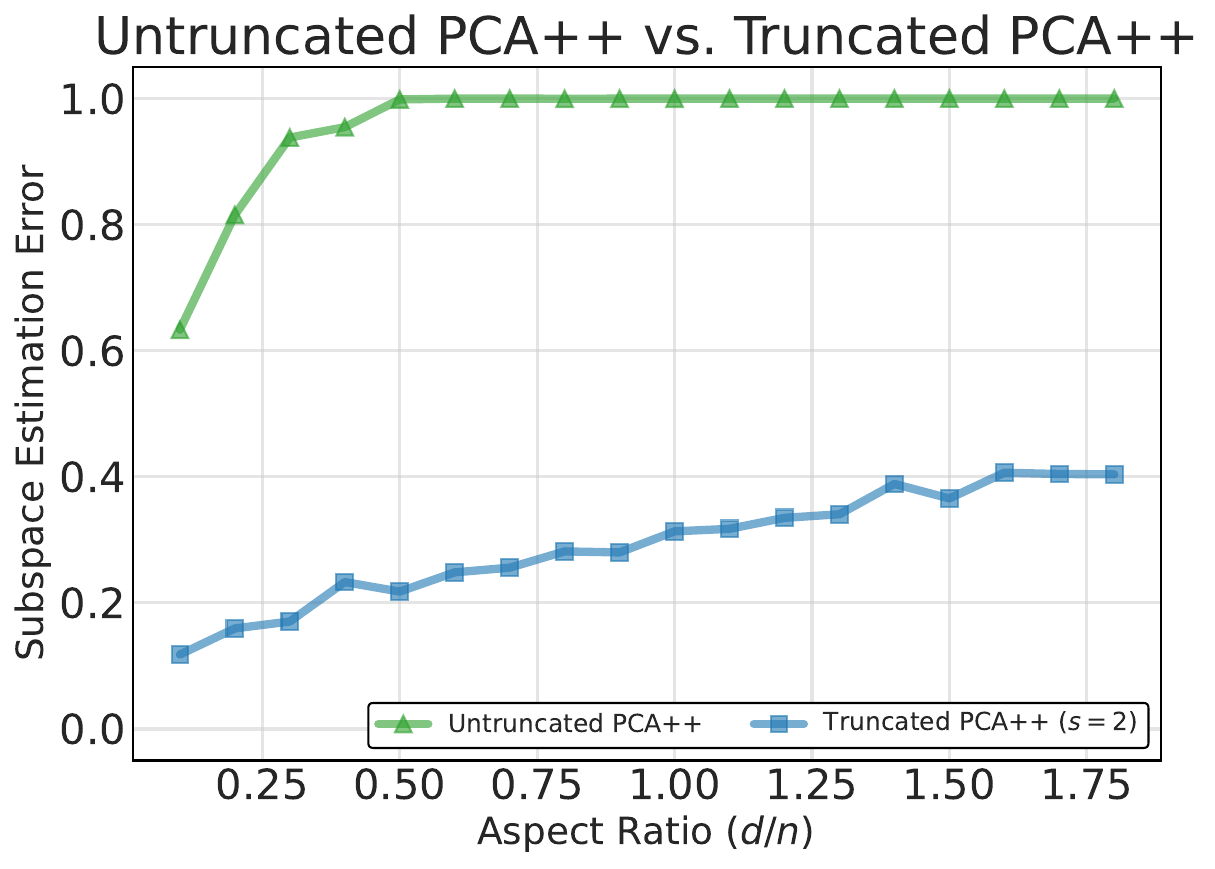}
}
\hspace{0.5cm}
\subfigure{
\vspace{0pt}
    \includegraphics[width=.45\linewidth]{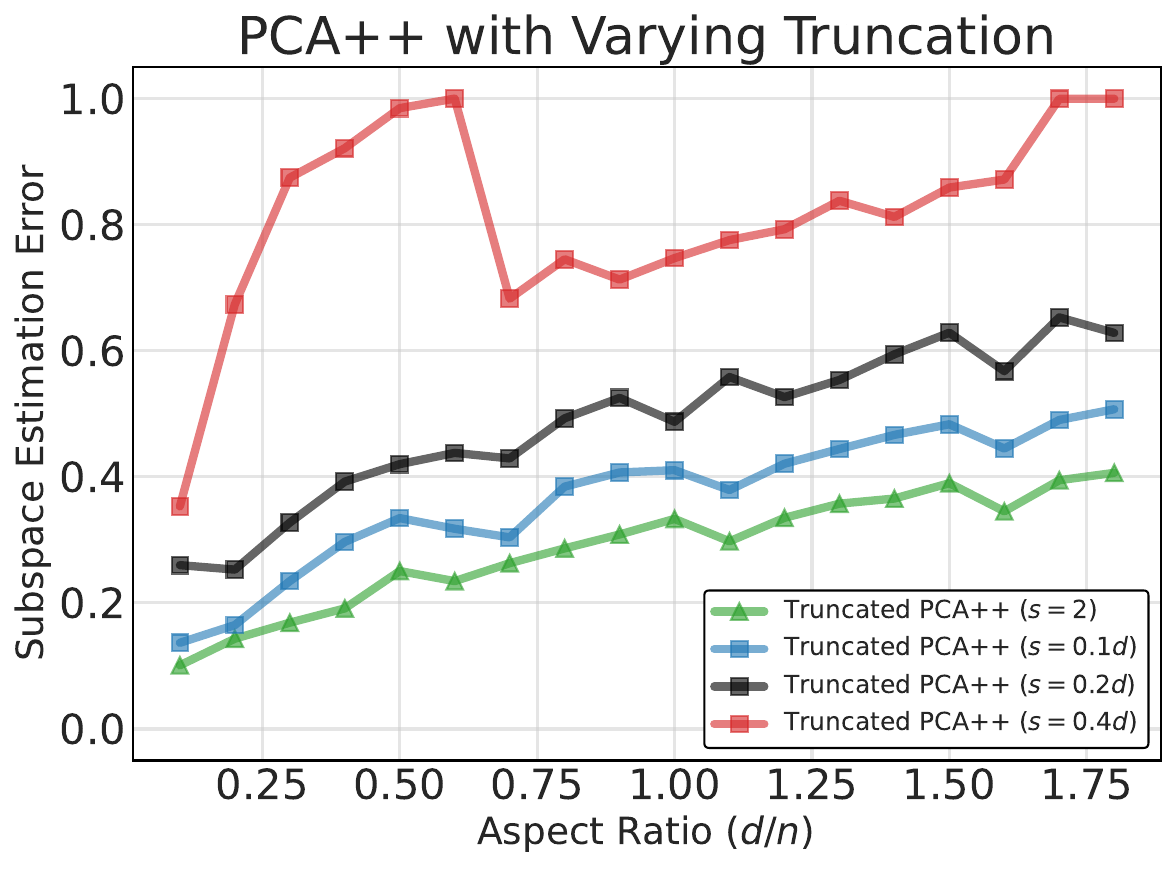}
}
    \caption{\small \textbf{Effect of covariance truncation on $\pxx$.} Results are for Example \ref{ex:counterexample}. \textbf{Left:} As $d/n$ increases, truncated $\pxx$ remains stable and accurate while untruncated $\pxx$ deteriorates sharply. \textbf{Right:} Truncated $\pxx$ with varying truncation ranks $s$ (fixed $s=2$; or $s$ as $0.1d, 0.2d, 0.4d$ of feature dimension $d$).}
    \label{fig:2}
\end{figure}

In very high dimensions, $S_n$ is often severely ill‐conditioned, so solving \eqref{eq:geneigen} directly may be numerically unstable. Figure \ref{fig:2} (left) shows how, even with the uniformity constraint, the generalized eigenvectors can drift away from the true signal when $d\gg n$. 
To restore stability, we replace $S_n$
by its rank-$s$ approximation $(S_n)_{\neig} \coloneqq \sum_{j=1}^{\neig} \eigval_{j} \eigvec_{j} \eigvec_{j}^\top$ where $\{\eigval_j,\eigvec_j\}_{j=1}^{s}$ are the top-$s$ eigenpairs of $S_n$, and $s\leq \ndim$ is a tuning parameter. We then solve
\begin{equation}\label{eq:truncopt}
\max_{V \in \RR^{\ndim \times \nsig}} \ \tr (V^\top \covmatsample_{\ndata}^+ V), \qquad
\text{s.t.} \qquad   V^\top (S_n)_{\neig} V = I_k,
\end{equation}
{\color{black}
By construction, $(S_n)_s$ is well‐conditioned on its $s$-dimensional leading subspace and discards the small, unstable directions, enforcing uniformity only where $S_n$ is reliable. As Figure~\ref{fig:2} (right) shows, an appropriate $s$ dramatically reduces error, whereas including too many dimensions re-introduces noise. Computationally, one simply replaces  $S_n$ with $(S_n)_s$ in Algorithm \ref{alg:generalized_eig} to obtain a stable solution. We provide detailed guidelines for choosing the value of this crucial hyperparameter in Appendix~\ref{app:hyperparam}.

Furthermore, this truncation strategy offers a significant computational advantage. Since we only need to find the top-$s$  generalized eigenvectors, we can avoid a full decomposition and instead employ efficient iterative solvers, such as the implicitly restarted Lanczos method (IRLM). This makes the approach scalable to very high-dimensional data, and the details of its computational complexity are discussed in Appendix~\ref{app:complex}.
}

We will analyze the truncated uniformity‐constrained contrastive PCA in two complementary high‐dimensional asymptotic regimes using tools from random matrix theory and the spiked covariance model \citep{Johnstone2001,baik2006eigenvalues,paul2007asymptotics}:
\begin{enumerate}
    \item[\color{PCAcolor}I.] \textbf{\color{PCAcolor}Fixed aspect ratio regime.} The signal eigenvalues $\{ \eigval_{\sig, j}\}_{j=1}^{\nsig}$ and background eigenvalues $\{ \eigval_{\bkg, j}\}_{j=1}^{\nbkg}$ remain constant, while the aspect ratio $\ndim/\ndata\rightarrow c > 0$ as both $\ndata,\ndim\rightarrow \infty$. 
    \item[\color{PCAcolor}II.] \textbf{\color{PCAcolor}Growing-spike regime.} Both signal spikes $\eigval_{\sig, j}$ and background eigenvalues $\eigval_{\bkg, j}$  diverge with $\ndata,\ndim\rightarrow \infty$, and the scaled aspect ratios $\frac{\ndim}{\ndata \eigval_{\sig, j}} \rightarrow \aspratio_{\sig, j}$ and $\frac{\ndim}{\ndata \eigval_{\bkg, j}} \rightarrow \aspratio_{\bkg, j}$, with $0 \leq \aspratio_{\sig, 1} < \dots < \aspratio_{\sig, \nsig} < \infty$ and $0 \leq \aspratio_{\bkg, 1} < \dots < \aspratio_{\bkg, \nbkg} < \infty$.
\end{enumerate}
Our analysis for the following high-dimensional regimes builds upon the asymptotic alignment between sample and population principal components of $S_n=\frac{1}{\ndata}X^\top X$ under model \eqref{eq:model}, as formalized in Lemma~\ref{lm:HDspike}. Denote the true population spikes and directions of $\EE[x x^\top]$ by $\{(\eigval_{\sig, j}, \eigvec_{\sig, j})\}_{j=1}^{\nsig}$ and $\{(\eigval_{\bkg, j},  \eigvec_{\bkg, j})\}_{j=1}^{\nbkg}$, sorted in descending order, and let $\{(\hat{\eigval}_{\sig, j}, \hat{\eigvec}_{\sig, j})\}_{j=1}^{\nsig}$ and $\{(\hat{\eigval}_{\bkg, j}, \hat{\eigvec}_{\bkg, j})\}_{j=1}^{\nbkg}$  be the corresponding sample eigenpairs obtained by matching each empirical eigenvalue of $S_n$ to its nearest population counterpart in magnitude. This matching preserves the separation between signal and background components in our high‐dimensional analysis. %Throughout our theoretical results for subspace estimation error (Theorems~\ref{thm:dist} and \ref{thm:dist2}), the distance between the estimated and true signal subspaces is quantified by the operator norm of the sine of their principal angles (See Appendix~\ref{sec:principleangles}).

\paragraph{\color{PCAcolor}I. Fixed aspect ratio regime.}

We now study the regime $\ndim/\ndata\rightarrow c > 0$ as both $\ndata,\ndim\rightarrow \infty$ and ask: when do the sample eigenvalues "stick out" of the Marcenko-Pastur bulk so that both signal and background directions remain identifiable?

\begin{assumption}[Detectable spikes]\label{assum:eigenvalue}
    The population eigenvalues $\{ \eigval_{\sig, j}\}_{j=1}^{\nsig}$ and $\{ \eigval_{\bkg, j}\}_{j=1}^{\nbkg}$ are all distinct and satisfy $\eigval_{\sig, j}$, $\eigval_{\bkg, j} \geq$ $\sqrt{\aspratio}$ for every $j$.
\end{assumption}

{\color{black}
\textbf{Remark.} This condition corresponds to the classical BBP threshold, a fundamental requirement in high-dimensional PCA for spiked covariance models. When the signal strength falls below this threshold, the leading eigenvalues of the sample covariance matrix are absorbed into the noise bulk of the Marcenko--Pastur distribution, rendering the signal statistically undetectable \citep{baik2006eigenvalues, paul2007asymptotics}. This phenomenon reflects an intrinsic information-theoretic limit of signal detection via PCA in the high-dimensional regime, rather than a limitation specific to our $\texttt{PCA++}$ method. Under this assumption, the hard-uniformity constrained PCA ($\texttt{PCA++}$) admits a simple and exact characterization of its asymptotic subspace error:
}

\begin{theorem}[Asymptotic subspace error under hard uniformity]\label{thm:dist}
    Under Assumptions \ref{asm:orthogonal}--\ref{asm:noise} and \ref{assum:eigenvalue}, and $s \geq k$, let $\widehat{\eigspace}_{\sig}$ be the top-$k$ subspace returned by the hard‐uniformity PCA in \eqref{eq:truncopt} and $\eigspace_{\sig}$ be the true population signal subspace. Then as $\npinfty$ with $\ndata / \ndim \rightarrow \aspratio \in (0, + \infty)$,
\begin{equation*}
    \operatorname{dist}(\widehat{\eigspace}_{\sig}, \mathcal{U}_{A})^2 \longrightarrow 1 - \frac{1 - \aspratio \eigval_{\sig, \nsig}^{-2}}{1 + \aspratio \eigval_{\sig, \nsig}^{-1}} \qas
\end{equation*}
\end{theorem}
We can see that when the weakest signal spike $\lambda_{A,k}\gg \sqrt{c}$, the error $1 - \frac{1 - \aspratio \eigval_{\sig, \nsig}^{-2}}{1 + \aspratio \eigval_{\sig, \nsig}^{-1}}$ is small, reflecting accurate recovery; As $\lambda_{A,k}\gg \sqrt{c}$ grows,  the error vanishes. Conversely, increasing the aspect ratio $c=d/n$ makes recovery harder and increases the limiting error.  Crucially, this result highlights the power of the uniformity constraint. Unlike the pure‐alignment method $\px$ (Example~\ref{ex:counterexample} and Theorem~\ref{thm:counterexample}), which can be overwhelmed by strong background spikes, the hard‐uniformity estimator $\pxx$
continues to recover the signal subspace, albeit with a controlled high‐dimensional bias, even when background eigenvalues grow arbitrarily large.
 
\paragraph{\color{PCAcolor}II. Growing-spike regime.}
In the growing-aspect-ratio regime, where the spikes grow proportionally to $\ndim/\ndata$ such that their "effective signal-to-noise ratios" converge to finite, distinct constants, we no longer require the lower‐bound from Assumption~\ref{assum:eigenvalue}: the growth of the eigenvalues ensures they remain asymptotically separable from the noise bulk. 
\begin{assumption}[Distinct growing spikes]\label{assum:eigenvalue2}
    The population eigenvalues $\{ \eigval_{\sig, j}\}_{j=1}^{\nsig}$ and $\{ \eigval_{\bkg, j}\}_{j=1}^{\nbkg}$ are all distinct.
\end{assumption}

{\color{black}
\textbf{Remark.} This assumption was made primarily for analytical convenience. With distinct eigenvalues, we can cleanly map each sample eigenvector to a population counterpart using existing results (Lemmas \ref{lm:HDspike} and \ref{lm:HDspike2}). \emph{Without this assumption}, if a set of eigenvalues were identical, we would instead estimate a  subspace  for those directions. Within that estimated subspace, it would be difficult to distinguish which basis vectors correspond to signal and which to background just by looking at $S_n$ alone, making it harder to explain how the uniformity constraint works on a per-direction basis.
The key insight is that even if the standard covariance $S_n$ has degenerate subspaces (i.e., multiple identical eigenvalues mixing signal and background components), the contrastive covariance $S_{n}^{+}$ resolves this ambiguity. Since $S_{n}^{+}$ has asymptotically zero energy on pure background directions, the generalized eigenvalue problem can still correctly identify the signal subspace and separate it from the background. We discuss the potential to relax this assumption in Appendix~\ref{app:discussion_ass43}.

}

The following theorem shows that,  the limiting subspace estimation error is a simple function of the weakest signal's effective SNR, $\aspratio_{\sig, k} = \frac{\ndim}{\ndata \eigval_{\sig, k}}$: 
\begin{theorem}\label{thm:dist2}
    Under Assumptions \ref{asm:orthogonal}--\ref{asm:noise} and \ref{assum:eigenvalue2}, and $s \geq k$,  as $\npinfty$,
\begin{equation*}
    \operatorname{dist}(\widehat{\eigspace}_{\sig}, \mathcal{U}_{A})^{2} \longrightarrow \frac{\aspratio_{\sig, \nsig}}{1 + \aspratio_{\sig, \nsig}} \qas
\end{equation*}
\end{theorem}
As $\aspratio_{\sig, k}$ decreases (i.e.\ the signal grows stronger relative to the ambient dimension $d/n$), the error vanishes. Moreover, by plugging $\aspratio_{\sig, j}  = \frac{\ndim}{\ndata \eigval_{\sig, j}}$ into the error expression from Theorem \ref{thm:dist}, one recovers exactly the same form in Theorem~\ref{thm:dist2}, confirming full consistency between the two high‐dimensional analyses.

\section{Experiments}\label{sec:experiments}

\paragraph{Simulation studies.}  
We designed a suite of simulations to test our theoretical predictions and to compare standard PCA, the alignment‐only method \(\px\), our uniformity‐constrained PCA++ (\(\pxx\)), and the corresponding high‐dimensional theory.  In every scenario, paired samples were drawn from the linear contrastive factor model (Eq.~\eqref{eq:model}), and performance was quantified by the sine of the principal angle between the estimated and true signal subspaces, averaged over fifty independent runs.

We first explored the role of truncation in stabilizing $\pxx$ when \(n=1000\) and \(\lambda_{A,1}=10,\;\lambda_{B,1}=500\).  Solving the generalized eigenproblem without truncation leads to erratic signal estimates once \(d/n\) exceeds about 1 (Figure~\ref{fig:2}, left).  However, projecting the sample covariance onto its top-\(s\) subspace with \(s=2\) fully restores recovery.  We then swept \(s\) over \(\{2,\,0.1d,\,0.2d,\,0.4d\}\) and found that moderate truncation (e.g.\ \(s=0.1d\)) achieves the best trade‐off: it discards the unstable noise directions while preserving enough dimensions to enforce the uniformity constraint (Figure~\ref{fig:2}, right).  
 
Next, we turned to the asymptotic regimes characterized by Theorems~\ref{thm:dist} and \ref{thm:dist2}.  Fixing $n=500$ and varying the aspect ratio $d/n$, we embedded a five‐dimensional signal subspace (variances [50, 25, 20, 15, 10]) in the first five coordinates and an orthogonal five‐dimensional background (variances [500, 400, 300, 200, 100]) in the last five.  Applying $\pxx$ with truncation rank $s=10$, we again computed the sine of the largest principal angle to the true signal, averaged over fifty runs.  

\begin{figure}[ht]\label{fig:high}
\centering
\subfigure{
\includegraphics[width=.45\linewidth]{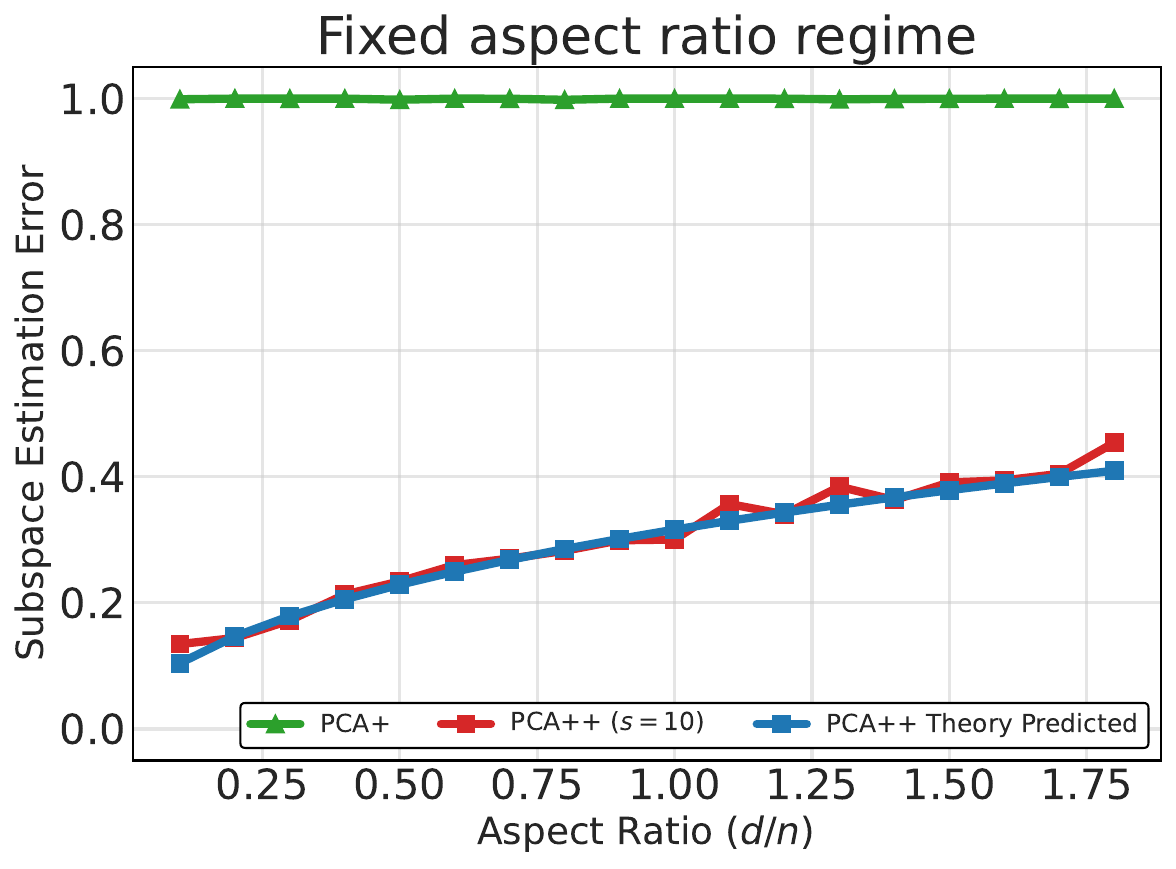}
}
\hspace{0.5cm}
\subfigure{
\includegraphics[width=.45\linewidth]{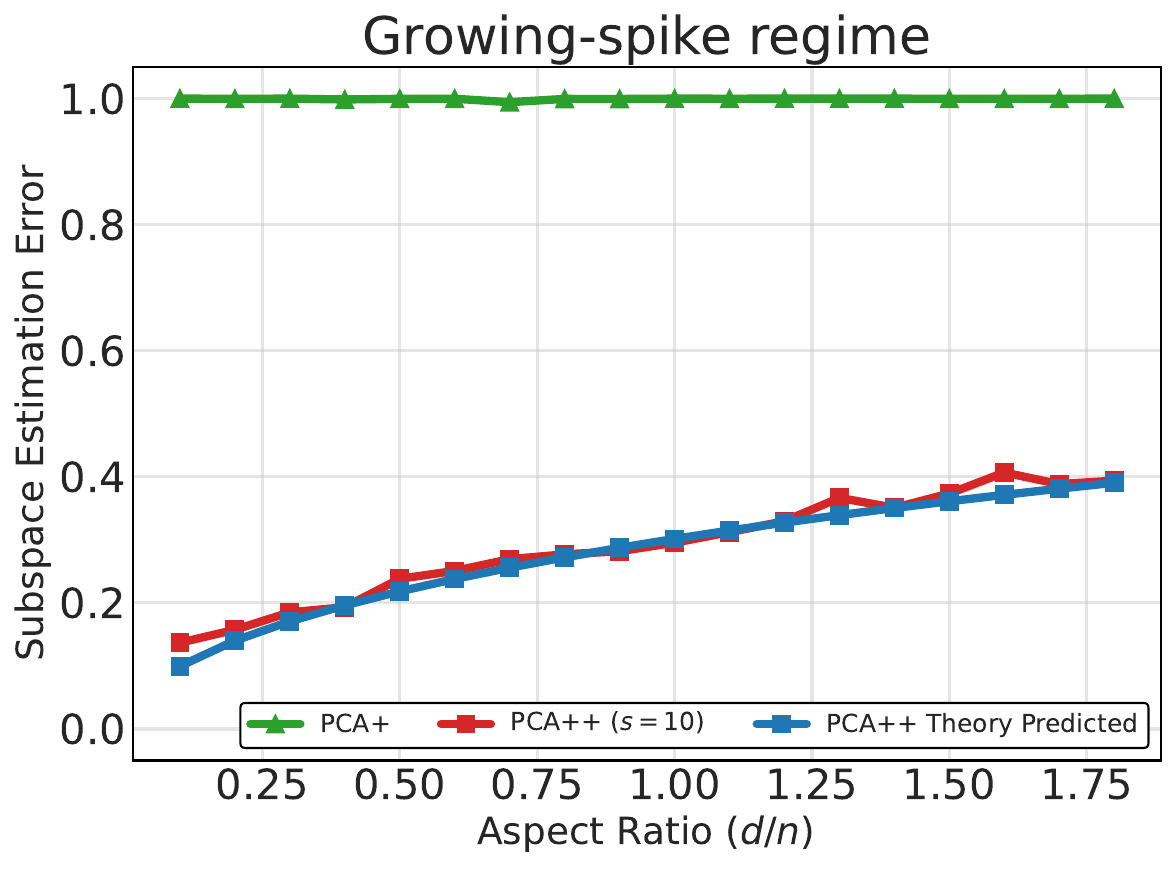}
}
    \caption{\small Empirical validation of theoretical predictions for $\pxx$. \textbf{Left:} Validation in the fixed aspect ratio regime for Theorem~\ref{thm:dist}. \textbf{Right:} Validation in the growing-spike regime for Theorem~\ref{thm:dist2}.}
    \label{fig:3}
\end{figure}

Under the fixed‐aspect‐ratio regime, the empirical $\pxx$ errors  trace our closed‐form curve almost exactly, while the alignment‐only $\px$ method diverges as $d/n$ increases (Figure~\ref{fig:3}, left).  In the growing‐spike regime—when both $d$ and spike variances are multiplied by ten—PCA++ continues to adhere to the simple limit $\frac{c}{1+c}$, but $\px$ collapses (Figure~\ref{fig:3}, right).  These results confirm that our theoretical predictions capture the precise high‐dimensional behavior of PCA++ in both regimes.
Further details and additional simulation results can be found in Appendix~\ref{app:simulation_details}.

% In the simplest setting, the signal and background each occupy a single direction in a \(d=800\) dimensional space, and we generated \(n=2000\) pairs.  The signal spike of magnitude \(\lambda_{A,1}=10\) lies along the first coordinate, while the background spike \(\lambda_{B,1}\) appears in the second coordinate.  By varying \(\lambda_{A,1}/\sqrt{\lambda_{B,1}}\) from 0.3125 to 0.666, we gradually strengthened the background relative to the signal.  As shown in Figure~\ref{fig:pcaplot_1} (left), both PCA and \(\px\) eventually lose alignment with the true signal direction, collapsing onto noise or drifting into the background axis.  In contrast, $\pxx$--even with a minimal truncation rank \(s=2\)--maintains near‐perfect alignment across the entire range of relative strengths.

% Next, we fixed \(n=500\) and varied the ambient dimension \(d\) so that the aspect ratio \(d/n\) ranges from 0.1 to 1.8, while holding \(\lambda_{A,1}=10\) and \(\lambda_{B,1}=500\).  Figure~\ref{fig:pcaplot_1} (right) shows that as \(d/n\) increases, both PCA and \(\px\) incur rapidly growing subspace error, whereas $\pxx$ remains stable and accurate even in the extreme high‐dimensional regime.

\begin{figure}[ht]
  \centering
  \subfigure{
    \includegraphics[width=0.3\textwidth]{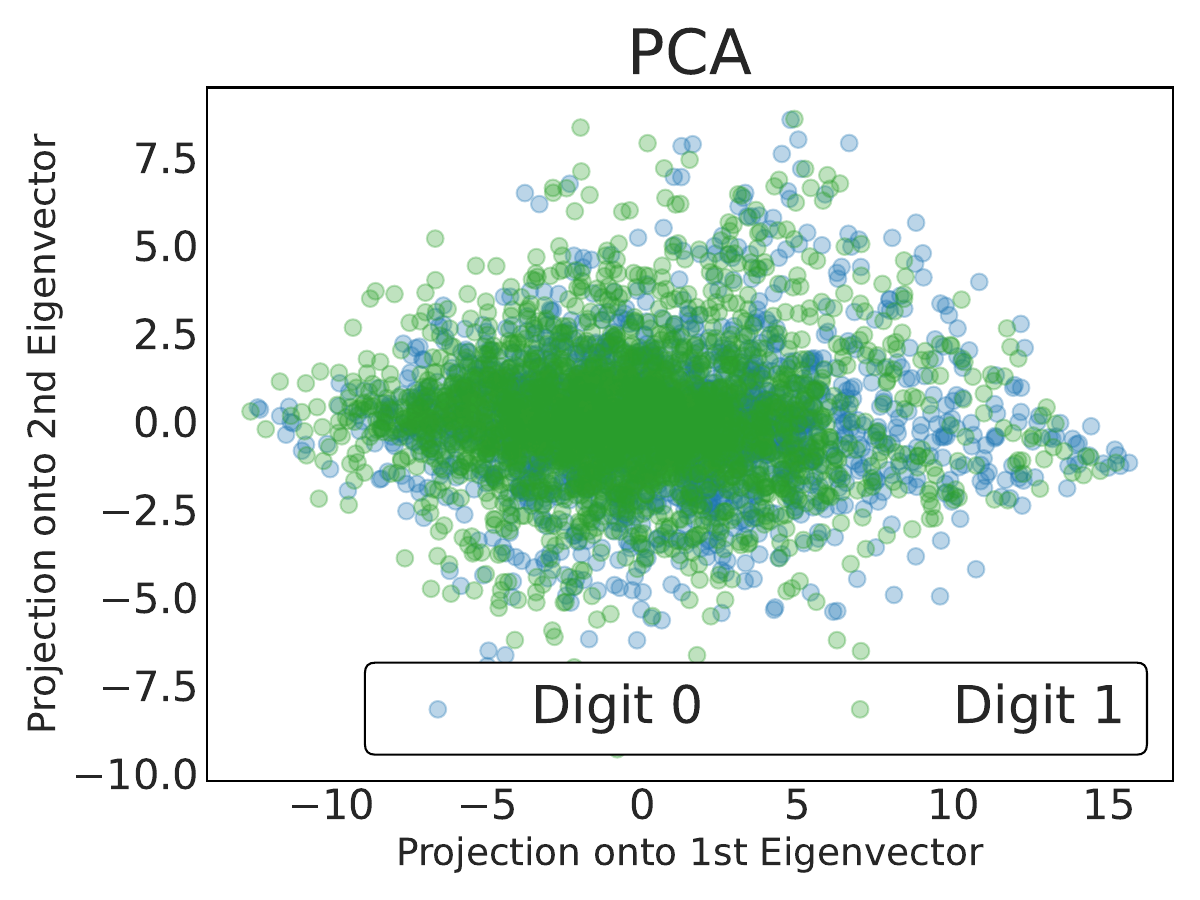}
    \label{fig:sub1}
  }
  \subfigure{
    \includegraphics[width=0.3\textwidth]{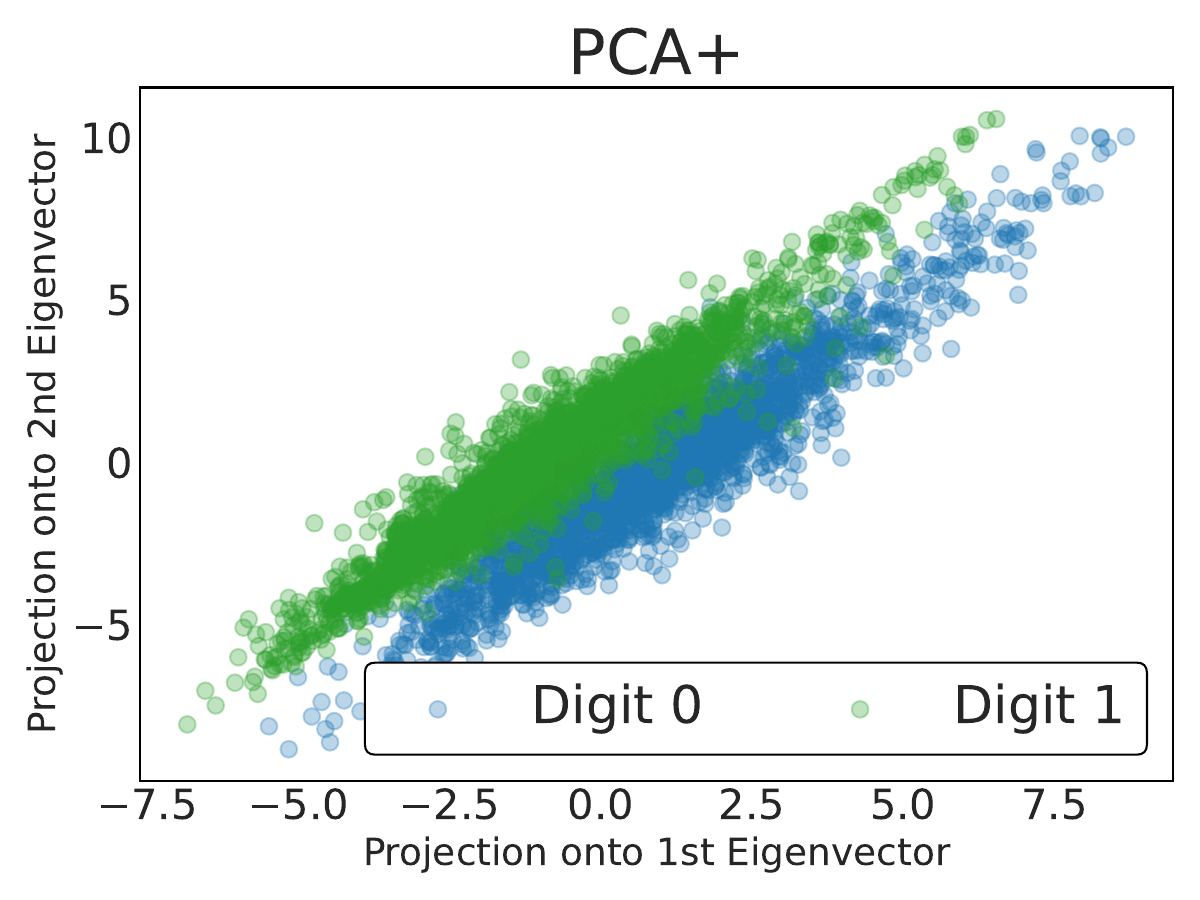}
    \label{fig:sub2}
  }
  \subfigure{
    \includegraphics[width=0.3\textwidth]{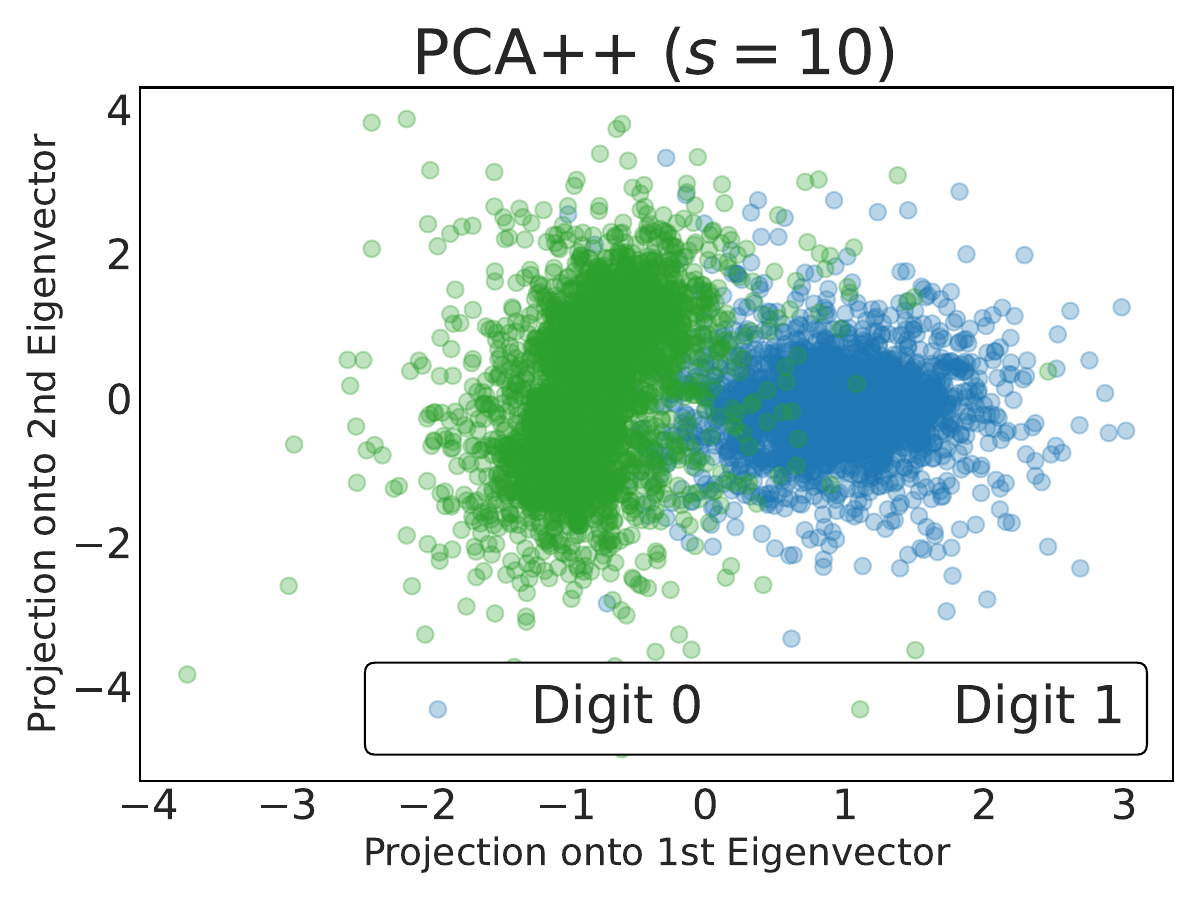}
    \label{fig:sub3}
  }
  \caption{\small 2D embeddings of noisy digit-over-grass images, standard PCA fails to separate classes. Contrastive $\px$ shows partial, misaligned separation. In contrast, our $\pxx$ achieves clear class separation predominantly along its first eigenvector, highlighting its superior ability to isolate the true signal and background noise.}
  \label{fig:MNIST}
\end{figure}

\paragraph{Corrupted MNIST Data.}
To visually assess signal disentanglement from structured background noise, we created a synthetic dataset of 5,000 paired images by superimposing MNIST digits \citep{deng2012mnist} (\texttt{'0'} or \texttt{'1'}) onto distinct ImageNet~\citep{deng2009imagenet} "grass" patches. Figure~\ref{fig:MNIST} compares the 2D embeddings obtained from standard PCA, $\px$, and $\pxx$. While standard PCA fails to separate classes and $\px$ shows only partial, misaligned separation, $\pxx$ clearly distinguishes the digits, with separation predominantly along its first learned eigenvector. This visually confirms that the uniformity constraint in $\pxx$ enables effective isolation of the true digit signal from the background.

\paragraph{Single-cell RNA sequencing data.}

\begin{figure}[ht]
\centering
\includegraphics[width=0.95\linewidth]{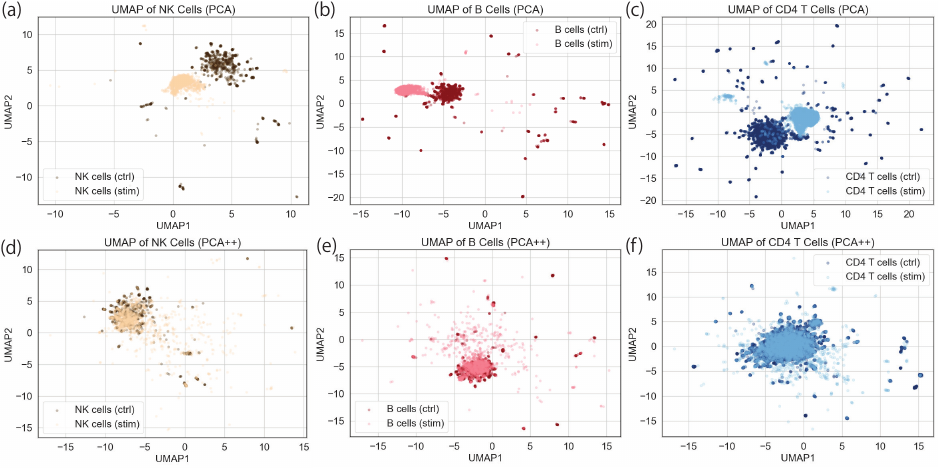}
\caption{\small \textbf{PCA vs.\ $\pxx$ embeddings.} We apply PCA and $\pxx$ to matched control and stimulated PBMCs (9,268 cells each) from the \cite{kang2018multiplexed} dataset and visualize the top 50 components using UMAP. (a–c) show PCA embeddings of CD4 T cells, B cells, and NK cells, where control and stimulated cells are often separated despite minimal biological response. (d–f) show the same cells under $\pxx$, where alignment across conditions improves, highlighting $\pxx$’s ability to isolate stable, condition-invariant structure.}
\label{fig:pca_vs_pxpp}
\end{figure}

We evaluated $\pxx$ on single-cell RNA-seq data from \cite{kang2018multiplexed}, comprising 14,619 control and 14,446 IFN-$\beta$–stimulated PBMCs across eight immune cell types. After matching 9,268 cells per condition via donor identity and local structural alignment, we extracted the top 50 components with both PCA and $\pxx$ and visualized each embedding (in Figure \ref{fig:pca_vs_pxpp}) using UMAP~\cite{mcinnes2018umap}. Standard PCA often separates control and stimulated cells of the same type, even when transcriptional changes are minimal, while $\pxx$ yields tightly overlapping clusters for invariant populations (e.g., CD4 T cells, B cells, NK cells), while still reflecting biologically meaningful dispersion in responsive cell types (e.g., monocytes in Figure \ref{fig:pcapp_umap} of the appendix). This demonstrates $\pxx$’s ability to isolate condition-invariant structure in complex single-cell data. Further details and additional simulation results can be found in Appendix \ref{app:real_data_details}.

\section{Discussion and Future Work}

{\color{black} This work revisits classical dimensionality reduction through a modern contrastive perspective, showing that alignment alone is insufficient for robust signal recovery in high-dimensional settings with strong structured noise. We demonstrate that incorporating an explicit uniformity constraint, a fundamental ingredient of recent contrastive learning theory, provides a principled safeguard against nuisance variation. The proposed method, $\pxx$, recovers latent signal subspaces in regimes where both standard PCA and alignment-only contrastive PCA ($\px$) provably fail.

\textbf{Uniformity as a robustness mechanism.} A key conceptual contribution of this work is clarifying the role of uniformity in contrastive learning. Prior studies emphasized its geometric effects (e.g., feature dispersion and isotropy), but its statistical role under structured noise was unclear. Our analysis shows that uniformity suppresses directions aligned with dominant background spikes, acting as a spectral filter that reweights principal directions based on their signal-to-background ratio rather than their raw variance.

%This connects contrastive learning with classical robust statistics, highlighting that structured interference cannot be eliminated by alignment alone and instead requires principled regularization.  

\textbf{Broader implications and future work.} Although studied in a linear setting, the insights extend beyond PCA: 
\textbf{(a)} \emph{Self-supervised learning.} $\pxx$ formalizes the benefit of uniformity in SimCLR-style losses and explains why explicit control of feature dispersion improves robustness to nuisance variation. \textbf{(b)} \emph{Multiview learning.} Our paired-data setting parallels CCA, but we show that variance-based alignment alone is insufficient under antagonistic background structure. The success of $\pxx$ suggests that modern representation learning in multiview settings may benefit from hard covariance constraints. Future extensions include contrastive sparse PCA \citep{zou2006sparse}, contrastive kernel PCA \citep{scholkopf1998nonlinear} and tensor PCA \citep{anandkumar2014tensor} for nonlinear/multiway data, and spectral clustering variants \citep{ng2001spectral} for community detection. These directions promise to expand uniformity-constrained inference across diverse high-dimensional settings.  }

\clearpage

\section*{Impact Statement}

This paper aims to advance the field of machine learning. While there may be societal impacts, none require specific attention here.

\section*{Acknowledgments}
We sincerely thank Yixuan Li for invaluable assistance with the real data analysis. This work was partially supported by NSERC Discovery Grants (RGPIN-2024-06780 and RGPIN-2018-06484) and FRQNT Team Research Project Grant (FRQNT 327788).

\bibliographystyle{apalike}
\bibliography{ref}

@inproceedings{NIPS2016_6b180037,
 author = {Sohn, Kihyuk},
 booktitle = {Advances in Neural Information Processing Systems},
 editor = {D. Lee and M. Sugiyama and U. Luxburg and I. Guyon and R. Garnett},
 pages = {},
 publisher = {Curran Associates, Inc.},
 title = {Improved Deep Metric Learning with Multi-class N-pair Loss Objective},
 url = {https://proceedings.neurips.cc/paper_files/paper/2016/file/6b180037abbebea991d8b1232f8a8ca9-Paper.pdf},
 volume = {29},
 year = {2016}
}

@INPROCEEDINGS{8578491,
  author={Wu, Zhirong and Xiong, Yuanjun and Yu, Stella X. and Lin, Dahua},
  booktitle={2018 IEEE/CVF Conference on Computer Vision and Pattern Recognition}, 
  title={Unsupervised Feature Learning via Non-parametric Instance Discrimination}, 
  year={2018},
  volume={},
  number={},
  pages={3733-3742},
  doi={10.1109/CVPR.2018.00393}}

@article{hubert1985comparing,
  title={Comparing partitions},
  author={Hubert, Lawrence and Arabie, Phipps},
  journal={Journal of Classification},
  volume={2},
  number={1},
  pages={193--218},
  year={1985},
  publisher={Springer},
  doi={10.1007/BF01908075}
}

@article{mcinnes2018umap,
  title={{UMAP}: Uniform manifold approximation and projection for dimension reduction},
  author={McInnes, Leland and Healy, John and Melville, James},
  journal={arXiv preprint arXiv:1802.03426},
  year={2018}
}

@article{van2008visualizing,
  title={Visualizing Data using t-{SNE}},
  author={van der Maaten, Laurens and Hinton, Geoffrey},
  journal={Journal of Machine Learning Research},
  volume={9},
  pages={2579--2605},
  year={2008}
}

@article{cca,
    author = {Hotelling, Harold},
    title = {RELATIONS BETWEEN TWO SETS OF VARIATES},
    journal = {Biometrika},
    volume = {28},
    number = {3-4},
    pages = {321-377},
    year = {1936},
    month = {12},
    issn = {0006-3444},
    doi = {10.1093/biomet/28.3-4.321},
    url = {https://doi.org/10.1093/biomet/28.3-4.321},
    eprint = {https://academic.oup.com/biomet/article-pdf/28/3-4/321/586830/28-3-4-321.pdf},
}

@article{SALLOUM2022108378,
title = {c{PCA}++: An efficient method for contrastive feature learning},
journal = {Pattern Recognition},
volume = {124},
pages = {108378},
year = {2022},
issn = {0031-3203},
doi = {https://doi.org/10.1016/j.patcog.2021.108378},
url = {https://www.sciencedirect.com/science/article/pii/S0031320321005586},
author = {Ronald Salloum and C.-C. Jay Kuo}
}

@article{fama1993common,
  title={Common risk factors in the returns on stocks and bonds},
  author={Fama, Eugene F and French, Kenneth R},
  journal={Journal of Financial Economics},
  volume={33},
  number={1},
  pages={3--56},
  year={1993},
  publisher={Elsevier}
}

@article{campbell2001have,
  title={Have individual stocks become more volatile? An empirical exploration of idiosyncratic risk},
  author={Campbell, John Y and Lettau, Martin and Malkiel, Burton G and Xu, Yexiao},
  journal={The Journal of Finance},
  volume={56},
  number={1},
  pages={1--43},
  year={2001},
  publisher={Wiley Online Library}
}

@article{fortin2017harmonization,
  title={Harmonization of multi-site diffusion tensor imaging data},
  author={Fortin, Jean-Philippe and Parker, Drew and Tun{\c{c}}, Birkan and Watanabe, Takanori and Elliott, Mark A and Ruparel, Kosha and Roalf, David R and Satterthwaite, Theodore D and Gur, Ruben C and Gur, Raquel E and others},
  journal={Neuroimage},
  volume={161},
  pages={149--170},
  year={2017},
  publisher={Elsevier}
}

@article{haghverdi2018batch,
  title={Batch effects in single-cell {RNA}-sequencing data are corrected by matching mutual nearest neighbors},
  author={Haghverdi, Laleh and Lun, Aaron TL and Morgan, Michael D and Marioni, John C},
  journal={Nature Biotechnology},
  volume={36},
  number={5},
  pages={421--427},
  year={2018},
  publisher={Nature Publishing Group}
}

@article{leek2010tackling,
  title={Tackling the widespread and critical impact of batch effects in high-throughput data},
  author={Leek, Jeffrey T and Scharpf, Robert B and Bravo, H{\'e}ctor Corrada and Simcha, David and Langmead, Benjamin and Johnson, W Evan and Geman, Donald and Baggerly, Keith and Irizarry, Rafael A},
  journal={Nature Reviews Genetics},
  volume={11},
  number={10},
  pages={733--739},
  year={2010},
  publisher={Nature Publishing Group UK London}
}

@inproceedings{chen2021exploring,
  title={Exploring simple siamese representation learning},
  author={Chen, Xinlei and He, Kaiming},
  booktitle={Proceedings of the IEEE/CVF Conference on Computer Vision and Pattern Recognition},
  pages={15750--15758},
  year={2021}
}

@inproceedings{gutmann2010noise,
  title={Noise-contrastive estimation: A new estimation principle for unnormalized statistical models},
  author={Gutmann, Michael and Hyv{\"a}rinen, Aapo},
  booktitle={Proceedings of The  International Conference on Artificial Intelligence and Statistics},
  pages={297--304},
  year={2010},
  organization={JMLR Workshop and Conference Proceedings}
}

@article{haochen2022beyond,
  title={Beyond separability: Analyzing the linear transferability of contrastive representations to related subpopulations},
  author={HaoChen, Jeff Z and Wei, Colin and Kumar, Ananya and Ma, Tengyu},
  journal={Advances in Neural Information Processing Systems},
  volume={35},
  pages={26889--26902},
  year={2022}
}

@article{baik2006eigenvalues,
  title={Eigenvalues of large sample covariance matrices of spiked population models},
  author={Baik, Jinho and Silverstein, Jack W},
  journal={Journal of Multivariate Analysis},
  volume={97},
  number={6},
  pages={1382--1408},
  year={2006},
  publisher={Elsevier}
}

@article{paul2007asymptotics,
  title={Asymptotics of sample eigenstructure for a large dimensional spiked covariance model},
  author={Paul, Debashis},
  journal={Statistica Sinica},
  pages={1617--1642},
  year={2007},
  publisher={JSTOR}
}

@article{Bai1993,
author = {Z. D. Bai and Y. Q. Yin},
title = {{Limit of the smallest eigenvalue of a large dimensional sample covariance matrix}},
volume = {21},
journal = {The Annals of Probability},
number = {3},
publisher = {Institute of Mathematical Statistics},
pages = {1275 -- 1294},
year = {1993}
}

@article{chen2021spectral,
  title={Spectral methods for data science: A statistical perspective},
  author={Chen, Yuxin and Chi, Yuejie and Fan, Jianqing and Ma, Cong and others},
  journal={Foundations and Trends in Machine Learning},
  volume={14},
  number={5},
  pages={566--806},
  year={2021},
  publisher={Now Publishers, Inc.}
}

@article{shen2016statistics,
  title={The statistics and mathematics of high dimension low sample size asymptotics},
  author={Shen, Dan and Shen, Haipeng and Zhu, Hongtu and Marron, JS},
  journal={Statistica Sinica},
  volume={26},
  number={4},
  pages={1747},
  year={2016},
  publisher={NIH Public Access}
}

@article{ghojogh2019eigenvalue,
  title={Eigenvalue and generalized eigenvalue problems: Tutorial},
  author={Ghojogh, Benyamin and Karray, Fakhri and Crowley, Mark},
  journal={arXiv preprint arXiv:1903.11240},
  year={2019}
}

@book{parlett1998symmetric,
  title={The symmetric eigenvalue problem},
  author={Parlett, Beresford N},
  year={1998},
  publisher={SIAM}
}

@inproceedings{chen2020simple,
  title={A simple framework for contrastive learning of visual representations},
  author={Chen, Ting and Kornblith, Simon and Norouzi, Mohammad and Hinton, Geoffrey},
  booktitle={International Conference on Machine Learning},
  pages={1597--1607},
  year={2020},
  organization={PMLR}
}

@inproceedings{he2020momentum,
  title={Momentum contrast for unsupervised visual representation learning},
  author={He, Kaiming and Fan, Haoqi and Wu, Yuxin and Xie, Saining and Girshick, Ross},
  booktitle={Proceedings of the IEEE/CVF Conference on Computer Vision and Pattern Recognition},
  pages={9729--9738},
  year={2020}
}

@inproceedings{wang2020understanding,
  title={Understanding contrastive representation learning through alignment and uniformity on the hypersphere},
  author={Wang, Tongzhou and Isola, Phillip},
  booktitle={International Conference on Machine Learning},
  pages={9929--9939},
  year={2020},
  organization={PMLR}
}

@article{bardes2021vicreg,
  title={{VICR}eg: Variance-invariance-covariance regularization for self-supervised learning},
  author={Bardes, Adrien and Ponce, Jean and LeCun, Yann},
  journal={arXiv preprint arXiv:2105.04906},
  year={2021}
}

@inproceedings{zbontar2021barlow,
  title={{B}arlow {T}wins: Self-supervised learning via redundancy reduction},
  author={Zbontar, Jure and Jing, Li and Misra, Ishan and LeCun, Yann and Deny, St{\'e}phane},
  booktitle={International Conference on Machine Learning},
  pages={12310--12320},
  year={2021},
  organization={PMLR}
}

@InProceedings{pmlr-v195-parulekar23a,
  title = 	 {InfoNCE Loss Provably Learns Cluster-Preserving Representations},
  author =       {Parulekar, Advait and Collins, Liam and Shanmugam, Karthikeyan and Mokhtari, Aryan and Shakkottai, Sanjay},
  booktitle = 	 {Proceedings of Thirty Sixth Conference on Learning Theory},
  pages = 	 {1914--1961},
  year = 	 {2023},
  editor = 	 {Neu, Gergely and Rosasco, Lorenzo},
  volume = 	 {195},
  series = 	 {Proceedings of Machine Learning Research},
  month = 	 {12--15 Jul},
  publisher =    {PMLR},
  url = 	 {https://proceedings.mlr.press/v195/parulekar23a.html}
}

@inproceedings{saunshi2022understanding,
  title={Understanding contrastive learning requires incorporating inductive biases},
  author={Saunshi, Nikunj and Ash, Jordan and Goel, Surbhi and Misra, Dipendra and Zhang, Cyril and Arora, Sanjeev and Kakade, Sham and Krishnamurthy, Akshay},
  booktitle={International Conference on Machine Learning},
  pages={19250--19286},
  year={2022},
  organization={PMLR}
}

@article{haochen2022theoretical,
  title={A theoretical study of inductive biases in contrastive learning},
  author={HaoChen, Jeff Z and Ma, Tengyu},
  journal={arXiv preprint arXiv:2211.14699},
  year={2022}
}

@article{oord2018representation,
  title={Representation learning with contrastive predictive coding},
  author={Oord, Aaron van den and Li, Yazhe and Vinyals, Oriol},
  journal={arXiv preprint arXiv:1807.03748},
  year={2018}
}

@article{bachman2019learning,
  title={Learning representations by maximizing mutual information across views},
  author={Bachman, Philip and Hjelm, R Devon and Buchwalter, William},
  journal={Advances in Neural Information Processing Systems},
  volume={32},
  year={2019}
}

@inproceedings{saunshi2019theoretical,
  title={A theoretical analysis of contrastive unsupervised representation learning},
  author={Saunshi, Nikunj and Plevrakis, Orestis and Arora, Sanjeev and Khodak, Mikhail and Khandeparkar, Hrishikesh},
  booktitle={International Conference on Machine Learning},
  pages={5628--5637},
  year={2019},
  organization={PMLR}
}

@inproceedings{mcallester2020formal,
  title={Formal limitations on the measurement of mutual information},
  author={McAllester, David and Stratos, Karl},
  booktitle={International Conference on Artificial Intelligence and Statistics},
  pages={875--884},
  year={2020},
  organization={PMLR}
}

@article{tschannen2019mutual,
  title={On mutual information maximization for representation learning},
  author={Tschannen, Michael and Djolonga, Josip and Rubenstein, Paul K and Gelly, Sylvain and Lucic, Mario},
  journal={arXiv preprint arXiv:1907.13625},
  year={2019}
}

@article{chen2021intriguing,
  title={Intriguing properties of contrastive losses},
  author={Chen, Ting and Luo, Calvin and Li, Lala},
  journal={Advances in Neural Information Processing Systems},
  volume={34},
  pages={11834--11845},
  year={2021}
}

@article{haochen2021provable,
  title={Provable guarantees for self-supervised deep learning with spectral contrastive loss},
  author={HaoChen, Jeff Z and Wei, Colin and Gaidon, Adrien and Ma, Tengyu},
  journal={Advances in Neural Information Processing Systems},
  volume={34},
  pages={5000--5011},
  year={2021}
}

@inproceedings{Daniel2023,
  author       = {Daniel D. Johnson and
                  Ayoub El Hanchi and
                  Chris J. Maddison},
  title        = {Contrastive Learning Can Find An Optimal Basis For Approximately View-Invariant
                  Functions},
  booktitle    = {The International Conference on Learning Representations,
                  {ICLR} 2023, Kigali, Rwanda, May 1-5, 2023},
  publisher    = {OpenReview.net},
  year         = {2023},
  url          = {https://openreview.net/forum?id=AjC0KBjiMu},
  bibsource    = {dblp computer science bibliography, https://dblp.org}
}

@inproceedings{tan2024contrastive,
  title={Contrastive Learning is Spectral Clustering on Similarity Graph},
  author={Tan, Zhiquan and Zhang, Yifan and Yang, Jingqin and Yuan, Yang},
  booktitle={International Conference
on Learning Representations},
  year={2024}
}

@article{ji2023power,
  title={The power of contrast for feature learning: A theoretical analysis},
  author={Ji, Wenlong and Deng, Zhun and Nakada, Ryumei and Zou, James and Zhang, Linjun},
  journal={Journal of Machine Learning Research},
  volume={24},
  number={330},
  pages={1--78},
  year={2023}
}

@article{bansal2024understanding,
  title={Understanding Contrastive Learning via {G}aussian Mixture Models},
  author={Bansal, Parikshit and Kavis, Ali and Sanghavi, Sujay},
  journal={arXiv preprint arXiv:2411.03517},
  year={2024}
}

@article{Johnstone2001,
 ISSN = {00905364, 21688966},
 URL = {http://www.jstor.org/stable/2674106},
 author = {Iain M. Johnstone},
 journal = {The Annals of Statistics},
 number = {2},
 pages = {295--327},
 publisher = {Institute of Mathematical Statistics},
 title = {On the Distribution of the Largest Eigenvalue in Principal Components Analysis},
 urldate = {2025-05-11},
 volume = {29},
 year = {2001}
}

@article{kang2018multiplexed,
  title={Multiplexed droplet single-cell {RNA}-sequencing using natural genetic variation},
  author={Kang, Hyun Min and Subramaniam, Meena and Targ, Sasha and Nguyen, Michelle and Maliskova, Lenka and McCarthy, Elizabeth and Wan, Eunice and Wong, Simon and Byrnes, Lauren and Lanata, Cristina M and others},
  journal={Nature Biotechnology},
  volume={36},
  number={1},
  pages={89--94},
  year={2018},
  publisher={Nature Publishing Group US New York}
}

@article{deng2012mnist,
  title={The mnist database of handwritten digit images for machine learning research},
  author={Deng, Li},
  journal={IEEE Signal Processing Magazine},
  volume={29},
  number={6},
  pages={141--142},
  year={2012},
  publisher={IEEE}
}

@inproceedings{deng2009imagenet,
  title={Image{N}et: A large-scale hierarchical image database},
  author={Deng, Jia and Dong, Wei and Socher, Richard and Li, Li-Jia and Li, Kai and Fei-Fei, Li},
  booktitle={2009 IEEE Conference on Computer Vision and Pattern Recognition},
  pages={248--255},
  year={2009},
  organization={IEEE}
}

@article{zou2006sparse,
  title={Sparse principal component analysis},
  author={Zou, Hui and Hastie, Trevor and Tibshirani, Robert},
  journal={Journal of Computational and Graphical Statistics},
  volume={15},
  number={2},
  pages={265--286},
  year={2006},
  publisher={Taylor \& Francis}
}

@article{scholkopf1998nonlinear,
  title={Nonlinear component analysis as a kernel eigenvalue problem},
  author={Sch{\"o}lkopf, Bernhard and Smola, Alexander and M{\"u}ller, Klaus-Robert},
  journal={Neural Computation},
  volume={10},
  number={5},
  pages={1299--1319},
  year={1998},
  publisher={MIT Press}
}

@article{anandkumar2014tensor,
  title={Tensor decompositions for learning latent variable models.},
  author={Anandkumar, Animashree and Ge, Rong and Hsu, Daniel J and Kakade, Sham M and Telgarsky, Matus and others},
  journal={Journal of Machine Learning Research},
  volume={15},
  number={1},
  pages={2773--2832},
  year={2014}
}

@article{ng2001spectral,
  title={On spectral clustering: Analysis and an algorithm},
  author={Ng, Andrew and Jordan, Michael and Weiss, Yair},
  journal={Advances in Neural Information Processing Systems},
  volume={14},
  year={2001}
}

@article{candes2011robust,
  title={Robust principal component analysis?},
  author={Cand{\`e}s, Emmanuel J and Li, Xiaodong and Ma, Yi and Wright, John},
  journal={Journal of the ACM},
  volume={58},
  number={3},
  pages={1--37},
  year={2011},
  publisher={ACM New York, NY, USA}
}

@article{abid2018exploring,
  title={Exploring patterns enriched in a dataset with contrastive principal component analysis},
  author={Abid, Abubakar and Zhang, Martin J and Bagaria, Vivek K and Zou, James},
  journal={Nature Communications},
  volume={9},
  number={1},
  pages={2134},
  year={2018},
  publisher={Nature Publishing Group UK London}
}

@article{li2020probabilistic,
  title={Probabilistic contrastive principal component analysis},
  author={Li, Didong and Jones, Andrew and Engelhardt, Barbara},
  journal={arXiv preprint arXiv:2012.07977},
  year={2020}
}

@article{boileau2020exploring,
  title={Exploring high-dimensional biological data with sparse contrastive principal component analysis},
  author={Boileau, Philippe and Hejazi, Nima S and Dudoit, Sandrine},
  journal={Bioinformatics},
  volume={36},
  number={11},
  pages={3422--3430},
  year={2020},
  publisher={Oxford University Press}
}

@article{zhou2023sparse,
  title={Sparse discriminant PCA based on contrastive learning and class-specificity distribution},
  author={Zhou, Qian and Gao, Quanxue and Wang, Qianqian and Yang, Ming and Gao, Xinbo},
  journal={Neural Networks},
  volume={167},
  pages={775--786},
  year={2023},
  publisher={Elsevier}
}

@inproceedings{zhang2025contrastive,
  title={Contrastive Functional Principal Component Analysis},
  author={Zhang, Eric and Li, Didong},
  booktitle={Proceedings of the AAAI Conference on Artificial Intelligence},
  volume={39},
  pages={22380--22388},
  year={2025}
}

%%%%%%%%%%%%%%%%%%%%%%%%%%%%%%%%%%%%%%%%%%%%%%%%%%%%%%%%%%%%
\clearpage

\appendix
\begin{center}
{\bf \Large Appendix} \\ \vspace{0.6cm} {\Large $\pxx$: How Uniformity Induces Robustness to
Background Noise in Contrastive Learning}
\end{center}

\section{Principal Angles}\label{sec:principleangles}

Denote the true subspace by \(\eigspace_{\sig} \coloneqq \spann\{\eigvec_{\sig,1}, \ldots, \eigvec_{\sig,\nsig}\}\) and let  \(\widehat{\eigspace}_{\sig}\) be an estimator obtained from a given learning method. We measure the estimation error between \(\eigspace_{\sig}\) and \(\widehat{\eigspace}_{\sig}\) via their principal angles, which capture the maximal deviations between corresponding directions in the two subspaces:
\begin{definition}[Principal angles]\label{def:principal_angles}
Let \(\eigspace\) and \(\eigspace'\) be \(\nsig\)-dimensional subspaces of \(\RR^{\ndim}\). Let the columns of  $U\in\RR^{\ndim\times \nsig}$ and $U'\in\RR^{\ndim\times \nsig}$ form orthonormal bases for \(\eigspace\) and \(\eigspace'\), respectively. Write the singular values of \(U^\top U'\) in descending order as  \(\sigma_1 \geq \sigma_2 \geq \cdots \geq \sigma_{\nsig} \geq 0\). The principal angles \(\theta_j\) between \(\eigspace\) and \(\eigspace'\) are defined by
\[
\theta_j \coloneqq \arccos(\sigma_j), \quad \text{for } j = 1, \ldots, \nsig.
\]
\end{definition}
These angles capture the worst-case misalignment between the two subspaces.  A common aggregate distance is
\[
\operatorname{dist}_{\sin,\|\cdot\|}(\eigspace, \eigspace') \coloneqq \| \sin \Theta \| \coloneqq \| \diag(\sin \theta_1, \ldots, \sin \theta_{\nsig}) \|,
\]
where $\|\cdot\|$ is any unitarily invariant matrix norm (e.g., the operator or Frobenius norm).
\begin{remark}
This metric is invariant under orthogonal transformations and relates directly to other subspace distances commonly employed in the literature, such as the projection Frobenius norm \(\|UU^\top - U' U'^{\top}\|_F\). Thus, it offers an interpretable measure to evaluate subspace estimation accuracy, particularly useful in high-dimensional analysis.
\end{remark}

\section{Generalized Eigenvalue Problem}
\subsection[Equivalence between constrained and generalized eigenvalue formulations]{Equivalence between \eqref{eq:constrained_opt} and \eqref{eq:geneigen}}\label{sec:GEV_proof}

Use variable $V'$ instead of $V$ in \eqref{eq:constrained_opt} (the reason will become clear later), and introduce a symmetric multiplier~$M \in\mathbb{R}^{k\times k}$ for \eqref{eq:constrained_opt}.
The Lagrangian is 
\[
L(V',M)
=
\tr\bigl(V'^{\top}S_{n}^{+}V'\bigr)
-
\tr\bigl[M(V'^{\top}S_{n}V' - I_{k})\bigr].
\]
Since $M$ is symmetric, we may rewrite
\[
L(V',M)
=
\tr(V'^{\top}S_{n}^{+}V')
-
\tr(M V'^{\top}S_{n}V')
+
\tr(M).
\]
We differentiate $L(V',M)$ w.r.t.\ $V'$ and then set $\nabla_{V'}L=\mathbf{0}$, we obtain
\begin{equation}\label{eq:Lequation}  
    \nabla_{V'}L = 2 S_{n}^{+} V' - 2 S_{n} V' M = \mathbf{0} \qquad \Longrightarrow \qquad S_{n}^{+}V' \;=\; S_{n}V'M.  
\end{equation}
Meanwhile, multiply $V'^{\top}$ to both sides of  \eqref{eq:Lequation}, we obtain 
\[
V'^{\top}S_{n}^{+}V' \;=\; V'^{\top}S_{n}V'M \qquad \Longrightarrow \qquad     M = V'^{\top}S_{n}^{+}V' 
\]
where we used $V'^{\top} S_{n} V' = I_{k}$.  The objective function we want to maximize is $\tr(V'^{\top}S_{n}^{+}V')$, which is equal to $\tr(M)$. The matrix $M = V'^{\top}S_n^{+}V'$ is a $k \times k$ real symmetric matrix. Therefore, it is orthogonally diagonalizable. Let $Q$ be a $k \times k$ orthogonal matrix ($Q^{\top}Q = QQ^{\top} = I_k$) such that $Q^{\top}M Q = \Lambda$, where $\Lambda = \text{diag}(\lambda_1, \lambda_2, \dots, \lambda_k)$ is a diagonal matrix containing the eigenvalues of $M$.

Define a new matrix $V = V'Q$. 
We check if $V$ satisfies the constraint:
\[
V^{\top}S_n V = (V'Q)^{\top}S_n (V'Q) = Q^{\top}V'^{\top}S_n V' Q = Q^{\top}I_k Q = Q^{\top}Q = I_k
\]
So $V$ also satisfies the constraint.
Now substitute $V' = VQ^{\top}$ into equation \eqref{eq:Lequation}:
\[
S_n^{+}(VQ^{\top}) = S_n (VQ^{\top}) M
\]
Post-multiply by $Q$:
\[
S_n^{+}VQ^{\top}Q = S_n VQ^{\top}M Q.
\]
We get
\begin{equation} \label{eq:diag_geneigen}
S_n^{+}V = S_n V \Lambda
\end{equation}
If we write $V = (v_1, v_2, \dots, v_k)$, then equation \eqref{eq:diag_geneigen} can be expressed column by column as:
\[
S_n^{+}v_j = \lambda_j S_n v_j \quad \text{for } j=1, \dots, k.
\]
This is precisely the generalized eigenvalue equation. Since the objective function is 
\[
\tr(V'^{\top}S_{n}^{+}V')= \tr(M)= \sum_{j=1}^k \lambda_j
\]
To maximize this sum, we must choose the $k$ generalized eigenvectors $v_j$ that correspond to the $k$ largest generalized eigenvalues $\lambda_j$.  The solution matrix $V$  will have these $k$ eigenvectors as its columns. Typically, the full generalized eigenvalue problem $S_n^{+}v_j = \lambda_j S_n v_j$ is solved for all $d$ possible eigenvectors and eigenvalues. Then, these eigenvalues are sorted in descending order, $\lambda_1\geq \lambda_2\geq \cdots \geq \lambda_d$, and the eigenvectors $v_1,\ldots,v_k$ corresponding to the $k$ largest eigenvalues are chosen to form the columns of $V=(v_1,\ldots,v_k)$.

\subsection{Generalized eigenvalue solver}\label{sec:Generalized_eigenvalue}
To compute the hard-uniformity PCA ($\pxx$) in \eqref{eq:constrained_opt}, we must solve the generalized eigenvalue problem
\begin{equation*}
    \covmatsample_{\ndata}^+ \eigvec_j=\eigval_j S_n \eigvec_j, \quad \forall j \in\{1, \ldots, \ndim\}.
\end{equation*}
where $\covmatsample_{\ndata}^+ = \frac{1}{2n} (X^{\top}X^{+} + X^{+\top}X)$ and $S =  \frac{1}{\ndata} X^\top X$. Equivalently, in matrix form,
\begin{equation*}
    \covmatsample_{\ndata}^+ V=S V \Lambda,
\end{equation*}
where $V = (\eigvec_1, \ldots, \eigvec_d)$ contains the eigenvectors as columns, $\Lambda = \diag(\eigval_1, \ldots, \eigval_d)$ is a diagonal matrix of eigenvalues.  

A stable and efficient procedure is as follows:
\begin{algorithm}
\caption{Generalized eigenvalue solver}
\label{alg:generalized_eig}
\begin{enumerate}
  \item Compute eigendecomposition: \( S_{n} V_{x} = V_{x} \Lambda_{x} \).
  \item Compute \( R \gets V_{x} (\Lambda_{x} + \varepsilon I_{d})^{-1/2} \)  where \(\varepsilon\) is a small regularizer to handle nearly zero eigenvalues.
  \item Project the contrastive covariance: \( M \gets R^{\top} \covmatsample_{\ndata}^+ R \).
  \item Compute eigendecomposition: \( M U  = U \Lambda \). 
  \item Recover generalized eigenvectors \( V \gets V_{x} U \) and generalized eigenvalues $\Lambda$.
  \end{enumerate}
\end{algorithm}

{ \color{black}
\section{On Selecting Hyperparameter \texorpdfstring{$s$}{s}\label{app:hyperparam}}
The choice of $s$ in truncated $\pxx$ controls a fundamental trade-off between the stability of the generalized eigenproblem and the effectiveness of the uniformity constraint:
\begin{itemize}
    \item A \textbf{small $s$} ensures that the inverse of the truncated covariance $(S_n)_s$ is well-conditioned, promoting numerical stability. However, it may discard dimensions needed to enforce uniformity effectively, potentially biasing the result.
    \item A \textbf{large $s$} enforces uniformity over a larger subspace but risks instability if $S_n$ is ill-conditioned, as including directions with near-zero eigenvalues can amplify noise. As shown in our experiments (Figure~\ref{fig:2}, left).
\end{itemize}

Based on this trade-off, we propose to \textbf{combine the following practical strategies} for selecting $s$:
\begin{itemize}
    \item \textbf{Information-based criterion (lower bound):} As in standard PCA, one can determine a minimum $s$ by examining the cumulative variance explained by the eigenvalues of the sample covariance $S_n$. For example, choose $s$ large enough to capture a significant portion (e.g., 90\%) of the total variance. This ensures the uniformity constraint operates on the most meaningful directions.
    \item \textbf{Stability-based criterion (upper bound):} To ensure numerical stability, one can monitor the \textbf{condition number} of the truncated matrix $(S_n)_s$, which is the ratio of its largest to its $s$-th eigenvalue $(\lambda_1(S_n) / \lambda_s(S_n))$. One should choose $s$ such that this ratio remains below a reasonable threshold, avoiding severe ill-conditioning.
\end{itemize}

In our experiments (e.g., Figure~\ref{fig:2}, right), we found that $\pxx$ is robust across a reasonable range of $s$.

\section{Computational Complexity}\label{app:complex}
The computational cost of $\pxx$ has two main components:
\begin{enumerate}
    \item \textbf{Covariance matrix formation:} We compute the sample covariance $S_n$ and the contrastive covariance $S_{n}^{+}$. For data matrices of size $n \times d$ (samples × features), forming these $d \times d$ matrices requires matrix multiplications with a complexity of $O(nd^{2})$. In the common high-dimensional setting where
$d\gg n$, this can be optimized to
$O(dn^{2})$ by first computing the $n\times n$ Gram matrix.
    \item \textbf{Generalized eigenvalue problem (GEP) solution:} The cost is dominated by the initial truncated eigendecomposition of  $S_n$ to rank $s$. Using an iterative solver like the IRLM, as implemented in \texttt{scipy}, this step has a complexity of approximately $O(sd^{2})$. Since the truncation rank $s$ is typically much smaller than $d$, this step is highly efficient.
\end{enumerate}

Overall, the complexity is comparable to standard PCA, primarily driven by the feature dimension $d$.

\paragraph{Empirical Scalability}

To demonstrate its practical performance, we benchmarked the runtime of $\pxx$ while varying the number of samples ($n$) and features ($d$). The results confirm our theoretical analysis.

\begin{table}[ht]
  \centering
  \caption{Computational cost of $\pxx$ (in seconds)}
  \label{tab:execution_time_simple}
  \begin{tabular}{l rrrr}
    \toprule
    $n$\textbackslash $d$ & \textbf{100} & \textbf{1000} & \textbf{5000} & \textbf{10000} \\
    \midrule
    \textbf{100} & 0.002 & 0.157 & 2.288 & 9.492 \\
    \textbf{1000} & 0.003 & 0.204 & 3.781 & 15.666 \\
    \textbf{5000} & 0.009 & 0.662 & 13.919 & 53.610 \\
    \textbf{10000} & 0.017 & 1.310 & 26.433 & 100.656 \\
    \bottomrule
  \end{tabular}
\end{table}

Setup: Truncation rank $s=10$, top $k=5$ eigenvectors estimated. Benchmarked on an Intel Xeon CPU @ 2.20GHz.

The benchmarks show that the runtime is dominated by the feature dimension $d$, scaling quadratically, while it scales nearly linearly with the number of samples $n$. This profile makes $\pxx$ computationally feasible for typical high-dimensional datasets with tens of thousands of features, confirming its practical scalability.

\section{Discussion of Theoretical Assumptions}\label{sec:Assumption}

\subsection{On Orthogonal signal and background (Assumption~\ref{asm:orthogonal})}
\textit{Why the core mechanism is robust to overlapping subspaces:} The key insight, enabled by the \textit{linearity of our factor model in~\eqref{eq:model}}, is that we can analyze the behavior of the covariance matrices in a shared basis. Even with overlap, the contrastive covariance $S_{n}^{+}$ isolates signal components, while the standard covariance $S_n$ accumulates variance from both signal and background in the shared directions.

Let's sketch this for a simple case. Because the model is linear, we can define an orthonormal basis that accounts for the overlap. Suppose $\text{span}(A)$ and $\text{span}(B)$ share a single direction $v_0$. We can decompose the subspaces as:

\begin{itemize}
    \item \textbf{Signal space:} $\text{span(A)}$ is spanned by $ \{ v_0, v_{A,1}, ..., v_{A,k-1} \} $, where $v_0$ is the shared part and $v_{A,i}$ are the pure signal directions, orthogonal to $v_0$.

    \item \textbf{Background space:} $\text{span}(B)$ is spanned by $ \{ v_0, v_{B,1}, ..., v_{B,m-1} \} $, where $v_{B,j}$ are the pure background directions.
\end{itemize}

The population covariances then become:

\begin{itemize}
    \item \textbf{Contrastive covariance:}
    $$E[S_{n}^+] = \lambda_{A} v_{0} v_{0}^\top + \sum_{i=1}^{k-1} \lambda_{A, i} v_{A, i} v_{A, i}^\top. $$
    
Here, $\lambda_{A,0}$ is the signal variance in direction $v_{0}$.
Thus, contributions from the pure background directions $\{v_{B,j}\}$ are still averaged out. The resulting expectation remains spanned only by the signal-related directions $ \{ v_0, v_{A,1}, ..., v_{A,k-1} \} $. The shared direction $v_{0}$ is \emph{not} cancelled.

    \item \textbf{Standard covariance:} 
$$E[S_{n}] = (\lambda_{A, 0}+\lambda_{B, 0}) v_{1} v_{1}^\top + \sum_{i=1}^{k-1} \lambda_{A, i} v_{A, i} v_{A, i}^\top + \sum_{i=1}^{m-1} \lambda_{B, i} v_{B, i} v_{B, i}^\top.$$ 
Here, $\lambda_{B, 0}$ is the background variance in direction $v_{0}$. The variance in the shared direction $v_{0}$ is amplified by both signal and background components.
\end{itemize}

\textbf{Intuition}--why $\pxx$ is robust to violations of orthogonality assumption: The robustness of $\pxx$ stems from how the generalized eigenvalue problem $S_{n}^{+} v = \lambda S_{n} v$ interacts with these modified covariance structures. We can analyze this through the lens of the asymptotic contrastive energy for each direction, as defined in our analysis for $\textbf{Lemma~\ref{lm:highdim_bound}}$.

\begin{itemize}
    \item \textbf{Pure background directions ($v_{B,j}$):} The contrastive energy $v_{B,j}^{\top} S_{n}^{+} v_{B,j}$ remains asymptotically zero, so these directions are filtered out. 

    \item \textbf{Shared direction ($v_0$):} The contrastive energy $v_0^{\top} S_{n}^{+} v_0$ is strictly positive due to the signal component. However, its variance in the standard covariance is now amplified by both signal ($\lambda_{A,0}$) and background ($\lambda_{B,0}$) components.
\end{itemize}

The resulting contrastive energy for shared direction will be smaller than that of a pure signal spike, but it will still be bounded away from zero. Therefore, the $\pxx$ objective still detects the shared direction as part of the signal space. The fundamental mechanism—isolating all directions with non-zero contrastive energy—remains intact.

\paragraph{Empirical validation with overlapping subspaces:}
To provide strong empirical evidence, we ran new simulations with non-orthogonal signal and background subspaces and will add these to the Appendix. 
We ran new simulations based on the setting in Section~\ref{sec:ExpmHighdim}, but introduced overlap between the signal and background subspaces. We aligned two background directions with the two weakest signal directions, creating a two-dimensional shared subspace. We tested this under two background noise levels.

\begin{itemize}
    \item \textbf{Signal variances:} $[50, 25, 20, 15, 10].$
    \item  \textbf{Moderate noise background:} $[500, 400, 300]$ (pure background) + $[25, 12.5]$ (shared background).
    \item \textbf{Large noise background:} $[500, 400, 300]$ (pure background) + $[100, 50]$ (shared background).
\end{itemize}

The results for the fixed aspect ratio regime and growing-spike regime are shown in Table \ref{tab:aspect_ratio_final_comparison_simple}--\ref{tab:aspect_ratio_comparison_v6_simple}. For demonstration, we compare against our original theoretical predictions (derived under orthogonality).

\begin{table}[h!]
  \centering
  \caption{Fixed aspect ratio with moderate overlapping noise}
  \label{tab:aspect_ratio_final_comparison_simple}
  \begin{tabular}{l cccccccccc}
    \toprule
    \textbf{Aspect Ratio} & 0.1 & 0.3 & 0.5 & 0.7 & 0.9 & 1.1 & 1.3 & 1.5 & 1.7 & 1.8 \\
    \midrule
    PCA++ & 0.137 & 0.207 & 0.269 & 0.287 & 0.317 & 0.311 & 0.356 & 0.388 & 0.416 & 0.431 \\
    PCA++ theory & 0.104 & 0.179 & 0.229 & 0.268 & 0.301 & 0.330 & 0.356 & 0.379 & 0.400 & 0.410 \\
    \bottomrule
  \end{tabular}
\end{table}

\begin{table}[h!]
  \centering
  \caption{Fixed aspect ratio with large overlapping noise}
  \label{tab:aspect_ratio_comparison_v4_simple}
  \begin{tabular}{l cccccccccc}
    \toprule
    \textbf{Aspect Ratio} & 0.1 & 0.3 & 0.5 & 0.7 & 0.9 & 1.1 & 1.3 & 1.5 & 1.7 & 1.8 \\
    \midrule
    PCA++ & 0.254 & 0.206 & 0.278 & 0.280 & 0.357 & 0.302 & 0.359 & 0.405 & 0.391 & 0.416 \\
    PCA++ theory & 0.104 & 0.179 & 0.229 & 0.268 & 0.301 & 0.330 & 0.356 & 0.379 & 0.400 & 0.410 \\
    \bottomrule
  \end{tabular}
\end{table}

\begin{table}[h!]
  \centering
  \caption{Growing-spike regime with moderate overlapping noise}
  \label{tab:aspect_ratio_comparison_v5_simple}
  \begin{tabular}{l cccccccccc}
    \toprule
    \textbf{Aspect Ratio} & 0.1 & 0.3 & 0.5 & 0.7 & 0.9 & 1.1 & 1.3 & 1.5 & 1.7 & 1.8 \\
    \midrule
    PCA++ & 0.121 & 0.197 & 0.216 & 0.261 & 0.299 & 0.334 & 0.369 & 0.374 & 0.400 & 0.426 \\
    PCA++ theory & 0.100 & 0.171 & 0.218 & 0.256 & 0.287 & 0.315 & 0.339 & 0.361 & 0.381 & 0.391 \\
    \bottomrule
  \end{tabular}
\end{table}

\begin{table}[h!]
  \centering
  \caption{Growing-spike regime with large overlapping noise}
  \label{tab:aspect_ratio_comparison_v6_simple}
  \begin{tabular}{l cccccccccc}
    \toprule
    \textbf{Aspect Ratio} & 0.1 & 0.3 & 0.5 & 0.7 & 0.9 & 1.1 & 1.3 & 1.5 & 1.7 & 1.8 \\
    \midrule
    PCA++ & 0.138 & 0.182 & 0.214 & 0.262 & 0.316 & 0.323 & 0.363 & 0.375 & 0.380 & 0.397 \\
    PCA++ theory & 0.100 & 0.171 & 0.218 & 0.256 & 0.287 & 0.315 & 0.339 & 0.361 & 0.381 & 0.391 \\
    \bottomrule
  \end{tabular}
\end{table}

\paragraph{Conclusion:} The results in Table \ref{tab:aspect_ratio_final_comparison_simple}--\ref{tab:aspect_ratio_comparison_v6_simple} show that $\pxx$ is remarkably robust, even when the orthogonality assumption is violated. In the moderate overlapping noise case, the empirical error of $\pxx$ continues to track the theoretical predictions remarkably well. This demonstrates that when the background variance in the shared subspace is not excessively large, the impact of the overlap is minimal.

In the large overlapping noise case, we observe a slight increase in estimation error, as expected. This is because the background noise in the shared directions becomes strong enough to reduce the effective signal-to-noise ratio, making recovery more challenging. Nevertheless, even in this challenging scenario, $\pxx$ remains stable and successfully recovers the signal subspace with controlled error, confirming that \textbf{perfect orthogonality is not a practical prerequisite for our method's success.}

\subsection{On the Gaussian latent factor assumption (Assumption~\ref{asm:latent_gaussian}):}

This assumption was made for analytical convenience, as it allows for the clean derivation of exact constants and closed-form error. However, we expect the core results to hold more generally for sub-Gaussian distributions.  

To provide strong empirical evidence for this claim, we have run new simulations where the Gaussian latent factors $(w_i, h_i)$ and noise $(\epsilon_i)$ from Assumptions~\ref{asm:latent_gaussian} and~\ref{asm:noise} were replaced with samples from a standardized $\text{Beta}(2,2)$ distribution (a symmetric, bounded, non-Gaussian distribution). We repeated the experiments from Section~\ref{sec:ExpmHighdim} (Figure~\ref{fig:3}) under this new setting.
    
The results in Table \ref{tab:aspect_ratio_comparison_v7_simple}--\ref{tab:aspect_ratio_comparison_v8_simple} show that the empirical performance of $\pxx$ under Beta-distributed noise  continues to align almost perfectly with our theoretical predictions, which were derived under the Gaussian assumption.

    \begin{table}[h!]
  \centering
  \caption{Fixed aspect ratio with Beta distribution}
  \label{tab:aspect_ratio_comparison_v7_simple}
  \begin{tabular}{l cccccccccc}
    \toprule
    \textbf{Aspect Ratio} & 0.1 & 0.3 & 0.5 & 0.7 & 0.9 & 1.1 & 1.3 & 1.5 & 1.7 & 1.8 \\
    \midrule
    PCA++ & 0.095 & 0.195 & 0.244 & 0.266 & 0.295 & 0.361 & 0.385 & 0.381 & 0.402 & 0.412 \\
    PCA++ theory & 0.104 & 0.179 & 0.229 & 0.268 & 0.301 & 0.330 & 0.356 & 0.379 & 0.400 & 0.410 \\
    \bottomrule
  \end{tabular}
\end{table}

    \begin{table}[h!]
  \centering
  \caption{Growing-spike regime with Beta distribution}
  \label{tab:aspect_ratio_comparison_v8_simple}
  \begin{tabular}{l cccccccccc}
    \toprule
    \textbf{Aspect Ratio} & 0.1 & 0.3 & 0.5 & 0.7 & 0.9 & 1.1 & 1.3 & 1.5 & 1.7 & 1.8 \\
    \midrule
    PCA++ & 0.112 & 0.171 & 0.223 & 0.260 & 0.296 & 0.331 & 0.348 & 0.360 & 0.389 & 0.409 \\
    PCA++ theory & 0.100 & 0.171 & 0.218 & 0.256 & 0.287 & 0.315 & 0.339 & 0.361 & 0.381 & 0.391 \\
    \bottomrule
  \end{tabular}
\end{table}
    
    This remarkable consistency provides strong evidence that the Gaussian assumption is a technical choice for analytical clarity rather than a strict requirement for the validity of our results, which appear to exhibit universality.

\subsection{On Assumption~\ref{assum:eigenvalue2} (Distinct growing spikes) \label{app:discussion_ass43}}

This assumption was made primarily for analytical convenience.  Even if the standard covariance $S_n$ has degenerate subspaces (i.e., multiple identical eigenvalues mixing signal and background components), the contrastive covariance $S_{n}^{+}$ resolves this ambiguity. Since $S_{n}^{+}$ has asymptotically zero energy on pure background directions, the generalized eigenvalue problem can still correctly identify the signal subspace and separate it from the background.

To empirically validate this claim, we have run a new set of simulations for the setting in Section~\ref{sec:ExpmHighdim} (Figure~\ref{fig:3}) but with:

\begin{itemize}
    \item Signal variances: $[50, 50, 20, 15, 10].$
    \item Background variances: $[500, 500, 300, 50, 50].$
\end{itemize}

The results in Table \ref{ass43}--\ref{ass43_1} below show that even with these degeneracies, the empirical subspace error for $\pxx$ continues to track our theoretical predictions, which were derived under the distinct spike assumption. This provides strong evidence that the assumption is a technical convenience rather than a practical necessity.

\begin{table}[h!]
  \caption{Fixed aspect ratio regime}
  \label{ass43}
  \begin{tabular}{lcccccccccc}
    \toprule
    \textbf{Aspect Ratio} & {0.1} & {0.3} & {0.5} & {0.7} & {0.9} & {1.1} & {1.3} & {1.5} & {1.7} & {1.8} \\
    \midrule
    $\pxx$ & 0.120 & 0.189 & 0.230 & 0.262 & 0.288 & 0.323 & 0.361 & 0.366 & 0.423 & 0.418 \\
    $\pxx$ theory & 0.104 & 0.179 & 0.229 & 0.268 & 0.301 & 0.330 & 0.356 & 0.379 & 0.400 & 0.410 \\
    \bottomrule
  \end{tabular}
\end{table}

\begin{table}[h!]
  \caption{Growing-spike regime}
  \label{ass43_1}
  \begin{tabular}{lcccccccccc}
    \toprule
    \textbf{Aspect Ratio} & {0.1} & {0.3} & {0.5} & {0.7} & {0.9} & {1.1} & {1.3} & {1.5} & {1.7} & {1.8} \\
    \midrule
    $\pxx$ & 0.136 & 0.195 & 0.228 & 0.270 & 0.307 & 0.321 & 0.353 & 0.384 & 0.403 & 0.384 \\
    $\pxx$ theory & 0.100 & 0.171 & 0.218 & 0.256 & 0.287 & 0.315 & 0.339 & 0.361 & 0.381 & 0.391 \\
    \bottomrule
  \end{tabular}
\end{table}

}

\section{Proofs}\label{sec:proof}
\subsection{Proof of Theorem~\ref{thm:consistency_vanillaPCA}}
\begin{proof}
Stacking $n$ paired samples into data matrices gives 
\begin{equation}\label{eq:model_matrix} 
X = \sigloadmat^\top \sig^\top + \bkgloadmat^\top \bkg^\top + \noisemat, \qquad
X^{+} = \sigloadmat^\top \sig^\top + \bkgloadmat'^\top \bkg^\top + \noisemat',
\end{equation}
where:
\[
\begin{aligned}
X &= (x_1, \ldots, x_{\ndata})^\top \in \RR^{\ndata \times \ndim}, \quad
X^{+} = (x_1^{+}, \ldots, x_{\ndata}^{+})^\top \in \RR^{\ndata \times \ndim}, \\
\sigloadmat &= (\sigload_1, \ldots, \sigload_{\ndata}) \in \RR^{\nsig \times \ndata}, \quad
\bkgloadmat = (\bkgload_1, \ldots, \bkgload_{\ndata}) \in \RR^{\nbkg \times \ndata}, \quad
\bkgloadmat' = (\bkgload_1', \ldots, \bkgload_{\ndata}') \in \RR^{\nbkg \times \ndata}, \\
\noisemat &= (\noise_1, \ldots, \noise_{\ndata})^\top \in \RR^{\ndata \times \ndim}, \quad
\noisemat' = (\noise_1', \ldots, \noise_{\ndata}')^\top \in \RR^{\ndata \times \ndim}.
\end{aligned}
\]
In the contrastive factor model, the covariance of each observation $x_i$ decomposes additively into signal, background, and noise:
\[
\EE[x_i x_i^\top] = \sig \sig^\top + \bkg \bkg^\top + I_{\ndim}.
\]
This decomposition ensures that the spectrum of \(\EE[x_i x_i^\top]\) directly reflects the signal, background, and noise contributions.

Recall that
\[
\covmatsample_{\ndata}^{+}
=\frac{1}{2n}\bigl(X^\top X^{+} + X^{+\top}X\bigr).
\]
Substituting the matrix form of the generative model~\eqref{eq:model_matrix}, we obtain the decomposition
\begin{equation*}
    \covmatsample_{\ndata}^{+} = \sig \sig^\top + \frac{1}{2}\left(\errormat_{\ndata} + \errormat_{\ndata}^\top \right),
\end{equation*}
where the error matrix $\errormat_{\ndata}$ collects all cross‐terms:
\begin{align}
\label{eq:errmatrix} \notag
\errormat_{\ndata}=& \frac{1}{\ndata} \Big(\sig(\sigloadmat \sigloadmat^\top - \ndata I_{\nsig})\sig^\top + \sig \sigloadmat \bkgloadmat'^\top \bkg^\top + \sig \sigloadmat \noisemat' \\ 
        &+ \bkg \bkgloadmat \sigloadmat^\top \sig^\top + \bkg \bkgloadmat \bkgloadmat'^\top \bkg^\top  + \bkg \bkgloadmat \noisemat'\\ \notag
        &+ \noisemat^\top \sigloadmat^\top \sig^\top + \noisemat^\top \bkgloadmat'^\top \bkg^\top + \noisemat^\top \noisemat'  \Big).
\end{align}

By Assumptions~\ref{asm:orthogonal}-\ref{asm:noise}, each latent factor matrix has independent, zero‐mean Gaussian columns, and the noise matrices are independent with zero mean.  Consequently,
\[
\EE[\errormat_{\ndata}]=\mathbf{0},
\]
and hence
\[
\EE\bigl[\covmatsample_{\ndata}^{+}]
=\sig \sig^\top,
\]
as claimed.
\end{proof}

\subsection{Lemma \ref{lm:highdim_bound}}

Lemma \ref{lm:highdim_bound} formalizes the behavior of the contrastive sample covariance matrix $S^+_n$
 when projected onto the sample signal and background eigenvectors derived from $S_n$:
\begin{lemma}[Contrastive energy of sample directions]\label{lm:highdim_bound}
    Under Assumptions~\ref{asm:orthogonal}-\ref{asm:noise} and~\ref{assum:eigenvalue},  as $\npinfty$ with $\ndim / \ndata \rightarrow \aspratio \in (0, + \infty)$, for each $1 \leq j \leq \nsig$
    \begin{equation*}
        \lim_{\npinfty} \frac{1}{\hat{\eigval}_{\sig, j}} \hat{\eigvec}_{\sig, j}^\top \covmatsample_{\ndata}^{+} \hat{\eigvec}_{\sig, j} \geq \lim_{\npinfty} \frac{1}{2} \left( 1 + \frac{\eigval_{\sig, j}}{\hat{\eigval}_{\sig, j}}\frac{1 - \aspratio \eigval_{\sig, j}^{-2}}{1 + \aspratio \eigval_{\sig, j}^{-1}} - \frac{(1 + \sqrt{c})^2}{\hat{\eigval}_{\sig, j}}  \right) \qas,
    \end{equation*}
    and for each $1\leq j \leq \nbkg$,
    \begin{equation*}
        \lim_{\npinfty} \frac{1}{\hat{\eigval}_{\bkg, j}} \hat{\eigvec}_{\bkg, j}^\top \covmatsample_{\ndata}^{+} \hat{\eigvec}_{\bkg, j} = 0 \qas.
    \end{equation*}
\end{lemma} 
In other words, in the fixed-$c$ limit the contrastive covariance
$\covmatsample_{\ndata}^{+}$ retains strictly positive "energy" along each true signal direction, but asymptotically vanishes along every background direction.

\paragraph{Proof of Lemma~\ref{lm:highdim_bound}}
\begin{proof}
We begin by noting a useful identity. For any vector $v \in \RR^{\ndim}$
\begin{equation}\label{eq:eqidt}
    \begin{split}
        v^\top \covmatsample_{\ndata}^{+} v &= \frac{1}{2\ndata} v^\top \left(  X^\top X^{+} + X^{+\top} X\right) v,\\
        &= \frac{1}{2\ndata} v^\top \left(  X^\top X^{+} + (X^\top X^{+})^\top\right) v,\\
        &= \frac{1}{\ndata} v^\top X^\top X^{+} v,
    \end{split}
\end{equation}
where the last equality follows from the symmetry property of quadratic forms. Specifically, for any matrix $M \in \RR^{\ndim \times \ndim}$ and vector $a \in \RR^{\ndim}$:
\begin{equation*}
    a^\top M a = a^\top M^\top a.
\end{equation*}
With this identity in hand, our task reduces to proving the following two limits as $\npinfty$ with $\ndim/\ndata \rightarrow \aspratio \in (0, +\infty)$:
\begin{align*}
        \lim_{\npinfty} \frac{1}{\hat{\eigval}_{\sig, j}} \hat{\eigvec}_{\sig, j}^\top \frac{1}{\ndata} X^\top X^{+} \hat{\eigvec}_{\sig, j} & \geq  \lim_{\npinfty} \frac{1}{2} \left( 1 + \frac{\eigval_{\sig, j}}{\hat{\eigval}_{\sig, j}}\frac{1 - \aspratio \eigval_{\sig, j}^{-2}}{1 + \aspratio \eigval_{\sig, j}^{-1}} - \frac{(1 + \sqrt{c})^2}{\hat{\eigval}_{\sig, j}}  \right), \quad  1\leq j \leq \nsig \qas,\\
        \lim_{\npinfty} \frac{1}{\hat{\eigval}_{\bkg, j}} \hat{\eigvec}_{\bkg, j}^\top \frac{1}{\ndata} X^\top X^{+} \hat{\eigvec}_{\bkg, j} & = 0, \quad 1\leq j \leq \nbkg \qas
    \end{align*}
\paragraph{Step 1:} Proof of $\lim_{\npinfty} \frac{1}{\hat{\eigval}_{\bkg, j}} \hat{\eigvec}_{\bkg, j}^\top \frac{1}{\ndata} X^\top X^{+} \hat{\eigvec}_{\bkg, j} = 0$ 

Recall the decomposition
\begin{equation}\label{eq:thmeq1}
    \begin{split}
        X^\top X^{+} - \ndata \sig \sig^\top =& \ndata \errormat_{\ndata}\\
        =& \sig(\sigloadmat \sigloadmat^\top - \ndata I_{\nsig})\sig^\top + \sig \sigloadmat \bkgloadmat'^\top \bkg^\top + \sig \sigloadmat \noisemat' \\
        &+ \bkg \bkgloadmat \sigloadmat^\top \sig^\top + \bkg \bkgloadmat \bkgloadmat'^\top \bkg^\top  + \bkg \bkgloadmat \noisemat'\\
        &+ \noisemat^\top \sigloadmat^\top \sig^\top + \noisemat^\top \bkgloadmat'^\top \bkg^\top + \noisemat^\top \noisemat'.
    \end{split}
\end{equation}
We analyze each term $M$ in $\errormat_{\ndata}$ to show
\begin{equation*}
    \frac{1}{\ndata} \hat{\eigvec}_{\bkg, j}^\top M \hat{\eigvec}_{\bkg, j} \overlineninfty 0 \qas.
\end{equation*}

\begin{enumerate}
\item[Case 1:] $M = \ndata \sig \sig^\top$

Since $\sig^\top \hat{\eigvec}_{\bkg, j}$ is the projection of $\hat{\eigvec}_{\bkg, j}$ onto the column space of $\sig$ and $\hat{\eigvec}_{\bkg, j}$ itself aligns with the column space o $\bkg$, Lemma~\ref{lm:HDspike} implies:
\begin{equation*}
     \sig^\top \hat{\eigvec}_{\bkg, j} = \left(\begin{array}{c}
\sqrt{\eigval_{\sig,1}}\eigvec_{\sig,1}^\top  \hat{\eigvec}_{\bkg, j}\\
\vdots \\
\sqrt{\eigval_{\sig,\nsig}}\eigvec_{\sig,\nsig}^\top \hat{\eigvec}_{\bkg, j}
\end{array}\right) \overlineninfty \vec{0} \in \RR^{\nsig} \qas.
\end{equation*}
Hence,
\begin{equation*}
    \hat{\eigvec}_{\bkg, j}^\top \sig \sig^\top \hat{\eigvec}_{\bkg, j} \overlineninfty 0 \qas.
\end{equation*}

\item[Case 2:] $M =  \sig(\sigloadmat \sigloadmat^\top - \ndata I)\sig^\top $

Since $\sigloadmat \sigloadmat^\top \in \RR^{\nsig \times \nsig}$ is of fixed dimension, the strong law of large numbers implies 
\begin{equation*}
    \frac{1}{\ndata} \sigloadmat \sigloadmat^\top \xrightarrow{\ndata \rightarrow +\infty} I_{\nsig} \qas.
\end{equation*}
Hence,
\begin{equation*}
    \frac{1}{\ndata}\hat{\eigvec}_{\bkg, j}^\top \sig(\sigloadmat \sigloadmat^\top - \ndata I)\sig^\top \hat{\eigvec}_{\bkg, j} \leq \left\| \frac{1}{\ndata}\sigloadmat \sigloadmat^\top -  I \right\|_{2} \left\|\sig^\top \hat{\eigvec}_{\bkg, j} \right\|_{2}^{2} \xrightarrow{\ndata \rightarrow +\infty} 0 \qas.
\end{equation*}
An analogous argument applies to the terms
\begin{equation*}
    \frac{1}{\ndata}\hat{\eigvec}_{\bkg, j}^\top \sig \sigloadmat \bkgloadmat^\top \bkg^\top \hat{\eigvec}_{\bkg, j}, \frac{1}{\ndata}\hat{\eigvec}_{\bkg, j}^\top \bkg \bkgloadmat^{+} \bkgloadmat^\top \bkg^\top \hat{\eigvec}_{\bkg, j} ~\text{and}~ \frac{1}{\ndata}\hat{\eigvec}_{\bkg, j}^\top \bkg \bkgloadmat^{+} \bkgloadmat^\top \bkg^\top \hat{\eigvec}_{\bkg, j}.
\end{equation*}
This shows that all such terms almost surely converge to zero.

\item[Case 3:] $M =  \bkg \bkgloadmat \noisemat'$

Since $Z'$ and $\hat{\eigvec}_{\bkg, j}$ are independent, we can condition on $\hat{\eigvec}_{\bkg, j}$ and write
\begin{equation*}
    Z' \hat{\eigvec}_{\bkg, j} = \left(\begin{array}{c}
z_{1}  \\
\vdots \\
z_{\ndata}
\end{array}\right),
\end{equation*}
where $z_{1}, \ldots, z_{\ndata}$ are $\iid$ Gaussian random variables with mean zero and variance $\|\hat{\eigvec}_{\bkg, j}\|_{2}^{2} = 1.$ These random variables are independent of $\hat{\eigvec}_{\bkg, j}$. Consequently, $\bkgloadmat$ is also independent of $Z' \hat{\eigvec}_{\bkg, j}$ and given $Z' \hat{\eigvec}_{\bkg, j}$, the entries of $\bkgloadmat Z' \hat{\eigvec}_{\bkg, j}$ are $\iid$ random variables with mean zero and variance $\|Z' \hat{\eigvec}_{\bkg, j}\|_{2}^{2}$. By the strong law of large numbers,
\begin{equation*}
\begin{split}
    \frac{1}{\ndata^2}\|\bkgloadmat Z' \hat{\eigvec}_{\bkg, j}\|_{2}^{2} &= \frac{1}{\ndata^2}(\|\bkgloadmat Z' \hat{\eigvec}_{\bkg, j}\|_{2}^{2} - \nbkg \|Z' \hat{\eigvec}_{\bkg, j}\|_{2}^{2}) + \frac{1}{\ndata^2}(\nbkg \|Z' \hat{\eigvec}_{\bkg, j}\|_{2}^{2} - \nbkg \ndata) + \frac{\nbkg \ndata}{\ndata^2}\\
    &\xrightarrow{\ndata \rightarrow +\infty} 0 \qas.
\end{split}    
\end{equation*}
Hence, 
\begin{equation*}
    \frac{1}{\ndata}\hat{\eigvec}_{\bkg, j}^\top \bkg \bkgloadmat \noisemat' \hat{\eigvec}_{\bkg, j} \leq \frac{1}{\ndata} \|\bkg^\top \hat{\eigvec}_{\bkg, j}\|_{2} \|\bkgloadmat Z' \hat{\eigvec}_{\bkg, j}\|_{2} \xrightarrow{\ndata \rightarrow +\infty} 0 \qas,
\end{equation*}
where the limit of $\|\bkg^\top \hat{\eigvec}_{\bkg, j}\|_{2}$ is bounded almost surely by Lemma~\ref{lm:HDspike}. A similar argument applies to show
\begin{equation*}
    \frac{1}{\ndata}\hat{\eigvec}_{\bkg, j}^\top \sig \sigloadmat \noisemat' \hat{\eigvec}_{\bkg, j} \xrightarrow{\ndata \rightarrow +\infty} 0 \qas.
\end{equation*}

\item[Case 4:] $M = \noisemat^\top \noisemat'$

Using the same reasoning as in Case 3, the entries of $\noisemat' \hat{\eigvec}_{\bkg, j}$ are $\iid$ Gaussian random variable with mean zero and variance 1, and are independent of both $\hat{\eigvec}_{\bkg, j}$ and $\noisemat$. Consequently, conditioning on $\noisemat \hat{\eigvec}_{\bkg, j}$, implies that 
$\frac{1}{\ndata} \hat{\eigvec}_{\bkg, j}^\top \noisemat^\top \noisemat' \hat{\eigvec}_{\bkg, j}$
follows a Gaussian distribution with mean zero and variance $\frac{1}{\ndata^2} \|\noisemat \hat{\eigvec}_{\bkg, j}\|_{2}^{2}$. By Lemma~\ref{lm:Bai-yin}, 
\begin{equation*}
    \lim_{\npinfty} \frac{1}{\ndata} \|\noisemat \hat{\eigvec}_{\bkg, j}\|_{2}^{2} \leq (1 + \sqrt{\aspratio})^2 < + \infty \qas.
\end{equation*}
Thus
\begin{equation*}
    \frac{1}{\ndata}\hat{\eigvec}_{\bkg, j}^\top \noisemat^\top \noisemat' \hat{\eigvec}_{\bkg, j} \overlineninfty 0 \qas.
\end{equation*}

\item[Case 5:] $M = \noisemat^\top \sigloadmat^\top \sig^\top$

From Cases 1, 2 and 4, we already know
\begin{align*}
    \| \sig^\top \hat{\eigvec}_{\bkg, j}\|_{2} &\overlineninfty 0 \qas,\\
    \frac{1}{\ndata} \sigloadmat \sigloadmat^\top &\overlineninfty I \qas,\\
    \frac{1}{\ndata} \|\noisemat \hat{\eigvec}_{\bkg, j}\|_{2}^{2} &\leq (1 + \sqrt{\aspratio})^2 \qas.
\end{align*}
We now use these facts to bound
\begin{align*}
    \frac{1}{\ndata}\hat{\eigvec}_{\bkg, j}^\top \noisemat^\top \sigloadmat^\top \sig^\top \hat{\eigvec}_{\bkg, j} &\leq  \frac{1}{\ndata} \|\noisemat \hat{\eigvec}_{\bkg, j}\|_{2} \| \sigloadmat^\top \sig^\top \hat{\eigvec}_{\bkg, j}\|_{2}\\
    &\leq \frac{1}{\ndata} \|\noisemat \hat{\eigvec}_{\bkg, j}\|_{2} \| \sigloadmat \sigloadmat^\top\|_{2} \| \sig^\top \hat{\eigvec}_{\bkg, j}\|_{2}\\
    &\overlineninfty 0 \qas.
\end{align*}

\item[Case 6:] $M = \noisemat^\top \bkgloadmat'^\top \bkg^\top$

By the same reasoning as in Case 3, the entries of $\bkgloadmat'^\top \bkg^\top \hat{\eigvec}_{\bkg, j}$ are $\iid$ Gaussian variable with mean zero and variance $\| \bkg^\top \hat{\eigvec}_{\bkg, j}\|$. According to Lemma~\ref{lm:HDspike}, $\| \bkg^\top \hat{\eigvec}_{\bkg, j}\|$ remains bounded almost surely. Consequently,
\begin{equation*}
    \frac{1}{\ndata}\hat{\eigvec}_{\bkg, j}^\top \noisemat^\top \bkgloadmat'^\top \bkg^\top \hat{\eigvec}_{\bkg, j} \leq \frac{1}{\ndata} \| \noisemat \hat{\eigvec}_{\bkg, j} \|_{2} \| \bkgloadmat'^\top \bkg^\top \hat{\eigvec}_{\bkg, j}\|_{2} \overlineninfty 0 \qas,
\end{equation*}
where the convergence follows from the strong law of large numbers and $\lim_{\npinfty} \frac{1}{\ndata}\| \noisemat \hat{\eigvec}_{\bkg, j} \|_{2}^2 \leq (1 + \sqrt{\aspratio})^2$ is shown in Case 4.
\end{enumerate}
Putting these cases together, all contributions of terms in equation~\eqref{eq:thmeq1} converge to zero. Dividing by $\hat{\eigval}_{\bkg, j}$ does not affect the limit, since $\hat{\eigval}_{\bkg, j}$ converges to a deterministic value by Lemma~\ref{lm:HDspike}. It follows that 
\begin{equation*}
    \frac{1}{\hat{\eigval}_{\bkg, j}} \hat{\eigvec}_{\bkg, j}^\top \frac{1}{\ndata} X^\top X^{+} \hat{\eigvec}_{\bkg, j} \xrightarrow{\ndata \rightarrow +\infty} 0 \qas,
\end{equation*}
completing the proof of Step 1.

\paragraph{Step 2:} Proof of $\lim_{\npinfty} \frac{1}{\hat{\eigval}_{\sig, j}} \hat{\eigvec}_{\sig, j}^\top \frac{1}{\ndata} X^\top X^{+} \hat{\eigvec}_{\sig, j}  \geq  \lim_{\npinfty} \frac{1}{2} \left( 1 + \frac{\eigval_{\sig, j}}{\hat{\eigval}_{\sig, j}}\frac{1 - \aspratio \eigval_{\sig, j}^{-2}}{1 + \aspratio \eigval_{\sig, j}^{-1}} - \frac{(1 + \sqrt{c})^2}{\hat{\eigval}_{\sig, j}}  \right) $

By the same reasoning employed in Step 1, each term $M$ appearing in equation~\eqref{eq:thmeq1} satisfies \begin{equation*}
    \frac{1}{\ndata} \hat{\eigvec}_{\sig, j}^\top M \hat{\eigvec}_{\sig, j} \xrightarrow{\ndata \rightarrow +\infty} 0 \qas,
\end{equation*}
with the exceptions of $\sig \sig^\top$ and $\noisemat^\top \sigloadmat^\top \sig^\top$. 
\begin{itemize}
    \item[Case 1:] $M = \sig \sig^\top$
    
We have
\begin{align*}
\hat{\eigvec}_{\sig, j}^\top \sig \sig^\top \hat{\eigvec}_{\sig, j} &=  \|\sig^\top \hat{\eigvec}_{\sig, j}\|_{2}^{2} 
= \left\| \left(\begin{array}{c}
\sqrt{\eigval_{\sig,1}}\eigvec_{\sig,1}^\top  \hat{\eigvec}_{\sig, j}\\
\vdots \\
\sqrt{\eigval_{\sig,\nsig}}\eigvec_{\sig,\nsig}^\top \hat{\eigvec}_{\sig, j}
\end{array}\right) \right\|_{2}^{2}
\overlineninfty \eigval_{\sig, j} \frac{1 - \aspratio \eigval_{\sig, j}^{-2}}{1 + \aspratio \eigval_{\sig, j}^{-1}} \qas,
\end{align*}
where the limit follows from Lemma~\ref{lm:HDspike}.

\item[Case 2:] $M = \noisemat^\top \sigloadmat^\top \sig^\top$

Note that
\begin{align*}
    X^\top X &= \sig \sigloadmat \sigloadmat^\top \sig^\top + \sig \sigloadmat \bkgloadmat^\top \bkg^\top + \sig \sigloadmat \noisemat \\
        &+ \bkg \bkgloadmat \sigloadmat^\top \sig^\top + \bkg \bkgloadmat \bkgloadmat^\top \bkg^\top  + \bkg \bkgloadmat \noisemat\\
        &+ \noisemat^\top \sigloadmat^\top \sig^\top + \noisemat^\top \bkgloadmat^\top \bkg^\top + \noisemat^\top \noisemat.
\end{align*}
Following the argument in Step 1 for each of these summands, we obtain
\begin{equation}\label{eq:eq2}
    \frac{1}{\ndata} \hat{\eigvec}_{\sig, j}^\top X^\top X \hat{\eigvec}_{\sig, j} - \eigval_{\sig, j} \frac{1 - \aspratio \eigval_{\sig, j}^{-2}}{1 + \aspratio \eigval_{\sig, j}^{-1}} - 2 \frac{1}{\ndata} \hat{\eigvec}_{\sig, j}^\top \sig \sigloadmat \noisemat \hat{\eigvec}_{\sig, j} - \frac{1}{\ndata} \hat{\eigvec}_{\sig, j}^\top \noisemat^\top \noisemat \hat{\eigvec}_{\sig, j} \overlineninfty 0 \qas.
\end{equation}
Since $\frac{1}{\ndata} \hat{\eigvec}_{\sig, j}^\top X^\top X \hat{\eigvec}_{\sig, j} = \hat{\eigval}_{\sig, j}$, it follows that
\begin{align*}
    \lim_{\npinfty} \frac{2}{\ndata} \hat{\eigvec}_{\sig, j}^\top \sig \sigloadmat \noisemat \hat{\eigvec}_{\sig, j} &= \lim_{\npinfty}\hat{\eigval}_{\sig, j} - \frac{1}{\ndata} \hat{\eigvec}_{\sig, j}^\top \noisemat^\top \noisemat \hat{\eigvec}_{\sig, j} - \eigval_{\sig, j} \frac{1 - \aspratio \eigval_{\sig, j}^{-2}}{1 + \aspratio \eigval_{\sig, j}^{-1}} \\
    &\geq \lim_{\npinfty} \hat{\eigval}_{\sig, j} - (1 + \sqrt{\aspratio})^2 - \eigval_{\sig, j} \frac{1 - \aspratio \eigval_{\sig, j}^{-2}}{1 + \aspratio \eigval_{\sig, j}^{-1}} \qas,
\end{align*}
where the last line follows from Lemma~\ref{lm:Bai-yin}.
\end{itemize}
Combining the contributions from $\sig \sig^\top$ and $\noisemat^\top \sigloadmat^\top \sig^\top$ with the negligible effects of all other terms yields the desired lower bound:
\begin{align}\label{eq:eq1}
    \lim_{\npinfty} \frac{1}{\hat{\eigval}_{\sig, j}} \hat{\eigvec}_{\sig, j}^\top \frac{1}{\ndata} X^\top X^{+} \hat{\eigvec}_{\sig, j} &= \lim_{\npinfty} \frac{1}{\hat{\eigval}_{\sig, j}} \hat{\eigvec}_{\sig, j}^\top \sig \sig^\top \hat{\eigvec}_{\sig, j}  + \lim_{\npinfty} \frac{1}{\hat{\eigval}_{\sig, j}} \frac{1}{\ndata} \hat{\eigvec}_{\sig, j}^\top \sig \sigloadmat \noisemat \hat{\eigvec}_{\sig, j}\\ \notag
    & \geq \lim_{\npinfty} \frac{1}{2} \left( 1 + \frac{\eigval_{\sig, j}}{\hat{\eigval}_{\sig, j}}\frac{1 - \aspratio \eigval_{\sig, j}^{-2}}{1 + \aspratio \eigval_{\sig, j}^{-1}} - \frac{(1 + \sqrt{c})^2}{\hat{\eigval}_{\sig, j}}  \right) \qas.
\end{align}
This completes the proof of Step 2, and hence the theorem.
\end{proof}

\subsection{Lemma \ref{lm:highdim_bound2}}

We extend the fixed-aspect-ratio analysis (Lemma~\ref{lm:highdim_bound} and Theorem~\ref{thm:dist}) to the growing-spike regime.
\begin{lemma}[Contrastive energy in growing-spike regime]\label{lm:highdim_bound2}
    Under Assumptions~\ref{asm:orthogonal}-\ref{asm:noise} and \ref{assum:eigenvalue2}, as $\npinfty$,  the contrastive energy along each sample signal direction satisfies
    \begin{equation*}
    \lim_{\npinfty} \frac{1}{\hat{\eigval}_{\sig, j}} \hat{\eigvec}_{\sig, j}^\top \covmatsample_{\ndata}^{+} \hat{\eigvec}_{\sig, j} \geq \frac{1}{2(1 + \aspratio_{\sig, j})} + \frac{1}{2(1 + \aspratio_{\sig, j})^2},
    \end{equation*}
    for each $1 \leq j \leq \nsig$, and
    \begin{equation*}
        \lim_{\npinfty} \frac{1}{\hat{\eigval}_{\bkg, j}} \hat{\eigvec}_{\bkg, j}^\top \covmatsample_{\ndata}^{+} \hat{\eigvec}_{\bkg, j} = 0 \qas.
    \end{equation*}
    for each background index $1\leq j \leq \nbkg$.
\end{lemma}
This lemma shows that--even when both dimension and spike strengths grow--the contrastive covariance 
 $\covmatsample_{\ndata}^{+}$, still concentrates nonzero contrastive "energy" on every true signal direction while vanishing on all background directions, thus preserving the clean separation needed for accurate subspace recovery under our hard‐uniformity constraint.

\paragraph{Proof of Lemma~\ref{lm:highdim_bound2}}
\begin{proof}
    The proof follows similar lines as that of Theorem~\ref{lm:highdim_bound}, with modifications to account for differences in the current setting. Specifically, we establish the following key results:
    \begin{enumerate}
        \item As $\npinfty$, $\frac{1}{\hat{\eigval}_{\bkg, j}} \| \sig^\top \hat{\eigvec}_{\bkg, j} \|^2 \rightarrow 0 \qas$, for all $1 \leq j \leq \nbkg$.
        
        \item As $\npinfty$, $\frac{1}{\hat{\eigval}_{\sig, j}} \| \sig^\top \hat{\eigvec}_{\sig, j} \|^2 \rightarrow \frac{1}{(1 + \aspratio_{\sig, j})^2} \qas$, for all $1 \leq j \leq \nbkg$.
        
        \item $\lim_{\npinfty} \frac{1}{\ndata\hat{\eigval}_{\bkg, j}} \eigval_{\max}(\noisemat^\top \noisemat) = \aspratio_{\bkg, j}$, for $1 \leq j \leq \nbkg$. An analogous result holds for $\hat{\eigval}_{\sig, j}$.
        
        \item $\lim_{\npinfty} \frac{2}{\ndata \hat{\eigval}_{\sig, j}} \hat{\eigvec}_{\sig, j}^\top \sig \sigloadmat \noisemat \hat{\eigvec}_{\sig, j} \geq \frac{\aspratio_{\sig, j}}{(1 + \aspratio_{\sig, j})^2}$.
    \end{enumerate}
    Using these results along with analogous arguments from the proof of Theorem~\ref{lm:highdim_bound}, we derive the following lower bound:
    \begin{align*}
    \lim_{\npinfty} \frac{1}{\hat{\eigval}_{\sig, j}} \hat{\eigvec}_{\sig, j}^\top \frac{1}{\ndata} X^\top X^{+} \hat{\eigvec}_{\sig, j} &= \lim_{\npinfty} \frac{1}{\hat{\eigval}_{\sig, j}} \hat{\eigvec}_{\sig, j}^\top \sig \sig^\top \hat{\eigvec}_{\sig, j}  + \lim_{\npinfty} \frac{1}{\hat{\eigval}_{\sig, j}} \frac{1}{\ndata} \hat{\eigvec}_{\sig, j}^\top \sig \sigloadmat \noisemat \hat{\eigvec}_{\sig, j}\\
    & \geq \frac{1}{2(1 + \aspratio_{\sig, j})} + \frac{1}{2(1 + \aspratio_{\sig, j})^2}.
\end{align*}
We now detail the proofs of the above assertions separately:
\begin{enumerate}
    \item[Case 1:] (Analysis of $\frac{1}{\hat{\eigval}_{\bkg, j}} \| \sig^\top \hat{\eigvec}_{\bkg, j} \|^2$)

    We have
    \begin{equation*}
        \frac{1}{\hat{\eigval}_{\bkg, j}} \|\sig^\top \hat{\eigvec}_{\bkg, j}\|_{2}^{2} = \left\| \left[\begin{array}{c}
\sqrt{\frac{\eigval_{\sig,1}}{\hat{\eigval}_{\bkg, j}}}\eigvec_{\sig,1}^\top  \hat{\eigvec}_{\bkg, j}\\
\vdots \\
\sqrt{\frac{\eigval_{\sig,\nsig}}{\hat{\eigval}_{\bkg, j}}}\eigvec_{\sig,\nsig}^\top \hat{\eigvec}_{\bkg, j}
\end{array}\right] \right\|_{2}^{2} \overlineninfty 0 \qas,
    \end{equation*}
    where the convergence follows from Lemma~\ref{lm:HDspike2} and the observation that
    \begin{equation*}
        \lim_{\npinfty}\frac{\eigval_{\sig,i}}{\hat{\eigval}_{\bkg, j}} = \frac{1}{1 + \aspratio_{j}} \frac{\eigval_{\sig,i}}{\eigval_{\bkg, j}} = \frac{1}{1 + \aspratio_{\bkg, j}} \frac{\aspratio_{\bkg, j}}{\aspratio_{\sig, i}} \leq \infty.
    \end{equation*}

    \item[Case 2:] (Analysis of $\frac{1}{\hat{\eigval}_{\sig, j}} \| \sig^\top \hat{\eigvec}_{\sig, j} \|^2$)
    
    Similarly, using Lemma~\ref{lm:HDspike2}, we have
    \begin{equation*}
        \frac{1}{\hat{\eigval}_{\sig, j}} \| \sig^\top \hat{\eigvec}_{\sig, j} \|^2 = \left\| \left[\begin{array}{c}
\sqrt{\frac{\eigval_{\sig,1}}{\hat{\eigval}_{\sig, j}}}\eigvec_{\sig,1}^\top  \hat{\eigvec}_{\sig, j}\\
\vdots \\
\sqrt{\frac{\eigval_{\sig,\nsig}}{\hat{\eigval}_{\sig, j}}}\eigvec_{\sig,\nsig}^\top \hat{\eigvec}_{\sig, j}
\end{array}\right] \right\|_{2}^{2} \overlineninfty (1 + \aspratio_{\sig, j})^{-2} \qas.
    \end{equation*}

    \item[Case 3:] (Analysis of $\frac{1}{\ndata\hat{\eigval}_{\bkg, j}} \eigval_{\max}(\noisemat^\top \noisemat)$)

    Since $\noisemat^\top \noisemat$ and $\noisemat \noisemat^\top$ share identical nonzero eigenvalues, Lemma~\ref{lm:Bai-yin} implies
    \begin{equation*}
        \lim_{\npinfty} \eigval_{\max}\left(\frac{1}{\ndim} \noisemat \noisemat^\top \right) = \lim_{\npinfty} \left(1 + \sqrt{\frac{\ndata}{\ndim}} \right)^2 = 1.
    \end{equation*}
    It then follows that
    \begin{equation*}
        \lim_{\npinfty} \frac{1}{\ndata\hat{\eigval}_{\bkg, j}} \eigval_{\max}(\noisemat^\top \noisemat) = \lim_{\npinfty} \frac{\ndim}{\ndata\hat{\eigval}_{\bkg, j}} \eigval_{\max}\left(\frac{1}{\ndim}\noisemat^\top \noisemat \right) = \aspratio_{\bkg, j},
    \end{equation*}
    for $1 \leq j \leq \nbkg$.

   \item[Case 4:] (Analysis of $\frac{2}{\ndata \hat{\eigval}_{\sig, j}} \hat{\eigvec}_{\sig, j}^\top \sig \sigloadmat \noisemat \hat{\eigvec}_{\sig, j}$)

    By arguments analogous to those in Case 2 of Step 2 in the proof of Lemma~\ref{lm:highdim_bound}, we deduce that
    \begin{align*}
    \lim_{\npinfty} \frac{2}{\ndata\hat{\eigval}_{\sig, j}} \hat{\eigvec}_{\sig, j}^\top \sig \sigloadmat \noisemat \hat{\eigvec}_{\sig, j} &= \lim_{\npinfty}1 - \frac{1}{\ndata\hat{\eigval}_{\sig, j}} \hat{\eigvec}_{\sig, j}^\top \noisemat^\top \noisemat \hat{\eigvec}_{\sig, j} - \frac{1}{\hat{\eigval}_{\sig, j}} \hat{\eigvec}_{\sig, j}^\top \sig \sig^\top \hat{\eigvec}_{\sig, j}\\
    &\geq 1 - \frac{\aspratio_{\sig, j}}{(1 + \aspratio_{\sig, j})} - \frac{1}{(1 + \aspratio_{\sig, j})^{2}}\\
    &= \frac{\aspratio_{\sig, j}}{(1 + \aspratio_{\sig, j})^2},
\end{align*}
 for $1 \leq j \leq \nbkg$, which establishes the claimed inequality. 
\end{enumerate}
The proof is thus complete.
\end{proof}

\subsection{Proof of Theorem~\ref{thm:CPCAlowdim}}

\begin{proof}
We analyze the distance between the estimated signal subspace \(\widehat{\eigspace}_\sig\) and its population counterpart \(\eigspace_\sig\) using the Davis-Kahan \(\sin\Theta\) theorem (Lemma~\ref{lm:sintheta}). Recall equation~\eqref{eq:errmatrix}
\begin{equation*}
    \covmatsample_{\ndata}^{+} = \sig \sig^\top + \frac{1}   {2}\left(\errormat_{\ndata} + \errormat_{\ndata}^\top \right),
\end{equation*}
where \(\sig \sig^\top\) represents the signal component and \(\errormat_{\ndata}\) represents the error matrix. Furthermore, the signal component is expressed as:
\begin{equation*}
    \sig \sig^\top = \sum_{j = 1}^{\nsig} \eigval_{\sig, j} \eigvec_{\sig, j} \eigvec_{\sig, j}^\top.
\end{equation*}
To apply Lemma~\ref{lm:sintheta}, we need to bound $\|\errormat_{\ndata}\|_2$ such that:
\begin{equation*}
    \|\errormat_{\ndata}\|_2 \leq \left(1 - \frac{1}{\sqrt{2}}\right) \left| \eigval_{\sig, \nsig} \right|.
\end{equation*}
\paragraph{Bounding the error matrix $\errormat_{\ndata}$.}Recall that the error matrix $\errormat_{\ndata}$~\eqref{eq:errmatrix}:
\begin{align*}
 \errormat_{\ndata}=& \frac{1}{\ndata} \Big(\sig(\sigloadmat \sigloadmat^\top - \ndata I_{\nsig})\sig^\top + \sig \sigloadmat \bkgloadmat'^\top \bkg^\top + \sig \sigloadmat \noisemat' \\ 
        &+ \bkg \bkgloadmat \sigloadmat^\top \sig^\top + \bkg \bkgloadmat \bkgloadmat'^\top \bkg^\top  + \bkg \bkgloadmat \noisemat'\\ \notag
        &+ \noisemat^\top \sigloadmat^\top \sig^\top + \noisemat^\top \bkgloadmat'^\top \bkg^\top + \noisemat^\top \noisemat'  \Big).
\end{align*}
We bound each term in $\errormat_{\ndata}$ using Lemma~\ref{lm:mxconcentrationgau}. For the first term, since the entries in $\sigloadmat$ are $\iid$ standard Gaussian random variables, we have:
\begin{align*}
    \left\| \sig \left(\frac{1}{\ndata} \sigloadmat \sigloadmat^\top - I_{\nsig}\right) \sig^\top \right\|  \leq \left\| \sig \right\|_{2}^{2} \left\|  \frac{1}{\ndata} \sigloadmat \sigloadmat^\top - I_{\nsig}  \right\|_{2} \lesssim \eigval_{\sig, 1} \sqrt{\frac{\nsig \log(\ndata + \ndim)}{\ndata}},
\end{align*}
provided that $\ndata \gtrsim \nsig \log^3 (\ndata + \ndim)$, with probability at least $1 - \mathcal{O}((\ndata + \ndim)^{-10})$.
Applying similar arguments for the remaining terms, we obtain:
\begin{align*}
    \left\| \errormat_{\ndata} \right\| &\lesssim \eigval_{\sig, 1} \sqrt{\frac{\nsig \log(\ndata + \ndim)}{\ndata}} + \eigval_{\bkg, 1} \sqrt{\frac{\nbkg \log(\ndata + \ndim)}{\ndata}} + \sqrt{\eigval_{\sig, 1}\eigval_{\bkg, 1}} \sqrt{\frac{\max(\nsig, \nbkg)  \log(\ndata + \ndim)}{\ndata}}\\
    & + \sqrt{\eigval_{\sig, 1}} \sqrt{\frac{\ndim  \log(\ndata + \ndim)}{\ndata}} + \sqrt{\eigval_{\bkg, 1}} \sqrt{\frac{\ndim  \log(\ndata + \ndim)}{\ndata}} + \sqrt{\frac{\ndim \log(\ndata + \ndim)}{\ndata}} + \frac{\ndim \log^{2}(\ndata + \ndim)}{\ndata}\\
    &= \Bigg( \eigval_{\sig, 1} \sqrt{\frac{\nsig}{\ndata}} + \eigval_{\bkg, 1} \sqrt{\frac{\nbkg}{\ndata}} + \sqrt{\eigval_{\sig, 1}\eigval_{\bkg, 1}} \sqrt{\frac{\max(\nsig, \nbkg)}{\ndata}}\\
    &\quad \quad \quad + \left(\sqrt{\eigval_{\sig, 1}} + \sqrt{\eigval_{\bkg, 1}} + 1\right)\sqrt{\frac{\ndim}{\ndata}} + \frac{\ndim \log^{3/2}(\ndata + \ndim)}{\ndata} \Bigg) \log^{1/2}(\ndata + \ndim).
\end{align*}
\paragraph{Sufficient condition for $\ndata$.}
To satisfy the bound $\|\errormat_{\ndata}\|_2 \leq \left(1 - \frac{1}{\sqrt{2}}\right) | \eigval_{\sig, \nsig} |$, it is sufficient to assume:
\begin{align*}
    \ndata &\geq C \Bigg( \frac{1}{\eigval_{\sig, \nsig}^{2}} \left( \nsig \eigval_{\sig, 1}^{2} + \nbkg \eigval_{\bkg, 1}^{2} + \max(\nsig, \nbkg) \eigval_{\sig, 1}\eigval_{\bkg, 1} + \ndim \left(\eigval_{\sig, 1} + \eigval_{\bkg, 1} + 1\right) \right) \log (\ndata + \ndim)\\
    &\quad \quad\quad\quad\quad+ \frac{\ndim \log^2 (\ndata + \ndim)}{\eigval_{\sig, \nsig}} + \nsig \log^3 (\ndata + \ndim) \Bigg) ,
\end{align*}
for some large enough constant $C > 0$. The condition in the theorem is established by observing that:
\begin{equation*}
    \frac{\ndim \log^2 (\ndata + \ndim)}{\eigval_{\sig, \nsig}} \leq \frac{\ndim \eigval_{\sig, 1}\log^2 (\ndata + \ndim)}{\eigval_{\sig, \nsig}^2}.
\end{equation*} This allows us to eliminate the second-to-last term in the above expression, and by similar reasoning, the last term can also be omitted. 

\paragraph{Final bound on subspace distance.}
Using Lemma~\ref{lm:sintheta}, the distance between the estimated subspace \(\widehat{\eigspace}_\sig\) and the true subspace \(\eigspace_\sig\) satisfies:
\begin{equation*}
    \operatorname{dist}(\widehat{\eigspace}_{\sig}, \eigspace_{A})  \leq 
    \frac{2 \|\errormat_{\ndata}\|_{2}}{\eigval(\sig \sig^\top)_{\nsig} - \eigval(\sig \sig^\top)_{\nsig+1}} = \frac{2 \|\errormat_{\ndata}\|_{2}}{\eigval_{\sig, \nsig}}.
\end{equation*}
Substituting the bound on $\|\errormat_{\ndata}\|_2$, we obtain:
\begin{align*}
    \operatorname{dist}(\widehat{\eigspace}_{\sig}, \eigspace_{A})
    &\lesssim \frac{1}{\lambda_{A, k}} \Bigg( \eigval_{\sig, 1} \sqrt{\frac{\nsig}{\ndata}} + \eigval_{\bkg, 1} \sqrt{\frac{\nbkg}{\ndata}} + \sqrt{\eigval_{\sig, 1}\eigval_{\bkg, 1}} \sqrt{\frac{\max(\nsig, \nbkg)}{\ndata}}\\
    &+ \left(\sqrt{\eigval_{\sig, 1}} + \sqrt{\eigval_{\bkg, 1}} + 1\right)\sqrt{\frac{\ndim}{\ndata}} + \frac{\ndim \log^{3/2}(\ndata + \ndim)}{\ndata} \Bigg) \log^{1/2}(\ndata + \ndim)
\end{align*}
where the last term in the parenthesis, $\frac{\ndim \log^{3/2}(\ndata + \ndim)}{\ndata}$, is dominated by the second-to-last term $\sqrt{\frac{\eigval_{\sig, 1} \ndim}{\ndata}}$, under the assumption: 
\begin{equation*}
    \ndata \gtrsim \frac{\ndim \lambda_{A, 1}}{\lambda_{A, k}^2} \log^{3}(\ndata + \ndim) \geq \frac{\ndim }{\lambda_{A, 1}} \log^{3}(\ndata + \ndim).
\end{equation*}
Therefore, the last term in the parenthesis can be omitted, and the final bound becomes:
\begin{align*}
    \operatorname{dist}(\widehat{\eigspace}_{\sig}, \eigspace_{A})
    &\lesssim \frac{1}{\lambda_{A, k}} \Bigg( \eigval_{\sig, 1} \sqrt{\frac{\nsig}{\ndata}} + \eigval_{\bkg, 1} \sqrt{\frac{\nbkg}{\ndata}} + \sqrt{\eigval_{\sig, 1}\eigval_{\bkg, 1}} \sqrt{\frac{\max(\nsig, \nbkg)}{\ndata}}\\
    &+ \left(\sqrt{\eigval_{\sig, 1}} + \sqrt{\eigval_{\bkg, 1}} + 1\right)\sqrt{\frac{\ndim}{\ndata}} \Bigg) \log^{1/2}(\ndata + \ndim).
\end{align*}
Thus, the theorem is proved.
\end{proof}

\subsection{Proof of Theorem~\ref{thm:counterexample}}
\begin{proof}
We aim to show that
\begin{equation*}
    \lim_{\npinfty} (\hat{\eigvec}_{1}^\top e_{1})^2 \leq 2 \frac{\eigval_{\sig, 1}}{\sqrt{\eigval_{\bkg, 1} \aspratio}}.
\end{equation*}
Let $\hat{\eigval}_{1}$ be the largest eigenvalue of $\covmatsample_{\ndata}^{+}$. Then
\begin{equation*}
    (\hat{\eigvec}_{1}^\top e_{1})^2 \leq \frac{e_{1}^\top \covmatsample_{\ndata}^{+} e_{1}}{\hat{\eigval}_{1}}.
\end{equation*}
Hence, it suffices to establish the following two statements as $\npinfty$ with $\ndim/\ndata \rightarrow \aspratio \in (0, +\infty)$
\begin{enumerate}
    \item $\lim_{\npinfty} e_{1}^\top \covmatsample_{\ndata}^{+} e_{1} = \eigval_{\sig, 1}$,

    \item  $\lim_{\npinfty} \hat{\eigval}_{1} \geq \frac{\sqrt{\eigval_{\bkg, 1}\aspratio}}{2}$.
\end{enumerate}
Once these are established, the desired bound follows immediately.
\paragraph{Step 1:} Proof of $\lim_{\npinfty} e_{1}^\top \covmatsample_{\ndata}^{+} e_{1} = \eigval_{\sig, 1}$ 
Using the identity~\eqref{eq:eqidt}, we aim to demonstrate instead:
\begin{equation*}
    \lim_{\npinfty} e_{1}^\top \frac{1}{\ndata} X^\top X^{+} e_{1} = \eigval_{\sig, 1}
\end{equation*}
Recall the decomposition~\eqref{eq:thmeq1}:
\begin{equation*}
    \begin{split}
        X^\top X^{+} - \ndata \sig \sig^\top =& \ndata \errormat_{\ndata}\\
        =& \sig(\sigloadmat \sigloadmat^\top - \ndata I_{\nsig})\sig^\top + \sig \sigloadmat \bkgloadmat'^\top \bkg^\top + \sig \sigloadmat \noisemat' \\
        &+ \bkg \bkgloadmat \sigloadmat^\top \sig^\top + \bkg \bkgloadmat \bkgloadmat'^\top \bkg^\top  + \bkg \bkgloadmat \noisemat'\\
        &+ \noisemat^\top \sigloadmat^\top \sig^\top + \noisemat^\top \bkgloadmat'^\top \bkg^\top + \noisemat^\top \noisemat'.
    \end{split}
\end{equation*}
Following a similar approach as in the Step 1 proof of Theorem~\ref{lm:highdim_bound}, we analyze each term in $M$ in $\errormat_{\ndata}$. For any matrix $M$ in $\errormat_{\ndata}$ we demonstrate:
\begin{equation*}
    \frac{1}{\ndata} e_{1}^\top M e_{1} \overlineninfty 0 \qas,
\end{equation*}
with the exception of $\noisemat^\top \sigloadmat^\top \sig^\top$, which is analyzed separately.
\begin{itemize}
    \item[Case 1:] $M = \ndata \sig \sig^\top$
    
    For this case, we compute: 
    \begin{align*}
        e_{1}^\top \sig \sig^\top e_{1} = \|  \sig^\top e_{1} \|_{2}^{2} = \left\|  \sqrt{\eigval_{\sig, 1}} e_{1}^\top e_{1} \right\|_{2}^{2}
        = \eigval_{\sig, 1}
    \end{align*}
    
    \item[Case 2:] $M = \noisemat^\top \sigloadmat^\top \sig^\top$
    
    Here, we write
    \begin{align*}
        \frac{1}{\ndata}e_{1}^\top \noisemat^\top \sigloadmat^\top \sig^\top e_{1} = \frac{\sqrt{\eigval_{\sig, 1}}}{\ndata} \sigloadmat \noisemat e_{1}
        = \frac{\sqrt{\eigval_{\sig, 1}}}{\ndata} \sum_{j = 1}^{\ndata} \sigload_{j} \noise_{j, 1}
        \overlineninfty 0 \qas,
    \end{align*}
    where $\noise_{j, 1}$ represents the first entry of the noise vector $\noise_{j}$. The convergence to zero follows because $\sigload_{j}$ and $\noise_{j, 1}$ are independent standard Gaussian random variables and the strong law of large numbers.
    
\end{itemize}
This completes the proof of Step 1.

\paragraph{Step 2:} Proof of $\lim_{\npinfty} \hat{\eigval}_{1} \geq \frac{\sqrt{\eigval_{\bkg, 1}\aspratio}}{2}$

Let 
\begin{equation*}
    v =  \frac{1}{\sqrt{2}}e_{2} + \frac{1}{\sqrt{2}} \frac{\noisemat^\top \bkgloadmat'^{\top}}{\| \noisemat^\top \bkgloadmat'^{\top}\|_{2}}.
\end{equation*}
It is sufficient to show that 
\begin{enumerate}
    \item $\lim_{\npinfty} v^\top \covmatsample_{\ndata}^{+} v \geq \frac{\sqrt{\eigval_{\bkg, 1}\aspratio}}{2}$,

    \item $\lim_{\npinfty} \|v\|_{2} = 1$.
\end{enumerate}

Using the identity~\eqref{eq:eqidt} and the decomposition~\eqref{eq:thmeq1}, we proceed by proving that, for each term $M$ in $\ndata \sig \sig^\top$ and $\errormat_{\ndata}$ the following holds:
\begin{equation*}
    \frac{1}{\ndata} v^\top M v \overlineninfty 0 \qas,
\end{equation*}
with the exception of $\noisemat^\top \bkgloadmat'^\top \bkg^\top$, which satisfies
\begin{equation}\label{eq:counterexmpleeq1}
    \lim_{\npinfty} \frac{1}{\ndata} v^\top \noisemat^\top \bkgloadmat'^\top \bkg^\top v \geq \frac{\sqrt{\eigval_{\bkg, 1}\aspratio}}{2}.
\end{equation}
\begin{itemize}
    \item[Case 1:] $M = \sig \sig^\top$
    
    From the definition of $v$, we have
    \begin{align*}
        \sig^\top v &= \sqrt{\eigval_{\sig, 1}} e_{1}^\top v\\
        &= \sqrt{\frac{\eigval_{\sig, 1}}{2}} \frac{e_{1}^\top \noisemat^\top \bkgloadmat'^{\top} }{\| \noisemat^\top \bkgloadmat'^{\top}\|_{2}}\\
        &= \sqrt{\frac{\eigval_{\sig, 1}}{2}} \frac{1}{\| \noisemat^\top \bkgloadmat'^{\top}\|_{2}} \sum_{j = 1}^{\ndata} \bkgload'_{j} \noise_{j, 1}.
    \end{align*}
    The term $\| \noisemat^\top \bkgloadmat'^{\top}\|_{2}$ satisfies
    \begin{align*}
        \frac{1}{\ndata \ndim}\| \noisemat^\top \bkgloadmat'^{\top}\|_{2}^{2} = \frac{1}{\ndim} \sum_{i = 1}^{\ndim} \frac{1}{\ndata} \left(\sum_{j = 1}^{\ndata} \bkgload'_{j} \noise_{j, i} \right)^{2}.
    \end{align*}
    Since $\bkgload'_{j}$ and $\noise_{j, i}$ are independent standard Gaussian random variables, the sum $\frac{1}{\sqrt{\ndata}} \sum_{j = 1}^{\ndata} \bkgload'_{j} \noise_{j, i} $ follows a Gaussian distribution with mean zero and variance $\sum_{j = 1}^{\ndata} \frac{1}{\ndata} \bkgload'^{2}_{j}$.  By the strong law of large numbers, this variance converges to 1 almost surely as $\ndata \rightarrow +\infty$. Consequently, we have:
    \begin{equation*}
        \frac{1}{\ndata \ndim}\| \noisemat^\top \bkgloadmat'^{\top}\|_{2}^{2} \overlineninfty 1 \qas.
    \end{equation*}
    Hence,
    \begin{align*}
        v^\top \sig \sig^\top v &= \| \sig^\top v \|_{2}^{2}=\frac{\eigval_{\sig, 1}}{2} \frac{1}{\| \noisemat^\top \bkgloadmat'^{\top}\|_{2}^{2}} \left( \sum_{j = 1}^{\ndata} \bkgload'_{j} \noise_{j, 1} \right)^{2} \overlineninfty 0 \qas.
    \end{align*}

    \item[Case 2:]  $M =  \sig(\sigloadmat \sigloadmat^\top - \ndata I)\sig^\top$
    
    The arguments for this case are analogous to Case 2 in Step 1 of the proof of Theorem~\ref{lm:highdim_bound}. The same reasoning applies directly to the terms $\sig(\sigloadmat \sigloadmat^\top - \ndata I)\sig^\top$, $
    \frac{1}{\ndata}\hat{\eigvec}_{\bkg, j}^\top \sig \sigloadmat \bkgloadmat^\top \bkg^\top \hat{\eigvec}_{\bkg, j}$, $\frac{1}{\ndata}\hat{\eigvec}_{\bkg, j}^\top \bkg \bkgloadmat^{+} \bkgloadmat^\top \bkg^\top \hat{\eigvec}_{\bkg, j}$ and $\frac{1}{\ndata}\hat{\eigvec}_{\bkg, j}^\top \bkg \bkgloadmat^{+} \bkgloadmat^\top \bkg^\top \hat{\eigvec}_{\bkg, j}$.

   \item[Case 3:] $M = \noisemat^\top \noisemat'$
    
    Since $v$ is independent of $\noisemat'$, the same arguments used in Case 3 and Case 4 of Step 1 in the proof of Theorem~\ref{lm:highdim_bound} can be applied here. This reasoning extends to the terms $\noisemat^\top \noisemat'$, $\bkg \bkgloadmat \noisemat'$ and $\sig \sigloadmat \noisemat'$.

    \item[Case 4:] $M = \noisemat^\top \sigloadmat^\top \sig^\top$
    
    From the Case 1, we know that 
    \begin{equation*}
        \sig^\top v \overlineninfty 0 \qas.
    \end{equation*}
    The rest of the proof follows exactly the reasoning in Case 5 of Step 1 in the proof of Theorem~\ref{lm:highdim_bound}.

    \item[Case 5:] $M = \noisemat^\top \bkgloadmat'^\top \bkg^\top$
    
    We write:
    \begin{align*}
        \frac{1}{\ndata} v^\top \noisemat^\top \bkgloadmat'^\top \bkg^\top v = \frac{1}{\sqrt{2}n}e_{2}^\top \noisemat^\top \bkgloadmat'^\top \bkg^\top v + \frac{1}{\sqrt{2}n} \frac{(\noisemat^\top \bkgloadmat'^{\top})^\top}{\| \noisemat^\top \bkgloadmat'^{\top}\|_{2}} \noisemat^\top \bkgloadmat'^\top \bkg^\top v
    \end{align*}
    We will show that the first term converges to zero, and the second term satisfies the inequality in~\eqref{eq:counterexmpleeq1}. 
    
    Firstly, consider the term $\bkg^\top v$:
    \begin{align}
        \bkg^\top v &= \sqrt{\eigval_{\bkg, 1}} e_{2}^\top v \notag\\ 
        &= \sqrt{\frac{\eigval_{\bkg, 1}}{2}} \left(1 + \frac{e_{2}^\top \noisemat^\top \bkgloadmat'^{\top} }{\| \noisemat^\top \bkgloadmat'^{\top}\|_{2}} \right) \notag\\
        &= \sqrt{\frac{\eigval_{\bkg, 1}}{2}} \left(1 + \frac{1}{\| \noisemat^\top \bkgloadmat'^{\top}\|_{2}} \sum_{j = 1}^{\ndata} \bkgload'_{j} \noise_{j, 2} \right) \notag\\ \label{eq:thmcounterexmpleeq2}
        &\overlineninfty \sqrt{\frac{\eigval_{\bkg, 1}}{2}} \qas,
    \end{align}
    where the convergence follows from the same reasoning as in Case 1. Hence,
    \begin{align*}
        \frac{1}{\sqrt{2}n} e_{2}^\top \noisemat^\top \bkgloadmat'^\top \bkg^\top v &= \frac{1}{\sqrt{2}} \bkg^\top v \frac{1}{\ndata} \sum_{j = 1}^{\ndata} \bkgload'_{j} \noise_{j, 2} \overlineninfty 0 \qas.
    \end{align*}
    Now, for the second term:
    \begin{align*}
        \lim_{\npinfty} \frac{1}{\sqrt{2}n} \frac{(\noisemat^\top \bkgloadmat'^{\top})^\top}{\| \noisemat^\top \bkgloadmat'^{\top}\|_{2}} \noisemat^\top \bkgloadmat'^\top \bkg^\top v &= \lim_{\npinfty} \frac{\sqrt{\eigval_{\bkg, 1}}}{2n} \| \noisemat^\top \bkgloadmat'^{\top}\|_{2}= \frac{\sqrt{\eigval_{\bkg, 1}\aspratio}}{2}.  
    \end{align*}
\end{itemize}
It remains to demonstrate that $\lim_{\npinfty} \|v\|_{2} = 1$. We have
\begin{align*}
    \|v\|_{2}^{2} &= \left(\frac{1}{\sqrt{2}}e_{2} + \frac{1}{\sqrt{2}} \frac{\noisemat^\top \bkgloadmat'^{\top}}{\| \noisemat^\top \bkgloadmat'^{\top}\|_{2}} \right)^\top \left(\frac{1}{\sqrt{2}}e_{2} + \frac{1}{\sqrt{2}} \frac{\noisemat^\top \bkgloadmat'^{\top}}{\| \noisemat^\top \bkgloadmat'^{\top}\|_{2}} \right) \\
    &= 1 + \frac{e_{2}^\top \noisemat^\top \bkgloadmat'^{\top} }{\| \noisemat^\top \bkgloadmat'^{\top}\|_{2}} \overlineninfty 1 \qas,
\end{align*}
where the convergence follows from the same arguments as in equation~\eqref{eq:thmcounterexmpleeq2}. 

This completes the proof of Step 2.
\end{proof}

\subsection{Proof of Theorem~\ref{thm:dist}}
\begin{proof}
    We begin by considering the sample covariance matrix $S_{\ndata}$ and write its eigendecomposition as
    \begin{equation*}
        S_{n} V_{x} = V_{x} \Lambda_{x},
    \end{equation*}
    where we order $V_{x} = (\hat{\eigvec}_{\sig,1}, \ldots, \hat{\eigvec}_{\sig, \nsig}, \hat{\eigvec}_{\bkg,1}, \ldots, \hat{\eigvec}_{\bkg, \nbkg}, \ldots)$ and $\Lambda_{x} = (\hat{\eigval}_{\sig,1}, \ldots, \hat{\eigval}_{\sig, \nsig}, \hat{\eigval}_{\bkg,1}, \ldots, \hat{\eigval}_{\bkg, \nbkg}, 0, \ldots)$. Set \(\Lambda_{x}^{-1/2}\) to be the diagonal matrix whose entries are
\(\hat{\eigval}_{k}^{-1/2}\) when \(\hat{\eigval}_{k}>0\) and \(0\) otherwise, and define the
whitened matrix. The transformed matrix $M_{n}$ can thus be expressed as:
    \begin{equation*}
        M = \Lambda_{x}^{-1/2} V_{x}^\top \covmatsample_{n}^+ V_{x} \Lambda_{x}^{-1/2}.
    \end{equation*}
    The matrix is nonzero only within its upper-left $(\nsig + \nbkg) \times (\nsig + \nbkg)$ submatrix:
    \begin{align*}
        & M_{n, (\nsig + \nbkg) \times (\nsig + \nbkg)} \\ 
        = & \left(\begin{array}{cccccc}
\frac{1}{\hat{\eigval}_{\sig,1}} \hat{\eigvec}_{\sig,1}^\top \covmatsample_{n}^+ \hat{\eigvec}_{\sig,1} & \cdots & \frac{1}{\sqrt{\hat{\eigval}_{\sig,1} \hat{\eigval}_{\sig, \nsig}}} \hat{\eigvec}_{\sig,1}^\top \covmatsample_{n}^+ \hat{\eigvec}_{\sig, \nsig} & \frac{1}{\sqrt{\hat{\eigval}_{\sig,1} \hat{\eigval}_{\bkg, 1}}} \hat{\eigvec}_{\sig,1}^\top \covmatsample_{n}^+ \hat{\eigvec}_{\bkg, 1} & \cdots &  \frac{1}{\sqrt{\hat{\eigval}_{\sig,1} \hat{\eigval}_{\bkg, \nbkg}}} \hat{\eigvec}_{\sig,1}^\top \covmatsample_{n}^+ \hat{\eigvec}_{\bkg, \nbkg}\\
\vdots && \vdots & \vdots && \vdots\\
\frac{1}{\sqrt{\hat{\eigval}_{\sig,\nsig} \hat{\eigval}_{\sig, 1}}} \hat{\eigvec}_{\sig, \nsig}^\top \covmatsample_{n}^+ \hat{\eigvec}_{\sig, 1} & \cdots & \frac{1}{\hat{\eigval}_{\sig, \nsig} } \hat{\eigvec}_{\sig,\nsig}^\top \covmatsample_{n}^+ \hat{\eigvec}_{\sig, \nsig} & \frac{1}{\sqrt{\hat{\eigval}_{\sig, \nsig} \hat{\eigval}_{\bkg, 1}}} \hat{\eigvec}_{\sig, \nsig}^\top \covmatsample_{n}^+ \hat{\eigvec}_{\bkg, 1} & \cdots &  \frac{1}{\sqrt{\hat{\eigval}_{\sig, \nsig} \hat{\eigval}_{\bkg, \nbkg}}} \hat{\eigvec}_{\sig, \nsig}^\top \covmatsample_{n}^+ \hat{\eigvec}_{\bkg, \nbkg} \\
\frac{1}{\sqrt{\hat{\eigval}_{\bkg,1} \hat{\eigval}_{\sig, 1}}} \hat{\eigvec}_{\bkg, 1}^\top \covmatsample_{n}^+ \hat{\eigvec}_{\sig, 1} & \cdots & \frac{1}{\sqrt{\hat{\eigval}_{\bkg, 1} \hat{\eigval}_{\sig, \nsig}}} \hat{\eigvec}_{\bkg, 1}^\top \covmatsample_{n}^+ \hat{\eigvec}_{\sig, \nsig} & \frac{1}{ \hat{\eigval}_{\bkg, 1}} \hat{\eigvec}_{\bkg, 1}^\top \covmatsample_{n}^+ \hat{\eigvec}_{\bkg, 1} & \cdots &  \frac{1}{\sqrt{\hat{\eigval}_{\bkg, 1} \hat{\eigval}_{\bkg, \nbkg}}} \hat{\eigvec}_{\bkg, 1}^\top \covmatsample_{n}^+ \hat{\eigvec}_{\bkg, \nbkg}\\
\vdots && \vdots & \vdots && \vdots\\
\frac{1}{\sqrt{\hat{\eigval}_{\bkg,\nbkg} \hat{\eigval}_{\sig, 1}}} \hat{\eigvec}_{\bkg, \nbkg}^\top \covmatsample_{n}^+ \hat{\eigvec}_{\sig, 1} & \cdots & \frac{1}{\sqrt{\hat{\eigval}_{\bkg, \nbkg} \hat{\eigval}_{\sig, \nsig}}} \hat{\eigvec}_{\bkg, \nbkg}^\top \covmatsample_{n}^+ \hat{\eigvec}_{\sig, \nsig} & \frac{1}{ \sqrt{ \hat{\eigval}_{\bkg, \nbkg}\hat{\eigval}_{\bkg, 1}}} \hat{\eigvec}_{\bkg, \nbkg}^\top \covmatsample_{n}^+ \hat{\eigvec}_{\bkg, 1} & \cdots &  \frac{1}{\hat{\eigval}_{\bkg, \nbkg}} \hat{\eigvec}_{\bkg, \nbkg}^\top \covmatsample_{n}^+ \hat{\eigvec}_{\bkg, \nbkg}
\end{array}\right).
    \end{align*}
We now show that, in the limit, this submatrix $M_{(\nsig + \nbkg) \times (\nsig + \nbkg)}$ vanishes except for the first $\nsig$ diagonal term entries, which remain positive by Lemma~\ref{lm:highdim_bound}.

\paragraph{Entries involving background directions.}
Using Lemma~\ref{lm:highdim_bound}, it follows directly that cross terms involving eigenvectors from the background subspace vanish asymptotically, for any \(1\le i,j\le\nbkg\),
\[
   \frac{1}{\sqrt{\hat{\eigval}_{\bkg,i}\hat{\eigval}_{\bkg,j}}}\,
   \hat{\eigvec}_{\bkg,i}^{\top}\covmatsample_{n}^{+}\hat{\eigvec}_{\bkg,j}
   \leq \frac{1}{\hat{\eigval}_{\bkg, i}} \hat{\eigvec}_{\bkg, i}^\top \covmatsample_{n}^+ \hat{\eigvec}_{\bkg, i} \frac{1}{\hat{\eigval}_{\bkg, j}} \hat{\eigvec}_{\bkg, j}^\top \covmatsample_{n}^+ \hat{\eigvec}_{\bkg, j} \;\xrightarrow{a.s.}\;0,
\]
and similarly, for \(1\le i\le\nsig,\;1\le j\le\nbkg\),
\[
   \frac{1}{\sqrt{\hat{\eigval}_{\sig,i}\hat{\eigval}_{\bkg,j}}}\,
   \hat{\eigvec}_{\sig,i}^{\top}\covmatsample_{n}^{+}\hat{\eigvec}_{\bkg,j}
   \leq  \frac{1}{\hat{\eigval}_{\sig, i}} \hat{\eigvec}_{\sig, i}^\top \covmatsample_{n}^+ \hat{\eigvec}_{\sig, i} \frac{1}{\hat{\eigval}_{\bkg, j}} \hat{\eigvec}_{\bkg, j}^\top \covmatsample_{n}^+ \hat{\eigvec}_{\bkg, j} \;\xrightarrow{a.s.}\;0.
\]

% Using Lemma~\ref{lm:highdim_bound}, we get
% \begin{align*}
%     \lim_{\npinfty} \frac{1}{ \sqrt{ \hat{\eigval}_{\bkg, i}\hat{\eigval}_{\bkg, j}}} \hat{\eigvec}_{\bkg, i}^\top \covmatsample_{n}^+ \hat{\eigvec}_{\bkg, j} &\leq \lim_{\npinfty} \frac{1}{\hat{\eigval}_{\bkg, i}} \hat{\eigvec}_{\bkg, i}^\top \covmatsample_{n}^+ \hat{\eigvec}_{\bkg, i} \frac{1}{\hat{\eigval}_{\bkg, j}} \hat{\eigvec}_{\bkg, j}^\top \covmatsample_{n}^+ \hat{\eigvec}_{\bkg, j} \overlineninfty 0 \qas, \\
%     \lim_{\npinfty} \frac{1}{ \sqrt{ \hat{\eigval}_{\sig, i}\hat{\eigval}_{\bkg, j}}} \hat{\eigvec}_{\sig, i}^\top \covmatsample_{n}^+ \hat{\eigvec}_{\bkg, j} &\leq \lim_{\npinfty} \frac{1}{\hat{\eigval}_{\sig, i}} \hat{\eigvec}_{\sig, i}^\top \covmatsample_{n}^+ \hat{\eigvec}_{\sig, i} \frac{1}{\hat{\eigval}_{\bkg, j}} \hat{\eigvec}_{\bkg, j}^\top \covmatsample_{n}^+ \hat{\eigvec}_{\bkg, j} \overlineninfty 0 \qas.
% \end{align*}

\paragraph{Off‑diagonal signal-signal entries.}
Fix \(1\le i<j\le\nsig\). A similar argument as in the derivation of equation~\eqref{eq:eq1} shows that
\begin{align*}
       \lim_{\npinfty} \frac{1}{\sqrt{\hat{\eigval}_{\sig,i} \hat{\eigval}_{\sig, j}}} \hat{\eigvec}_{\sig, i}^\top \covmatsample_{n}^+ \hat{\eigvec}_{\sig, j} &= \underbrace{%
       \lim_{\npinfty} \frac{1}{\sqrt{\hat{\eigval}_{\sig,i} \hat{\eigval}_{\sig, j}}} \hat{\eigvec}_{\sig, i}^\top \sig \sig^\top \hat{\eigvec}_{\sig, j}}_{=:T_{1}}\\
     &\;+\;
     \underbrace{%
       \lim_{\npinfty} \frac{1}{\sqrt{\hat{\eigval}_{\sig,i} \hat{\eigval}_{\sig, j}}} \frac{1}{2\ndata} \hat{\eigvec}_{\sig, i}^\top (\sig \sigloadmat \noisemat + \noisemat^\top \sigloadmat^\top \sig^\top)\hat{\eigvec}_{\sig, j}}_{=:T_{2}}.
\end{align*}

\begin{enumerate}
\item $T_{1}$: We have
        \begin{equation*}
    \lim_{\npinfty} \frac{1}{\sqrt{\hat{\eigval}_{\sig,i} \hat{\eigval}_{\sig, j}}} \hat{\eigvec}_{\sig, i}^\top \sig \sig^\top \hat{\eigvec}_{\sig, j} = \lim_{\npinfty} \frac{1}{\sqrt{\hat{\eigval}_{\sig,i} \hat{\eigval}_{\sig, j}}} \sum_{l = 1}^{\nsig} \hat{\eigval}_{\sig, l} \hat{\eigvec}_{\sig, i}^\top \eigvec_{\sig, l} \eigvec_{\sig, l}^\top \hat{\eigvec}_{\sig, j} = 0,
    \end{equation*}
    where the last equality follows directly from Lemma~\ref{lm:HDspike}.
    
        \item $T_{2}$:
        Similar reasoning to equation~\eqref{eq:eq2} yields:
        \begin{align*}
    \lim_{\npinfty} \frac{1}{\sqrt{\hat{\eigval}_{\sig,i} \hat{\eigval}_{\sig, j}}} \frac{1}{\ndata} \hat{\eigvec}_{\sig, i}^\top (\sig \sigloadmat \noisemat + \noisemat^\top \sigloadmat^\top \sig^\top)\hat{\eigvec}_{\sig, j} &= \lim_{\npinfty} \frac{1}{\sqrt{\hat{\eigval}_{\sig,i} \hat{\eigval}_{\sig, j}}} \hat{\eigvec}_{\sig, i}^\top \frac{1}{\ndata} X^\top X \hat{\eigvec}_{\sig, j}\\ 
    - \lim_{\npinfty} \frac{1}{\sqrt{\hat{\eigval}_{\sig,i} \hat{\eigval}_{\sig, j}}} &\hat{\eigvec}_{\sig, i}^\top \sig \sig^\top \hat{\eigvec}_{\sig, j}  - \lim_{\npinfty} \frac{1}{\sqrt{\hat{\eigval}_{\sig,i} \hat{\eigval}_{\sig, j}}} \hat{\eigvec}_{\sig, i}^\top \frac{1}{\ndata} \noisemat^\top \noisemat \hat{\eigvec}_{\sig, j}\\
    = - \lim_{\npinfty} \frac{1}{\sqrt{\hat{\eigval}_{\sig,i} \hat{\eigval}_{\sig, j}}} &\hat{\eigvec}_{\sig, i}^\top \frac{1}{\ndata} \noisemat^\top \noisemat \hat{\eigvec}_{\sig, j}.
    \end{align*}

To demonstrate the almost sure convergence of the remaining expression to zero, we apply a leave-one-out decoupling approach:
\begin{itemize}
    \item For each sample $1 \leq l \leq \ndata$ delete row $l$ of $X$ and let $\hat{\eigval}_{i}^{-l}$ be the corresponding eigenvectors. These eigenvectors satisfy independence from the omitted data point $x_{l}$ and Davis-Kahan perturbation bounds implies $\| \hat{\eigval}_{i} - \hat{\eigval}_{i}^{-l}\|_{2} = O_{p}(\frac{1}{\ndata})$.
    \item Express the term as:
    \begin{equation*}
        \hat{\eigvec}_{\sig, i}^\top \frac{1}{\ndata} \noisemat^\top \noisemat \hat{\eigvec}_{\sig, j} = \frac{1}{\ndata} \sum_{l = 1}^{\ndata} (z_{l}^\top \hat{\eigvec}_{\sig, i}^{(-l)})^\top (z_{l}^\top \hat{\eigvec}_{\sig, j}^{(-l)}) + o_{p}(1/\sqrt{n}),
    \end{equation*}
    where $z_{l}$ the $l$-th row of $\noisemat$, is independent of $\hat{\eigvec}_{\sig, i}^{(-l)}$ and $\hat{\eigvec}_{\sig, j}^{(-l)}$.
    \item  Following the proof of Lemma~\ref{lm:orthGauss}, $\sum_{l = 1}^{\ndata} (z_{l}^\top \hat{\eigvec}_{\sig, i}^{(-l)})^\top (z_{l}^\top \hat{\eigvec}_{\sig, j}^{(-l)})/\ndata$ is an average of $\iid$ zero‑mean variables with finite variance, hence it converges to zero almost surely by the strong law of large numbers.
\end{itemize}
\end{enumerate}
Consequently, we establish:
\begin{equation*}
    \lim_{\npinfty} \frac{1}{\sqrt{\hat{\eigval}_{\sig,i} \hat{\eigval}_{\sig, j}}} \frac{1}{\ndata} \hat{\eigvec}_{\sig, i}^\top (\sig \sigloadmat \noisemat + \noisemat^\top \sigloadmat^\top \sig^\top)\hat{\eigvec}_{\sig, j} = - \lim_{\npinfty} \frac{1}{\sqrt{\hat{\eigval}_{\sig,i} \hat{\eigval}_{\sig, j}}} \hat{\eigvec}_{\sig, i}^\top \frac{1}{\ndata} \noisemat^\top \noisemat \hat{\eigvec}_{\sig, j} = 0,
\end{equation*}
proving that the off-diagonal entries vanish asymptotically.

\paragraph{Positive definite submatrix $M_{n, \nsig \times \nsig}$}
We have demonstrated that the asymptotic limit of the matrix $M_{n, (\nsig + \nbkg) \times (\nsig + \nbkg) }$ except for its first $\nsig$ diagonal term, which remain strictly positive by Lemma~\ref{lm:highdim_bound}. Consequently, asymptotically, exactly $\nsig$ sample eigenvectors correspond to positive eigenvalues, yielding the estimated signal set:
\begin{equation*}
    V = (\hat{\eigvec}_{\sig,1}, \ldots, \hat{\eigvec}_{\sig, \nsig}).
\end{equation*}

\paragraph{Evaluate the estimated subspace}
Finally, we quantify the accuracy of the estimated subspace $\widehat{\eigspace}_{\sig}$ relative to the true signal subspace $\mathcal{U}_{A}$. By Lemma~\ref{lm:HDspike}, it follows that:
\begin{align*}
    \lim_{\npinfty} V^\top \sig &= \lim_{\npinfty} \left(\begin{array}{ccc}
\hat{\eigvec}_{\sig,1}^\top \eigvec_{\sig,1} & \cdots & \hat{\eigvec}_{\sig,1}^\top \eigvec_{\sig, \nsig} \\
\vdots && \vdots\\
 \hat{\eigvec}_{\sig, \nsig}^\top  \eigvec_{\sig, 1} & \cdots & \hat{\eigvec}_{\sig,\nsig}^\top \eigvec_{\sig, \nsig}
\end{array}\right) \\
&= \diag \left(
\sqrt{\frac{1 - \aspratio \eigval_{1}^{-2}}{1 + \aspratio \eigval_{1}^{-1}}}, \ldots, \sqrt{\frac{1 - \aspratio \eigval_{\nsig}^{-2}}{1 + \aspratio \eigval_{\nsig}^{-1}}} \right)
\end{align*}
Hence, the asymptotic squared subspace distance is given by:
\begin{align*}
    \lim_{\npinfty} \operatorname{dist}(\widehat{\eigspace}_{\sig}, \mathcal{U}_{A})^2 &= \lim_{\npinfty}\max_{1 \leq j \leq \nsig} \sin^{2} (\arccos \sigma_{j}) \\
    & =\lim_{\npinfty}\max_{1 \leq j \leq \nsig} 1 - \sigma_{j}^2\\
    & = 1 - \frac{1 - \aspratio \eigval_{\nsig}^{-2}}{1 + \aspratio \eigval_{\nsig}^{-1}}.
\end{align*}
% Moreover, let $\check{\eigvec} = \left(
% \frac{1}{\sqrt{\eigval_{\sig,1}}} \eigvec_{\sig,1}, \ldots,
% \frac{1}{\sqrt{\eigval_{\sig,\nsig}}} \eigvec_{\sig,\nsig} \right)$, by equation~\eqref{eq:eq1}, the $\nsig  \times \nsig$ submatrix on the left up corner of the matrix $\check{A}$ satisfies
% \begin{equation*}
%     \lim_{\npinfty} \check{A}_{\nsig \times \nsig} = \lim_{\npinfty}\check{\eigvec}^\top \sig \sig^\top \check{\eigvec} + \lim_{\npinfty}\check{\eigvec}^\top (\sig \sigloadmat + \noisemat^\top) (\sig \sigloadmat + \noisemat^\top)^\top \check{\eigvec} - \lim_{\npinfty} \check{\eigvec}^\top \noisemat^\top \noisemat \check{\eigvec},
% \end{equation*}
% where the second term is positive semi-definit. 

% Next, we show $\check{A}_{\nsig \times \nsig}$ is asymptotically strictly positive definite. It is sufficient to see that
% \begin{align*}
%     \eigval_{\min} \left(\sig \sig^\top - \noisemat^\top \noisemat \right) &\geq \eigval_{\min} \left(\sig \sig^\top\right) - \eigval_{\max} (\noisemat^\top \noisemat)\\
%     & \geq 
% \end{align*}
\end{proof}

\subsection{Proof of Theorem~\ref{thm:dist2}}
\begin{proof}
The proof follows directly from the arguments detailed in the proof of Theorem~\ref{thm:dist}, in conjunction with Lemma~\ref{lm:HDspike2} and Lemma~\ref{lm:highdim_bound2}. Accordingly, the algorithm asymptotically yields:
\begin{equation*}
    V = (\hat{\eigvec}_{\sig,1}, \ldots, \hat{\eigvec}_{\sig, \nsig}).
\end{equation*}
Consequently, we have:
\begin{equation*}
    \lim_{\npinfty} V^\top \sig = \diag\left(
(1 + \aspratio_{\sig, 1})^{-1/2} (1 + \aspratio_{\sig, \nsig})^{-1/2} \right)
\end{equation*}
and the squared subspace distance asymptotically satisfies:
\begin{align*}
    \lim_{\npinfty} \operatorname{dist}(\widehat{\eigspace}_{\sig}, \mathcal{U}_{A})^2 &= \lim_{\npinfty}\max_{1 \leq j \leq \nsig} \sin^{2} (\arccos \sigma_{j}) \\
    & =\lim_{\npinfty} \max_{1 \leq j \leq \nsig} 1 - \sigma_{j}^2\\
    & = 1 - \frac{1}{1 + \aspratio_{\sig, \nsig}}\\
    & = \frac{\aspratio_{\sig, \nsig}}{1 + \aspratio_{\sig, \nsig}}.
\end{align*}    
\end{proof}

\section{Preliminary Lemmas}\label{sec:prelemma}

This section collects technical tools that will be invoked repeatedly in the proofs of our main results.

\begin{lemma}[Lemma 3.5 in~\cite{chen2021spectral}]\label{lm:mxconcentrationgau}
    Assume that the entries in matrices $\sigloadmat \in \RR^{\nsig \times \ndata}$ and $Z \in \RR^{\ndim \times \ndata}$ are $\iid$ standard Gaussian random variables, if $\ndata \gtrsim \nsig \log^3 (\ndata + \ndim)$, with probability at least $1 - \mathcal{O}((\ndata + \ndim)^{-10})$ one has
    \begin{align*}
        \begin{split}
           \frac{1}{\ndata} \left\| \sigloadmat Z^\top \right\| &\lesssim \sqrt{\frac{\ndim \log(\ndata + \ndim)}{\ndata}},\\
           \left\| \frac{1}{\ndata}\sigloadmat \sigloadmat^\top - I_{\nsig} \right\| &\lesssim \sqrt{\frac{\nsig \log(\ndata + \ndim)}{\ndata}},\\
           \left\| \frac{1}{\ndata}Z Z^\top - I_{\ndim} \right\| &\lesssim \sqrt{\frac{\ndim \log(\ndata + \ndim)}{\ndata}} + \frac{\ndim \log^{2}(\ndata + \ndim)}{\ndata}.
        \end{split}
    \end{align*}
\end{lemma}
The next lemma is a convenient corollary of the classical Davis-Kahan $\sin\Theta$ theorem.
\begin{lemma}[Corollary 2.8 in \cite{chen2021spectral}]\label{lm:sintheta}
    Let $M^{*}$ and $M=M^{*}+E$ be two $n \times n$ real symmetric matrices. We express the eigendecomposition of $M^{*}$ and $M$ as follows
\begin{align*}
M^{*} & =\sum_{j=1}^n \lambda_j^{*} v_j^{*} v_j^{* \top}
=\begin{pmatrix}
U^* & U_\perp^*
\end{pmatrix} \begin{pmatrix}
\Lambda^* & 0 \\
0 & \Lambda_\perp^*
\end{pmatrix} \begin{pmatrix}
{U^*}^\top \\
{U_\perp^*}^\top
\end{pmatrix}.\end{align*}
Here, $\{\lambda_j\}_{j=1}^{\ndim}$ (resp. $\{\lambda_j^{*}\}_{j=1}^{\ndim}$) denote the eigenvalues of $M$ (resp. $M^{*}$), and $v_j$ (resp. $v_j^{*}$) stands for the eigenvector associated with the eigenvalue $\lambda_j$ (resp. $\lambda_j^{*}$). Additionally, we take
\begin{equation*}
\Lambda \coloneqq\operatorname{diag}\left(\left[\lambda_1, \cdots, \lambda_r\right]\right) \in \RR^{r \times r}, \quad \quad U\coloneqq\left[v_1, \cdots, v_r\right] \in \mathbb{R}^{n \times r}.
\end{equation*}
The matrices $U^{*}, U_{\perp}^{*}, \Lambda^{*}$, and $\Lambda_{\perp}^{*}$ are defined analogously. Suppose that $|\lambda_1^{*}| \geq|\lambda_2^{*}| \geq \cdots \geq|\lambda_r^{*}|>$ $|\lambda_{r+1}^{*}\| \geq \cdots |\lambda_n^{*}|$ and $|\lambda_1| \geq|\lambda_2| \geq \cdots \geq|\lambda_n|$ (i.e., the eigenvalues are sorted by their magnitudes). If $\|E\|<(1-1 / \sqrt{2})(|\lambda_r^{*}|-|\lambda_{r+1}^{*}|)$, then
\begin{equation*}
    \operatorname{dist}(\mathcal{U}, \mathcal{U}^{*}) \leq \sqrt{2}\|\sin \Theta\| \leq \frac{2\|E U^{*}\|}{|\lambda_r^{*}\|-|\lambda_{r+1}^{*}\|} \leq \frac{2\|E\|}{|\lambda_r^{*}|-|\lambda_{r+1}^{*}|},
\end{equation*}
where $\mathcal{U}$ (resp. $\mathcal{U}^{*}$) is the subspace spanned by $U$ (resp. $U^{*}$).

\end{lemma}

By combining Theorem 2 and Remark 1 of \cite{Bai1993}, we obtain the following lemma. 
\begin{lemma}\label{lm:Bai-yin}
    Assume that the feature vector $x$ has \iid~entries with zero mean, unit variance, and bounded $4$-th moment. As $\ndata, \ndim \rightarrow \infty$, $0 \leq \aspratio<\infty$,
    \begin{align*}
        \lim_{\npinfty} \eigval_{\min}^{+} (X^\top X/\ndata) &= (1 - \sqrt{\aspratio})^{2} \qas \\
        \lim_{\npinfty} \eigval_{\max} (X^\top X/\ndata) &= (1 + \sqrt{\aspratio})^{2} \qas
    \end{align*}
    where $\eigval_{\min}^{+}$ denotes the smallest positive eigenvalue.
\end{lemma}

\begin{lemma}[\citep{baik2006eigenvalues} and \citep{paul2007asymptotics}]\label{lm:HDspike}
Under model~\eqref{eq:model}, let $\eigval_{1} > \eigval_{2} > \ldots > \eigval_{\nsig + \nbkg}> \sqrt{\aspratio} \in (0, + \infty)$ denote the top $\nsig + \nbkg$ eigenvalues. Then, as $\ndim/\ndata \rightarrow \aspratio$ with $\npinfty$,
\begin{equation*}
\begin{split}
    \hat{\eigval}_{j} &\rightarrow 1 + \eigval_{j} + \aspratio \frac{1 + \eigval_{j}}{\eigval_{j}} \qas,\\
    a^\top \hat{\eigvec}_{j} \hat{\eigvec}_{j}^\top b &\rightarrow  \frac{1 - \aspratio \eigval_{j}^{-2}}{1 + \aspratio \eigval_{j}^{-1}} a^\top \eigvec_{j} \eigvec_{j}^\top b \qas,
\end{split}
\end{equation*}
for any deterministic unit vectors $a,b \in \RR^{\ndim}$ and $1 \leq j\leq \nsig + \nbkg$.
\end{lemma}

\begin{lemma}[Theorem 5.1 in \citep{shen2016statistics}]\label{lm:HDspike2}
Under model~\eqref{eq:model}, let $\eigval_{1} > \eigval_{2} > \ldots > \eigval_{\nsig + \nbkg}$ denote the top $\nsig + \nbkg$ eigenvalues. Then, as $\npinfty$ with $\frac{\ndim}{\ndata \eigval_{j}}  \rightarrow \aspratio_{j}\in (0, + \infty)$ and $\eigval_{\nsig + \nbkg} \rightarrow \infty$,
\begin{equation*}
\begin{split}
    \hat{\eigval}_{j}/\eigval_{j} &\rightarrow (1 + \aspratio_{j}) \qas,\\
    \eigvec_{i}^\top \hat{\eigvec}_{j}  &\rightarrow  (1 + \aspratio_{j})^{-1/2} \mathbbm{1}(j=i) \qas,
\end{split}
\end{equation*}
for $1 \leq i,j \leq \nsig + \nbkg$.
\end{lemma}
\begin{proof}
    The original statement of Theorem 5.1 in \citep{shen2016statistics} covers explicitly the results of both the sample eigenvalues and the alignment between the $j$-th sample and population eigenvectors when $i = j$. Here, we extend the arguments to handle the off-diagonal scenario $i \neq j$. Let
    \begin{equation*}
        \hat{\eigvec}_{j} = (\hat{\eigvec}_{1,j}, \dots, \hat{\eigvec}_{\ndim,j})^\top, \quad j = 1, \dots, \ndim
    \end{equation*}
    denote the eigenvectors of the sample covariance. Without loss of generality, we adopt the standard Euclidean basis $\{e_j\}_{j=1}^{\ndim}$ for the population eigenvectors, setting $\eigvec_{j} = e_{j}$. Under this choice, it suffices to demonstrate that, as $\npinfty$,
    \begin{equation*}
        \hat{\eigvec}_{i, j} \rightarrow 0 \qas,
    \end{equation*} 
    for all $i \neq j$. We first consider the scenario with $j = 1$. By Theorem 5.1 (equation (8.21)) of \citep{shen2016statistics}, it follows directly that:
    \begin{equation*}
        \sum_{i = 2}^{\nsig + \nbkg} \hat{\eigvec}_{i, 1}^{2} \overlineninfty 0 \qas.
    \end{equation*}
    Hence, we conclude $\hat{\eigvec}_{i, 1}\rightarrow 0$ a.s. for all $i\neq 1$, proving the claim for $j=1$. 
    
    Next, for the case, $j=2$, equations (8.23) and (8.24) in Theorem 5.1 from \citep{shen2016statistics} imply, as $\npinfty$
    \begin{equation*}
            \sum_{i = 2}^{\nsig + \nbkg} \hat{\eigvec}_{i, 2}^{2} \rightarrow \frac{1}{1 + \aspratio_{2}} \qas, \quad \sum_{i = 2}^{\nsig + \nbkg} \hat{\eigvec}_{2, i}^{2} \rightarrow \frac{1}{1 + \aspratio_{2}} \qas. 
    \end{equation*}
    Consequently, we have $\hat{\eigvec}_{i, 2} \rightarrow 0$ a.s. for all $i\neq 2$. Furthermore, from equation (8.17) in \citep{shen2016statistics}, we have:
    \begin{equation*}
        \sum_{i = i}^{\nsig + \nbkg} \hat{\eigvec}_{1, i}^{2} \overlineninfty \frac{1}{1 + \aspratio_{1}} \qas.
    \end{equation*}
    which directly implies $\hat{\eigvec}_{1, 2} \rightarrow 0$ a.s. as $\npinfty$. Thus, the result also holds for $j=2$. 
    
    The above arguments can be iteratively applied to subsequent indices $j=3,4,\dots,\nsig+\nbkg$ by analogous reasoning. Hence, we obtain the general result for arbitrary $i$ and $j$, completing the proof.
\end{proof}

\begin{lemma}[Asymptotic orthogonality of Gaussian projections]\label{lm:orthGauss}
Let \( a, b \in \mathbb{R}^d \) be unit vectors satisfying \( a^\top b = 0 \), and let \( X \in \mathbb{R}^{n \times d} \) be a random matrix with entries \( X_{ij} \overset{\text{i.i.d.}}{\sim} \mathcal{N}(0, 1) \). Then:
\[
\frac{1}{n} a^\top X^\top X b \xrightarrow{\text{a.s.}} 0 \quad \text{as } \npinfty.
\]
\end{lemma}

\begin{proof}
We first rewrite the expression as:
\[
\frac{1}{n} a^\top X^\top X b = \frac{1}{n} (X a)^\top (X b) = \frac{1}{n} \sum_{j=1}^n (Xa)_j (Xb)_j.
\]
Let \( z_i \coloneqq (Xa)_j (Xb)_j \), so that:
\[
\frac{1}{n} a^\top X^\top X b = \frac{1}{n} \sum_{i=1}^n z_j.
\]
Each row \( X_{j\bullet} \in \mathbb{R}^d \) of \( X \) is an i.i.d.\ sample from \( \mathcal{N}(0, I_d) \). Then:
\[
(Xa)_j = X_{j\bullet}^\top a, \quad (Xb)_j = X_{j\bullet}^\top b,
\]
and thus:
\[
z_i = (X_{j\bullet}^\top a)(X_{j\bullet}^\top b).
\]

Since \( X_{j\bullet} \sim \mathcal{N}(0, I_d) \), and \( a, b \in \mathbb{R}^d \) are fixed vectors, the random variables \( X_{j\bullet}^\top a \) and \( X_{j\bullet}^\top b \) are jointly Gaussian with:
\[
\EE[X_{j\bullet}^\top a] = 0, \quad \EE[X_{j\bullet}^\top b] = 0, \quad \text{and} \quad \text{Cov}(X_{j\bullet}^\top a, X_{j\bullet}^\top b) = a^\top b = 0.
\]
Hence, \( X_{j\bullet}^\top a \) and \( X_{j\bullet}^\top b \) are uncorrelated zero-mean Gaussian random variables and thus independent.

Therefore, since both are independent and centered,
\[
\EE[z_i] = \EE[(X_{j\bullet}^\top a)(X_{j\bullet}^\top b)] = \EE[X_{j\bullet}^\top a] \cdot \EE[X_{j\bullet}^\top b] = 0.
\]

Also, \( z_1, \ldots, z_n \) are i.i.d.\ with finite variance. Then by the strong law of large numbers,
\[
\frac{1}{n} \sum_{j=1}^n z_j \xrightarrow{\text{a.s.}} \EE[z_j] = 0.
\]
Thus,
\[
\frac{1}{n} a^\top X^\top X b \xrightarrow{\text{a.s.}} 0.
\]
\end{proof}

\section{Additional Details for Numerical Simulations (Section \ref{sec:experiments})}\label{app:simulation_details}
Below we describe in additional details of the synthetic data configurations and evaluation procedures underlying our numerical experiments. In all experiments we measure the subspace estimation error as the sine of the principal angle between the estimated and true signal spaces, averaged over 50 Monte Carlo trials.

\subsection{One-signal-one background-model--varying background strength}
We generated \(n=2000\) paired samples in \(d=800\) dimensions from the linear contrastive factor model (Eq.~\eqref{eq:model}).  The true signal subspace is one‐dimensional ($k=1$): $A = [\sqrt{\eigval_{\sig, 1}}, 0, \ldots, 0]^{\top}$ places a spike of magnitude \(\lambda_{A,1}=10\) in coordinate~1.  In $\bkg = [0, \sqrt{\eigval_{\bkg, 1}}, 0, \ldots, 0]^{\top}$, background variation is confined to the orthogonal coordinate~2 ($m=1$), with strength \(\lambda_{B,1}\) varied so that \(\lambda_{A,1}/\sqrt{\lambda_{B,1}}\) ranges over \(\{0.3125,\dots,0.666\}\).  For each method (PCA, $\px$, $\pxx$), we compute the sine of the principal angle between the estimated vector and the true signal direction (the first standard basis vector $e_{1}$).  A fixed truncation rank \(s=2\) is used to stabilize \(\pxx\) (though truncation does not affect this one‐dimensional signal case).  Results in Figure~\ref{fig:pcaplot_1} (left) show that, as background strength grows, PCA collapses to noise, \(\px\) eventually drifts onto the background axis, while \(\pxx\) remains tightly aligned with the true signal.

\subsection{One-signal-one background-model--varying aspect ratios}
We fixed \(n=500\) and varied \(d\) so that \(d/n\) spans \(\{0.1,\dots,1.8\}\).  Both signal and background spikes remain at \(\lambda_{A,1}=10\) in coordinate~1 and \(\lambda_{B,1}=500\) in coordinate~2.  We apply ordinary PCA, \(\px\), and \(\pxx\) with truncation \(s=2\).  The sine of the principal angle to \(e_{1}\) is plotted versus \(d/n\).  Figure~\ref{fig:pcaplot_1} (right) shows that while PCA and \(\px\) errors grow unbounded for large \(d/n\), $\pxx$ error remains uniformly low.

\subsection{Stability of \texorpdfstring{$\pxx$}{pxx}: truncation effects}
Figure~\ref{fig:2} (left) illustrates the instability of untruncated $\pxx$ for \(n=1000\) as \(d/n\) increases from 0.1 to 1.8.  With the same one‐dimensional signal (\(\lambda_{A,1}=10\)) and background (\(\lambda_{B,1}=500\)), direct solution of the generalized eigenproblem yields erratic directions at high aspect ratios.  Replacing the sample covariance \(S_n\) by its rank‐\(s\) approximation with \(s=2\) restores stable recovery.

Figure~\ref{fig:2} (right) examines $\pxx$’s sensitivity to \(s\).  Again with \(n=1000\), \(\lambda_{A,1}=10\), \(\lambda_{B,1}=500\), and \(d/n\in[0.1,1.8]\), we vary \(s\in\{2,\,0.1d,\,0.2d,\,0.4d\}\).  Moderate truncation (e.g.\ \(s=0.1d\)) best balances bias and variance, discarding noisy directions while retaining the subspace needed to enforce uniformity.

\subsection{Empirical high‐dimensional asymptotics}\label{sec:ExpmHighdim}
We fix \(n=500\) and let \(d/n\) vary for the fixed‐aspect‐ratio regime (Theorem~\ref{thm:dist}), using a five‐dimensional signal 
$\nsig = 5$. The signal loading matrix $A \in \mathbb{R}^{d \times 5}$ was defined such that the signal occupied the first five feature dimensions:
    \[ A = \begin{bmatrix} 
        \sqrt{\lambda_{\sig,1}} & 0 & 0 & 0 & 0\\ 
        0 & \sqrt{\lambda_{\sig,2}} &0 & 0 & 0 \\ 
        0 & 0 &\sqrt{\lambda_{\sig, 3}} & 0 & 0 \\
        0 & 0 & 0 & \sqrt{\lambda_{\sig, 4}} & 0 \\
        0 & 0 & 0 & 0 & \sqrt{\lambda_{\sig, 5}} & \\
        
        \vdots & \vdots & \vdots & \vdots & \vdots\\ 
        0 & 0 & 0 & 0 & 0
        \end{bmatrix}, 
    \]
    with the signal component variance per dimension $[\lambda_{\sig,1}, \lambda_{\sig,2}, \lambda_{\sig,3}, \lambda_{\sig,4}, \lambda_{\sig,5}]$ $ =$ $[50,25,20,15,10]$. The true signal subspace $\mathcal{U}_{A}$ is spanned by $[e_1, \dots, e_5]$. The background loading matrix $B \in \mathbb{R}^{\ndim \times 5}$ was defined orthogonally to $A$, affecting the last five feature dimensions :
    \[ B = \begin{bmatrix} 
        0 & 0 & 0 & 0 & 0\\ 
        \vdots & \vdots & \vdots & \vdots & \vdots\\ 
         0 & 0 & 0 & 0 & \sqrt{\lambda_{\bkg,5}}\\ 
        0 & 0 & 0 & \sqrt{\lambda_{\bkg,4}} & 0\\
        0 & 0 & \sqrt{\lambda_{\bkg,3}} & 0 & 0\\
        0 & \sqrt{\lambda_{\bkg,2}} & 0 & 0 & 0\\
        \sqrt{\lambda_{\bkg,1}} & 0 & 0 & 0 & 0\\
        \end{bmatrix}, 
    \]
    with the signal component variance per dimension $[\lambda_{\bkg,1}, \lambda_{\bkg,2}, \lambda_{\bkg,3}, \lambda_{\bkg,4}, \lambda_{\bkg,5}]$ $=$ $[500,400,300,200,100]$. Truncation rank is \(s=10\).  For the growing‐spike regime (Theorem~\ref{thm:dist2}), we scale both \(d\) and all spikes by a factor of 10.  
    As shown in Figure \ref{fig:3}, the empirical subspace
error of $\pxx$ tracks the theoretical prediction almost exactly, even as the spikes and dimension diverge, whereas $\px$ consistently fails to align with the true signal.

\subsection{Empirical validation with fixed aspect ratio}

To assess performance under a fixed aspect ratio of \(d/n=0.4\), we conducted simulations comparing standard PCA, \(\px\), $\pxx$, and the theoretical prediction.  Data were generated from the linear contrastive factor model (Eq.~\eqref{eq:model}) with signal and background each occupying five orthogonal directions in a \(p\)-dimensional space.  The signal subspace consisted of the first five canonical axes, with variances \([\lambda_{\sig,1}, \lambda_{\sig,2}, \lambda_{\sig,3}, \lambda_{\sig,4}, \lambda_{\sig,5}] = [20,20,15,10,10]\), while the background spanned the last five axes with variances \([\lambda_{\bkg,1}, \lambda_{\bkg,2}, \lambda_{\bkg,3}, \lambda_{\bkg,4}, \lambda_{\bkg,5}] = [500,500,200,100,100]\) in the “large noise’’ regime and \([100,100,50,25,25]\) in the “mild noise’’ regime.  We varied the sample size \(n\in\{100,500,5000\}\), keeping \(d=0.4\,n\), and performed 50 independent trials for each setting.  Subspace error was measured by the largest principal angle between the estimated and true signal subspaces.

Under large background noise (Table~\ref{tab:1}), both PCA and \(\px\) fail completely, yielding error near one, whereas $\pxx$ achieves low error (0.304 at \(n=100\), dropping to 0.212 at \(n=5000\)) with small variability (standard deviation from 0.037 to 0.005).  Even when background noise is mild (Table~\ref{tab:2}), \(\px\) only improves its average error at large \(n\) (0.222 at \(n=5000\)) but remains highly unstable ($sd=0.062$).  In contrast, $\pxx$ matches or betters \(\px\)’s mean error (0.212) while reducing its standard deviation to 0.004.  These results demonstrate that $\pxx$ not only lowers estimation error but also yields far more consistent recovery of the signal subspace than either PCA or the alignment‐only method.

\begin{table}[ht]
  \caption{Fixed Aspect Ratio (\(d/n=0.4\)) under Large Background Noise}
  
  \label{tab:1}
  \centering
  \begin{tabular}{lcccc}
    \toprule
    Setting     & PCA     & $\px$ &  $\pxx$ & $\pxx$ theory\\
    \midrule
    $n = 100, p = 40$ & 0.999 (0)  & 0.999 (0) & 0.304 (0.037) &0.205    \\
    $n = 500, p = 200$     & 0.999 (0) & 0.999 (0) & 0.225 (0.010) & 0.205      \\
    $n = 5000, p = 2000$     & 0.999 (0) & 0.999 (0.001) & 0.212 (0.005) & 0.205 \\
    \bottomrule
  \end{tabular}
\end{table}

\begin{table}[ht]
  \caption{Fixed Aspect Ratio (\(d/n=0.4\)) under Mild Background Noise}
  
  \label{tab:2}
  \centering
  \begin{tabular}{lcccc}
    \toprule
    Setting     & PCA     & $\px$ &  $\pxx$ & $\pxx$ theory\\
    \midrule
    $n = 100, p = 40$ & 0.999 (0)  & 0.996 (0.011) & 0.293 (0.038) & 0.205    \\
    $n = 500, p = 200$     & 0.999 (0) & 0.713 (0.287) & 0.227 (0.010) & 0.205      \\
    $n = 5000, p = 2000$     & 0.999 (0) & 0.222 (0.062) & 0.212 (0.004) & 0.205 \\
    \bottomrule
  \end{tabular}
\end{table}

\section{Additional Details for Real Data Applications (Section \ref{sec:experiments})}\label{app:real_data_details}

\subsection{Corrupted MNIST data}

This section details the experimental setup for the qualitative evaluation on noisy MNIST digits, the results of which are presented in Figure~\ref{fig:MNIST}. The aim was to visually assess signal disentanglement from structured background noise using a dataset of $n=5000$ paired synthetic images $(x_i, x_i^+)$.

\paragraph{Data generation.} The signal components were images of digits \texttt{'0'} and \texttt{'1'} from the MNIST dataset~\citep{deng2012mnist}. Each original $28\times28$ grayscale MNIST image had its pixel values normalized to the range $[0,1]$; let $M_i$ denote such a processed digit. Backgrounds were sourced from the "grass" synset of the ImageNet dataset~\citep{deng2009imagenet}. Full ImageNet images were converted to grayscale, from which $28\times28$ patches were randomly cropped. Each cropped grass patch was then also normalized so its pixel values ranged from $[0,1]$. Let $G_{i}$ be independently sampled and processed grass patches. Paired images were then constructed through pixel-wise addition:
\begin{align*}
    x_i &=  M_i \\
    x_i^+ &= 0.25 \cdot M_i + G_{i}.
\end{align*}
This setup uses the same MNIST digit $M_i$ for both $x_i$ and its positive pair $x_i^+$, but with different grass backgrounds, and with the MNIST signal scaled down by a factor of $0.25$ in the $x_i^+$ view. Each resulting   synthetic image was flattened into a $p=784$ dimensional vector. Finally, before applying the PCA methods, the set of all $x_i$ vectors (forming matrix $X$) was globally mean-centered (i.e. the mean vector of $X$ was subtracted from each row $x_i$). Similarly, the set of all $x_i^+$ vectors (forming matrix $X^+$) was globally mean-centered. % Optional: Pixel values were subsequently clipped to the $[0,1]$ range.

\subsection{Single-cell RNA sequencing data}

We investigate the impact of immune stimulation on peripheral blood mononuclear cells (PBMCs) using single-cell RNA sequencing data from  \cite{kang2018multiplexed}. This dataset includes scRNA-seq profiles from 8 human donors collected under two conditions: control and interferon-$\beta$ (IFN-$\beta$) stimulation. The processed data comprise 14,619 control cells and 14,446 stimulated cells, spanning 8 curated immune cell types. These include both IFN-$\beta$–responsive populations (e.g., monocytes) and cell types with minimal transcriptional response (e.g., B cells), providing a natural testbed for studying condition-invariant structure. While cells are not explicitly paired in the raw data, shared donor identities and consistent annotations allow post hoc matching for comparative analysis. Each cell is labeled with a curated cell type, offering a stable reference to assess shifts in population structure under perturbation. Our goal is to demonstrate that the proposed $\pxx$ method more effectively captures biologically meaningful, condition-invariant structure in real-world single-cell data. To this end, we apply both PCA and $\pxx$, and visualize the resulting embeddings by applying UMAP~\cite{mcinnes2018umap} to the top 50 components of each projection. We compare the structures side-by-side across control and stimulated conditions to assess the preservation and alignment of cell-type clusters.

To prepare the dataset for contrastive analysis, we first filtered the gene space to retain only informative and biologically relevant features. We excluded non-coding genes (e.g., lncRNA, miRNA, pseudogenes) by annotating Ensembl gene IDs using the mygene API and removing entries with non-protein-coding biotypes. We then selected the top 2,000 highly variable genes (HVGs) using the Seurat v3 method implemented in Scanpy. This yielded a final feature space shared across 29,065 cells, comprising 14,619 control and 14,446 stimulated cells.

Because cells are not explicitly paired in the original dataset, we constructed cell pairs post hoc using donor identity and local structural alignment. For each donor, we performed independent normalization, log-transformation, and PCA on cells from the control and stimulated conditions. We clustered cells within each condition using Leiden clustering and computed the centroids of each cluster in the 50-dimensional PCA space. Clusters were matched across conditions by cosine similarity of their centroids. Within each matched pair of clusters, we applied nearest-neighbor matching in PCA space to identify at most 300 well-aligned cell pairs. This strategy ensured both inter-condition comparability and intra-donor consistency, resulting in 9,268 matched control and 9,268 matched stimulated cells across all 8 donors.

To evaluate the stability of cellular identity across conditions, we apply PCA and our proposed $\pxx$ to the full set of matched control and stimulated cells. Both methods produce 50-dimensional latent embeddings, which we visualize using UMAP. In the $\pxx$ framework, the stimulated cells are treated as the target dataset and the control cells as the target-plus dataset, enabling the method to extract components that emphasize shared structure while suppressing condition-specific variation. This design aligns with our goal of capturing invariant cell populations under perturbation.

\begin{figure}[ht]
\centering
\includegraphics[width=1.0\linewidth]{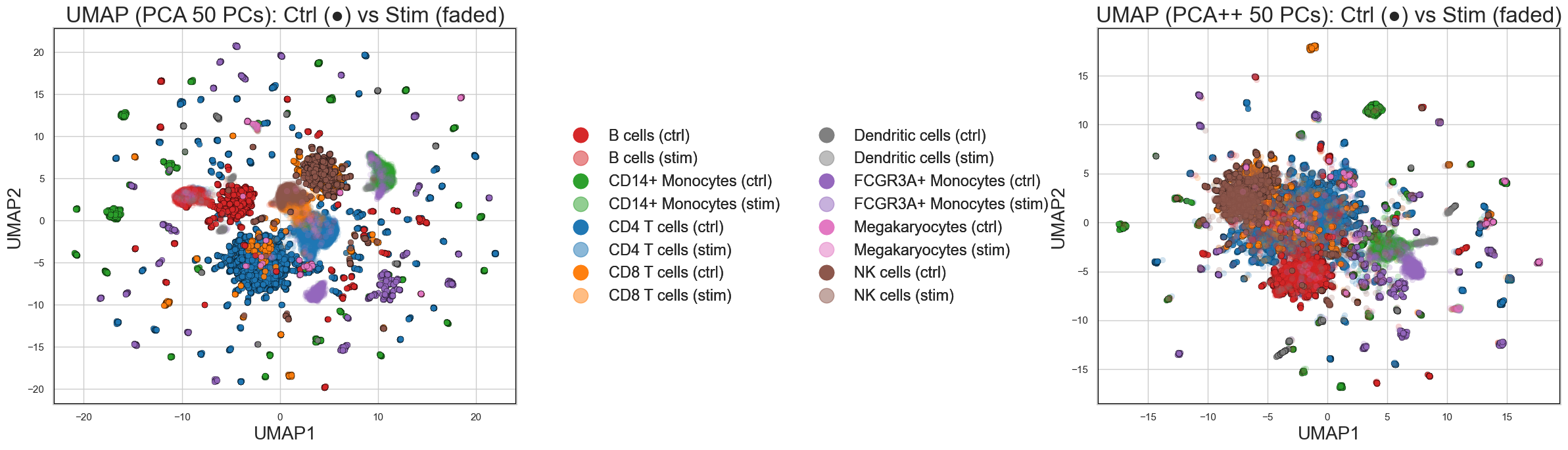}
\caption{\small \textbf{UMAP visualizations of PCA (left) vs.\ $\pxx$ (right) on all cells.} Both plots show UMAP projections of the top 50 components applied to all matched PBMCs. The PCA embedding (left) separates control and stimulated cells—even within the same cell type—due to confounding condition-specific variance. In contrast, the $\pxx$ embedding (right), computed using stimulated cells as the target and control cells as auxiliary input, better aligns the two conditions. Invariant populations such as CD4 T cells and B cells remain coherent across stimulation, while more responsive types (e.g., monocytes) exhibit increased spread, reflecting true biological variability.}
\label{fig:pcapp_umap}
\end{figure}

The left panel of Figure \ref{fig:pcapp_umap} shows the UMAP projection of PCA embeddings. While cell types are well-separated, the embeddings exhibit clear batch effects—control and stimulated cells of the same type often appear as disconnected clusters. This suggests that PCA captures both cell identity and stimulation-driven variance. In contrast, the right panel of Figure \ref{fig:pcapp_umap} shows UMAP on $\pxx$ embeddings. Here, stimulated and control cells are more co-localized, especially for stable populations like CD4 T cells and B cells, indicating that $\pxx$ better preserves condition-invariant structure. Although some separation remains for highly responsive populations (e.g., monocytes), $\pxx$ enables clearer alignment of shared cellular manifolds across conditions.

To better interpret the results, we highlight key observations from Figure \ref{fig:pca_vs_pxpp}, which contrasts UMAP visualizations derived from PCA (a, b, c) and $\pxx$ (d, e, f) projections. In the PCA embeddings, cell types are well separated in both control and stimulated conditions; however, control and stimulated cells of the same type frequently occupy non-overlapping regions. This indicates that standard PCA captures both cell identity and condition-induced variation, leading to separation even within the same cell type. In contrast, $\pxx$ (d, e, f) produces embeddings where control and stimulated cells are better aligned within each cell type. Although the overall separation between cell types becomes slightly less distinct, the coherence of shared structure across conditions improves markedly. For example, invariant populations such as CD4 T cells, B cells, and NK cells remain tightly clustered regardless of stimulation, highlighting $\pxx$’s ability to emphasize stable transcriptional programs. Importantly, these findings are consistent with known biology: these cell types are relatively insensitive to IFN-$\beta$ stimulation and exhibit minimal transcriptional changes. In contrast, more responsive cell types such as monocytes appear more dispersed, reflecting biologically meaningful variation rather than projection-induced noise. These results demonstrate that $\pxx$ more effectively disentangles cell identity from stimulation effects, enabling clearer interpretation of shared cellular structure across perturbed and unperturbed states.

{\color{black} 
we have also included a comprehensive comparison across several methods, including standard PCA, alignment-only $\px$ (which lacks the uniformity constraint), $\pxx$, as well as additional baseline dimensionality reduction techniques (UMAP \citep{mcinnes2018umap}, t-SNE \citep{van2008visualizing}, Robust PCA~\citep{candes2011robust}). All methods were evaluated on the same single-cell RNA-seq data, focusing on their ability to recover cell-type groupings in both control and stimulated conditions. 

The adjusted Rand index [\citealp[ARI]{hubert1985comparing}]  (lower is better in this context) for each method is summarized in Table \ref{tab:real}.

\begin{table}[ht]
  \centering
  \caption{Performance comparison across different cell types}
  \label{tab:real}
  \begin{tabular}{lcccc}
    \toprule
    & Cells & B Cells & NK Cells & All \\
    \midrule
    PCA & 0.2084 & 0.2977 & 0.2502 & 0.1478 \\
    $\px$ & 0.0036 & 0.0242 & 0.0299 & 0.014 \\
    $\pxx$ & 0.0008 & 0.005 & 0.0049 & 0.0172 \\
    U-MAP & 0.2031 & 0.2506 & 0.2078 & 0.1392 \\
    t-SNE & 0.0233 & 0.0967 & 0.0477 & 0.0217 \\
    Robust PCA & 0.0399 & 0.1907 & 0.1314 & 0.0443 \\
    \bottomrule
  \end{tabular}
\end{table}

As shown, both $\px$ and $\pxx$ dramatically reduce the ARI compared to standard PCA and other popular methods, indicating much better mixing of the two conditions within cell types. Importantly, adding the uniformity constraint in $\pxx$ further improves the results over $\px$, especially in B cells and NK cells. These quantitative results highlight the utility of both the alignment and uniformity components. $\pxx$ achieves superior separation of invariant populations compared to alternative approaches.
}

{\color{black}

\section{Additional Discussion on cPCA\label{app:cpca_pcaplus}}

Although there is a fundamental distinction between \texttt{cPCA} and $\pxx$, one reviewer suggested that \texttt{cPCA} could nonetheless be adapted to address the problem solved by $\pxx$. Specifically, the reviewer proposed constructing a synthetic foreground and background via
$x_{f} = (x + x^{+})/2$ and $x_{b} = (x - x^{+})/2$, and then applying \texttt{cPCA} to the pair $(x_{f}, x_{b})$.
While this construction yields a \texttt{cPCA}-like method that is theoretically valid at the population level, our analysis and experiments demonstrate that it is substantially less stable and reliable in finite-sample regimes compared to $\pxx$.

We can see that at the population level, the signal is perfectly isolated. Under the contrastive factor model, the population covariances of the synthesized data are:
\[
\Sigma_f = \mathbb{E}[x_f x_f^{\top}] = AA^{\top} + \frac{1}{2}BB^{\top} + \frac{1}{2}I_d
\]
\[
\Sigma_b = \mathbb{E}[x_b x_b^{\top}] = \frac{1}{2}BB^{\top} + \frac{1}{2}I_d
\]
Therefore, the difference $\Sigma_f - \Sigma_b = AA^{\top}$ perfectly recovers the signal covariance. The practical challenge, however, arises from finite-sample estimation error. Define foreground matrix $X_{f} = (X + X^{+})/2$ and background matrix $X_{b} = (X - X^{+})/2$. The sample covariances of the synthesized data are $\widehat{\Sigma}_f = X_{f}^{\top} X_{f}/n$ and  $\widehat{\Sigma}_b = X_{b}^{\top} X_{b}/n$, respectively.  

\textbf{Regarding cPCA:} The reviewer's proposed approach relies on the matrix subtraction $\hat{\Sigma}_{f} - \alpha \hat{\Sigma}_{b}$, this subtraction of two large, noisy matrices can amplify noise, potentially overwhelming the true signal and leading to unstable eigenvectors.

\textbf{Regarding cPCA++:} 
A \texttt{cPCA}-like extension, referred to as \texttt{cPCA++} \citep{SALLOUM2022108378}, has also been proposed in prior work. This method identifies contrastive directions by solving a generalized eigenvalue problem of the form
$\widehat{\Sigma}_{b}^{-1} \widehat{\Sigma}_{f}$,
where $\widehat{\Sigma}_{f}$ and $\widehat{\Sigma}_{b}$ are the empirical covariance matrices of the foreground and background data, respectively. As we demonstrate in our paper (Figure \ref{fig:2}), directly inverting $\widehat{\Sigma}_{b}$ is numerically unstable when dimension $d$ is large. The reviewer might suggest stabilizing this by using a truncated pseudoinverse, similar to our $\pxx$. However, this reveals a deeper conceptual issue: this procedure is equivalent to projecting the foreground data $X_{f}$ onto the principal subspace of the background data $X_{b}$ and then performing PCA. Since the synthesized background $X_{b}$ contains no signal $A$ by construction, its principal subspace is also signal-free. Projecting the foreground onto this signal-free subspace would annihilate the very signal we aim to recover. Therefore, a truncated \texttt{cPCA++} would fundamentally fail.

\subsection*{Empirical validation:}

To investigate these different approaches empirically, we performed a new set of experiments comparing $\pxx$ against alternative methods proposed by the reviewer. 

\begin{enumerate}
    \item \textbf{Baselines on synthesized data:} Following the reviewer's suggestion, we evaluated methods that operate on synthesized foreground $X_{f} = (X + X^{+})/2$ and background  $X_{b} = (X - X^{+})/2$ data. This includes standard \texttt{cPCA} and \texttt{cPCA++} (based on a difference of covariances). This allows us to directly test the stability of the "subtract-then-decompose" approach versus our "decompose-from-cross-covariance" method.
    \item \textbf{CCA:} We also introduced CCA as a canonical and highly relevant baseline. We apply it by treating the paired data matrices, $X$ and $X^+$, as the two views. In our model, the shared signal is the sole source of population-level correlation between these views. Therefore, CCA, which finds directions of maximal correlation, is theoretically suited for recovering the signal subspace and serves as a strong benchmark.
\end{enumerate}
 
We consider the following experiment setup:
\begin{itemize}
    \item \textbf{Experimental setup:} The experiment follows the $n=500$, $k=5$ fixed aspect ratio setting from Sec E.4. For `cPCA`, we set $\alpha=1$. This is the most principled choice as it yields an unbiased population estimator 
$$\mathbb{E}[\hat{\Sigma}_{f} - \hat{\Sigma}_{b}] = AA^{\top} + BB^{\top}/2 + I_d/2 - BB^{\top}/2 - I_d/2 = AA^{\top}.$$ We tested two scenarios: one with moderate background noise and one with strong background noise.

\item \textbf{Signal variances:} $[50, 25, 20, 15, 10]$.

\item \textbf{Moderate noise background:} $[100, 50, 40, 30, 20]$.

\item \textbf{Large noise background:} $[500, 400, 300, 200, 100]$.
\end{itemize}
 
\subsection*{Experimental results (subspace error):}
 
The results are provided in the following Table \ref{tab:aspect_ratio_comparison} and \ref{tab:aspect_ratio_comparison2}. As the results clearly show, the \texttt{cPCA} adaptation is highly unstable even with its optimal hyperparameter setting $\alpha=1$. This, combined with the conceptual issue in the truncated \texttt{cPCA++} approach, demonstrates that these methods are not well-suited for this problem. In contrast, $\pxx$ method remains stable, highlighting the robustness of our GEP formulation.  

\begin{table}[ht]
\centering
\begin{tabular}{lcccccccccc}
\toprule
Aspect Ratio & 0.1 & 0.3 & 0.5 & 0.7 & 0.9 & 1.1 & 1.3 & 1.5 & 1.7 & 1.8 \\
\midrule
\texttt{cPCA} & 0.293 & 0.819 & 0.430 & 0.977 & 0.284 & 0.626 & 0.813 & 0.581 & 1.000 & 0.964 \\
\texttt{cPCA++} & 1.000 & 1.000 & 1.000 & 1.000 & 1.000 & 1.000 & 1.000 & 0.999 & 0.999 & 1.000 \\
\texttt{CCA} & 0.160 & 0.488 & 0.999 & 1.000 & 1.000 & 1.000 & 1.000 & 0.999 & 1.000 & 1.000 \\
\pxx & 0.125 & 0.212 & 0.250 & 0.275 & 0.311 & 0.347 & 0.375 & 0.388 & 0.415 & 0.401 \\
\pxx\ theory & 0.104 & 0.179 & 0.229 & 0.268 & 0.301 & 0.330 & 0.356 & 0.379 & 0.400 & 0.410 \\
\bottomrule
\end{tabular}
\caption{Performance comparison across different aspect ratios with moderate noise background.}
\label{tab:aspect_ratio_comparison}
\end{table}

\begin{table}[ht]
\centering
\begin{tabular}{lcccccccccc}
\toprule
Aspect Ratio & 0.1 & 0.3 & 0.5 & 0.7 & 0.9 & 1.1 & 1.3 & 1.5 & 1.7 & 1.8 \\
\midrule
\texttt{cPCA} & 1.000 & 1.000 & 0.999 & 0.941 & 0.998 & 0.999 & 0.998 & 1.000 & 0.998 & 0.996 \\
\texttt{cPCA++} & 1.000 & 1.000 & 1.000 & 1.000 & 1.000 & 1.000 & 1.000 & 1.000 & 1.000 & 1.000 \\
\texttt{CCA} & 0.154 & 0.369 & 0.999 & 1.000 & 1.000 & 1.000 & 1.000 & 1.000 & 1.000 & 1.000 \\
\pxx & 0.156 & 0.185 & 0.230 & 0.283 & 0.321 & 0.339 & 0.379 & 0.353 & 0.410 & 0.447 \\
\pxx\ theory & 0.104 & 0.179 & 0.229 & 0.268 & 0.301 & 0.330 & 0.356 & 0.379 & 0.400 & 0.410 \\
\bottomrule
\end{tabular}
\caption{Performance comparison across different aspect ratios with large noise background.}
\label{tab:aspect_ratio_comparison2}
\end{table}

\section{Additional discussion on the relationship to CCA\label{app:cca}}

While both CCA and $\pxx$ methods aim to find shared structure between paired datasets, they differ crucially in their objectives and constraints, leading to significant performance gaps in noisy, high-dimensional settings.

\subsection*{Objective functions and constraints:}
\begin{itemize}
    \item \textbf{Standard CCA:} Seeks two distinct projection matrices, $U$ and $V$, that maximize the correlation between the projected views $U^{\top} X$ and $V^{\top} X^+$. The objective for the leading component is:
$$\max_{u,v} \frac{u^{\top} (X^{\top} X^+) v}{\sqrt{(u^{\top} (X^{\top} X) u) (v^{\top} (X^{+T} X^+) v)}}$$
The key instability lies in the denominator, which normalizes by variance from both views ($X$ and $X^{+}$). In our problem setting, both views contain large, independent background and noise components, making this normalization highly susceptible to noise amplification. Although one can also revise   CCA for the positive-pair setting by imposing constrain $U=V$, the objective becomes a Generalized Rayleigh Quotient, which, to our knowledge, does not have the same closed-form solution as standard CCA.
\item \textbf{$\pxx$:} Our formulation makes two critical changes that enhance stability 
\begin{itemize}
    \item \textbf{Shared projection space:} It enforces $U=V$, searching for a single, shared subspace that captures the signal common to both views.
    \item \textbf{Robust normalization:} Instead of normalizing by the noisy variance from both views, PCA++ maximizes the shared covariance $v^{\top} (X^{\top} X^+) v$ subject to a hard uniformity constraint on only one view $v^{\top} (X^{\top} X) v = 1$. This constraint acts as a powerful regularizer, anchoring the solution to a more stable variance structure and preventing distortion from the background and noise present in the second view $X^+$.
\end{itemize}
\end{itemize}

\subsection*{Empirical Performance:}

This difference in formulation has a dramatic impact on performance. As our new experiments (provided in Table \ref{tab:aspect_ratio_comparison} and \ref{tab:aspect_ratio_comparison2} of Appendix \ref{app:cpca_pcaplus}) demonstrate:
\begin{itemize}
    \item \textbf{Standard CCA is highly unstable} in the settings we study. It fails to recover the signal, with its performance collapsing as dimensionality or noise increases.
    \item \textbf{$\pxx$ remains stable and robust}, successfully recovering the signal subspace across all tested regimes.
\end{itemize}
This confirms that our specific choice of a shared projection space and a single-view uniformity constraint is essential for reliable signal recovery in this problem setting. 

}

\end{document}